\documentclass{article}

\usepackage{microtype}
\usepackage{graphicx}
\usepackage{xcolor}
\usepackage{booktabs} % for professional tables

\usepackage{hyperref}
\usepackage{amsthm}
\usepackage{amsmath}
\usepackage{amssymb}
\usepackage{booktabs}
\usepackage{float}
\usepackage{multirow}
\usepackage{array}
\usepackage{graphicx}  %插入图片的宏包
\usepackage{float}  %设置图片浮动位置的宏包
\usepackage{subcaption}
\usepackage{color}
\usepackage{amsmath}
\theoremstyle{plain}
\newtheorem{theorem}{Theorem}[section]

\theoremstyle{definition}
\newtheorem{definition}[theorem]{Definition}

\theoremstyle{remark}

\usepackage[ruled,vlined]{algorithm2e}
\newcommand{\jiawei}[1]{{\color{black}#1}}
\newcommand{\xiaowei}[1]{{\color{black}#1}}
\newcommand{\yq}[1]{{\color{black}#1}}
\usepackage[accepted]{icml2021}

\icmltitlerunning{ICML 2024}

\begin{document}

\twocolumn[
\icmltitle{SuDA: Support-based Domain Adaptation for Sim2Real Motion Capture
with Flexible Sensors}

% It is OKAY to include author information, even for blind
% submissions: the style file will automatically remove it for you
% unless you've provided the [accepted] option to the icml2021
% package.

% List of affiliations: The first argument should be a (short)
% identifier you will use later to specify author affiliations
% Academic affiliations should list Department, University, City, Region, Country
% Industry affiliations should list Company, City, Region, Country

% You can specify symbols, otherwise they are numbered in order.
% Ideally, you should not use this facility. Affiliations will be numbered
% in order of appearance and this is the preferred way.
\icmlsetsymbol{equal}{*}

\begin{icmlauthorlist}
\icmlauthor{Jiawei Fang}{xmu}
\icmlauthor{Haishan Song}{xmu}
\icmlauthor{Chengxu Zuo}{xmu}
\icmlauthor{Xiaoxia Gao}{xidian}
\icmlauthor{Xiaowei Chen}{xmu}
\icmlauthor{Shihui Guo}{xmu}
\icmlauthor{Yipeng Qin}{Cardiff}
\end{icmlauthorlist}

\icmlaffiliation{xmu}{School of Informatics, Xiamen University, Fujian, China}
\icmlaffiliation{xidian}{Department of Computer Science and Technology, Xidian University, Shanxi, China}
\icmlaffiliation{Cardiff}{School of Computer Science\&Informatics, Cardiff University, Wales, UK}

\icmlcorrespondingauthor{Shihui Guo}{guoshihui@xmu.edu.cn}
% \icmlcorrespondingauthor{Eee Pppp}{ep@eden.co.uk}

% You may provide any keywords that you
% find helpful for describing your paper; these are used to populate
% the "keywords" metadata in the PDF but will not be shown in the document
\icmlkeywords{Machine Learning, ICML}

\vskip 0.3in
]

% this must go after the closing bracket ] following \twocolumn[ ...

% This command actually creates the footnote in the first column
% listing the affiliations and the copyright notice.
% The command takes one argument, which is text to display at the start of the footnote.
% The \icmlEqualContribution command is standard text for equal contribution.
% Remove it (just {}) if you do not need this facility.

\printAffiliationsAndNotice{}
% leave blank if no need to mention equal contribution

% \printAffiliationsAndNotice{\icmlEqualContribution} 
% otherwise use the standard text.

\begin{abstract}
% \jiawei{Flexible sensors show promise for human motion capture due to their wearability, privacy preservation, and minimal constraints on natural movement. However, existing flexible sensor-based Mocap methods are deep-learning based, and their training requires large and diverse labeled data.}
\yq{Flexible sensors hold promise for human motion capture (MoCap), offering advantages such as wearability, privacy preservation, and minimal constraints on natural movement. However, existing flexible sensor based MoCap methods rely on deep learning and necessitate large and diverse labeled datasets for training.}
 % While machine learning models excel at modeling the complex non-linear relationships between sensor signals and motions that result from variations across users, motions, and wearing positions, their training requires large and diverse labeled data. 
These data typically need to be collected in MoCap studios with specialized equipment and substantial manual labor, making them difficult and expensive to obtain at scale. Thanks to the high-linearity of flexible sensors,  we address this challenge by proposing a novel \emph{Sim2Real} Mocap solution based on domain adaptation, eliminating the need for labeled data yet achieving comparable accuracy to supervised learning. Our solution relies on a novel \emph{Support-based} Domain Adaptation method, namely \textbf{SuDA}, which aligns the \emph{supports} of the predictive functions rather than the instance-dependent distributions between the source and target domains. \yq{Extensive} \jiawei{experimental results demonstrate the effectiveness of our method and its superiority over state-of-the-art distribution-based domain adaptation methods in our task.}
% The mapping between flexible sensor signals and motion is usually built with machine learning approaches due to their ability to model complex non-linear relationships and adapt to variations across users, motions, and wearing positions.
% However, obtaining accurate motion labels to pair with sensor data typically requires specialized equipment in controlled studios and substantial manual labor, posing formidable challenges in gathering diverse, realistic training data at scale.

% This makes our adaptation method applicable to unpaired simulated and real data with inherently different body shapes, motions, sensor positions, etc., as long as they share similar ranges of joint motion.

\end{abstract}

\section{Introduction}

\jiawei{Human motion capture (MoCap) has garnered significant attention for its vast potential in comprehending human intentions and its capacity to control physical devices and systems effectively~\cite{yi2022physical, shen2023x}.} 
\jiawei{As an alternative to vision-based and inertial-based solutions, flexible sensors are emerging as they offer the benefits of bio-compatibility, high stretchability, lightweight, and easy integration into clothing, enabling long-term monitoring of human physical status~\cite{zazoum2022recent,chen2023dispad}. Hence, there has been a growing demand to incorporate wearable flexible sensors into MoCap~\cite{luo2021learning, zhang2022wearable, luo2023intelligent}.}

% \yq{As a promising form of human-computer interaction, MoCap finds applications in diverse fields including robotics, haptics, biomechanical studies, healthcare, and the metaverse.}
% As a new type of human-computer interaction, Mocap could be applied in diverse fields, including robotics, haptics bio-mechanical studies, health care and metaverse. }
% Optical tracking and inertial sensors are prevalent in mocap solutions for film, animation, and gaming industries, with some studies exploring integrated approaches combining both sensor types. While these solutions have excelled in professional applications, they may not be optimal for consumer-level use, where the emphasis is on prioritizing user experience comfort and convenience.

\begin{figure}[t]
% \begin{figure}[bp]
% \centering
% \hspace{=0.01cm}
\centering
% \flushleft
\includegraphics[width=0.90\linewidth]{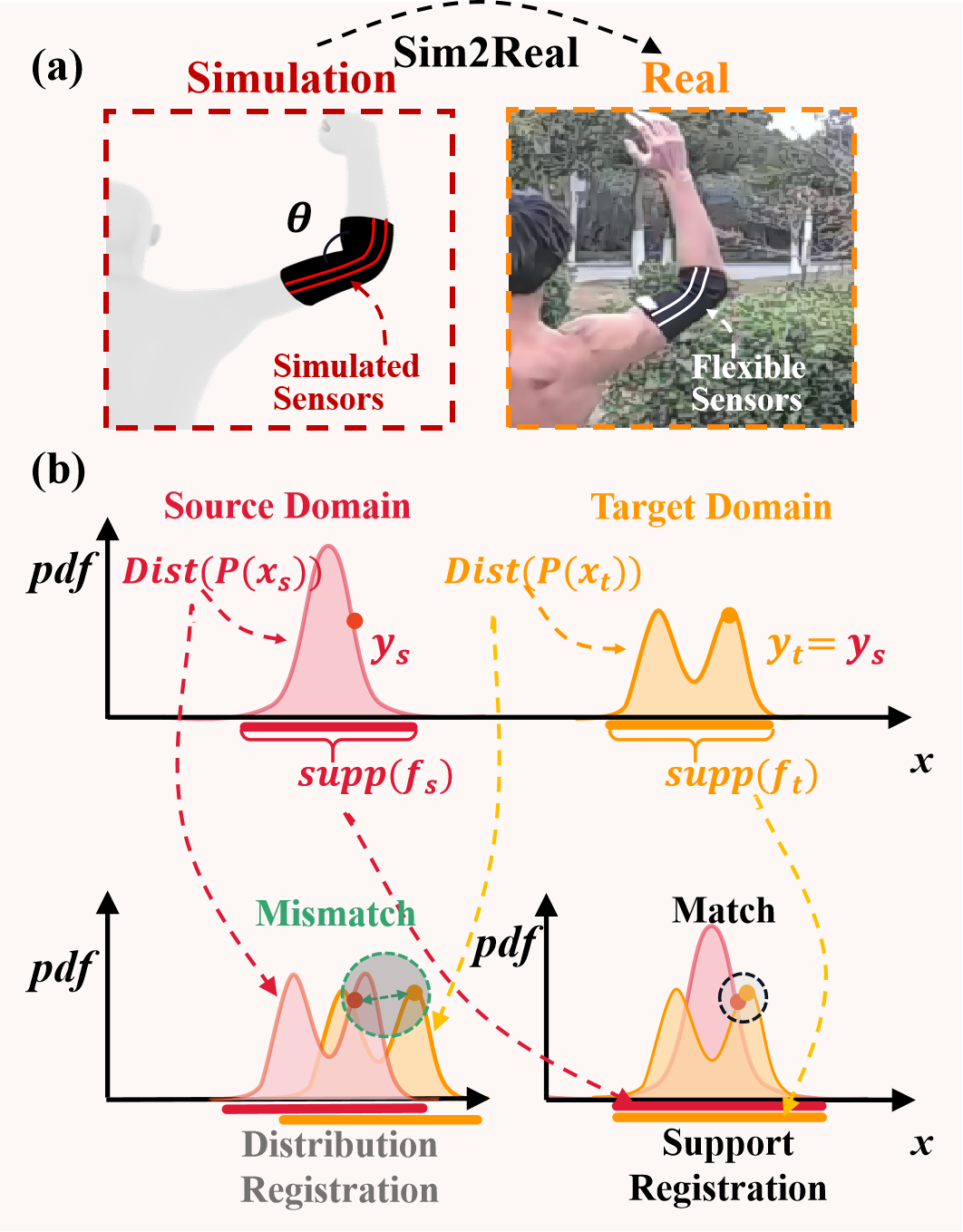}
\vspace{-2mm}
\caption{(a) Our Sim2Real approach simulates flexible sensors to predict joint angle in the real world. (b) \textbf{Distribution  Registration} registers the feature distributions {\textcolor[RGB]{222,26,55}{\textbf{$Dist(P(x_s))$}}} and {\textcolor[RGB]{255,152,0}{\textbf{$Dist(P(x_t))$}}} of the source and target domains, which fails to match points of the same label {\textcolor[RGB]{222,26,55}{\textbf{$y_s$}}}$=${\textcolor[RGB]{255,152,0}{\textbf{$y_t$}}} when the two distributions are inherently different.
In contrast, the proposed \textbf{Support Registration} registers the function supports {\textcolor[RGB]{222,26,55}{\textbf{$supp(f_s)$}}} and {\textcolor[RGB]{255,152,0}{\textbf{$supp(f_t)$}}} that are independent of the specific distributions, thereby guaranteeing successful domain adaptation.}
\vspace{-5mm}

\label{fig:teaser}
\end{figure}

% \jiawei{As an alternative to vision-based and inertial-based solutions, flexible sensors are emerging as they offer the benefits of bio-compatibility, high stretchability, lightweight, and easy integration into clothing, enabling long-term monitoring of human physical status~\cite{zazoum2022recent,chen2023dispad}. Hence, there has been a growing demand to incorporate wearable, flexible sensors into Mocap~\cite{luo2021learning, zhang2022wearable, luo2023intelligent} }

% human-computer interfaces~\cite{liu2022functionalized}, soft robotics~\cite{ying2021anti}, etc.

Deep neural networks have become the leading methodology for flexible sensor-based MoCap~\cite{mathis2020primer, zuo2023self} due to their strengths in modeling the complex relationships between sensor signals and motions. However, the conflict between data-hungry deep neural networks and the costly, labor-intensive process of collecting paired motion and sensor data poses a challenge in the real-world application of flexible sensor-based Mocap. Specifically, acquiring such paired data involves setting up motion capture studios equipped with both sensor reading collection equipment and experienced staff, which is expensive and labor-intensive to conduct at scale. A full-fledged optical MoCap setup typically costs between \$25K and \$500k.

Thanks to the high-linearity of flexible sensors, our paper addresses the conflict mentioned above by proposing a novel {\it Sim2Real} solution, which eliminates the need for any paired motion and sensor data yet achieves comparable accuracy.
Specifically, a {\it support-based} domain adaptation (SuDA) method is proposed to align the mapping between sensor and motion from the simulated (source) domain to the real-world (target) domain in a simple yet effective way.
In contrast to state-of-the-art distribution-based domain adaptation methods ({\it i.e.}, DiDA) that register the instance-dependent distributions between the source and target domains (Fig.~\ref{fig:teaser}), SuDA registers the {\it supports} of the predictive functions instead, making it independent of distribution variations resulting from body shapes, motions, sensor positions, etc.. Without loss of generality, we empirically evaluate our solution by predicting the joint angle using a smart pad equipped with two flexible sensors and worn on users' joints.
Extensive experimental results demonstrate the effectiveness of our method and its superiority over state-of-the-art distribution-based domain adaptation methods in human motion capture.

Our contributions are three-fold:
\begin{itemize}
    % \item we introduce a \emph{Sim2Real} method that achieves comparable accuracy to real data without any real data label, which could significantly reduce data annotation costs.
    \item We propose a \emph{Sim2Real} solution for human motion capture using flexible sensors, eliminating the need for an expensive real-world data collection process yet achieving comparable accuracy to supervised learning.
    \item We propose a novel {\it support-based} domain adaptation method (SuDA) that \jiawei{registers} predictive function supports rather than data distributions, leading to more accurate and generalizable results. 
    % \item we propose an innovative transfer learning method based on domain support registration to reduce gap between simulation and real data. Experimental results demonstrate its significant superiority over state-of-the-art domain distribution based transfer learning methods. This makes our adaptation method applicable to unpaired simulated and real data with inherently different body shapes, motions, sensor positions, etc., as long as they share similar ranges of joint motion.
    \item We conduct extensive experiments covering various application scenarios (different users, motions, wearing positions and joints) to show the effectiveness of our method.

\end{itemize}
\vspace{-0.5cm}
\section{Related Work}

% Head 2

\subsection{Motion Capture with Flexible Sensors}

Human motion capture has been actively explored due to its competence in learning profound knowledge about human motion from raw sensor inputs~\cite{wang2022distributionaware}. 
In addition to established motion capture systems using the multi-camera matrix (such as Vicon), alternative methods \xiaowei{mainly} include vision-based~\cite{tian2023recovering} and IMU-based approaches~\cite{yi2022physical,mollyn2023imuposer}.

Flexible sensors have been used for long-term monitoring of human physical status, precisely motion capture~\cite{zhang2022wearable,chen2023dispad}, human-computer interfaces~\cite{liu2022functionalized, zhang2020sensock,fang2023handwriting}, soft robotics~\cite{guo2023touch}, etc., for their advantages of bio-compatibility, high stretch-ability, lightweight, and ease of integration within clothing~\cite{zazoum2022recent}.

For MoCap, existing methods have explored the use of flexible sensors in tracking the motion of the full body~\cite{chen2023fullbody}, upper body~\cite{jin2020identifying,zhou2023mocapose}, fingers~\cite{glauser2019interactive}, lower limbs \cite{kwak2020wireless}, elbow joints \cite{abro2019novel,chen2023dispad}, and knee joints \cite{yu2021development}.

% The problem we aim to address is how to effectively and in real-time tackle arbitrary sensor drifts. In our work, we elucidate the use of a simple theory called SuDA to define all sensor data drifts as support drifts. This allows SuDA to effectively handle arbitrary drifts, achieving results comparable to supervised learning. Additionally, we extend SuDA for real-time applications, enabling quick transfer learning within an extremely short timeframe (e.g., the duration of a single elbow flexion).

% Along with such exploration, it has been noted that flexible sensors face critical domain-mismatch challenges across different users, motion types, and wearing sessions, where the trained models fail to generalize~\cite{chen2023dispad,kang2022dual}. 
\yq{However, all these methods face critical data scarcity challenges, requiring collection across different users, motion types, and wearing sessions~\cite{chen2023dispad,kang2022dual}. This typically involves MoCap studios, specialized equipment, and significant manual labor, making large-scale data acquisition difficult and expensive.}
% \jiawei{Along with such exploration, it has been noted that flexible sensors face critical data hungry challenges, where data across different users, motion types, and wearing sessions~\cite{chen2023dispad,kang2022dual} typically need to be collected in MoCap studios with specialized equipment and substantial manual labor, making them difficult and expensive to obtain at scale. }

\subsection{Sim2Real Paradigm}
The Sim2Real paradigm trains models on simulated data and then applies them to solve real-world problems, thus successfully alleviating the data scarcity problem in machine learning~\cite{hofer2021sim2real,nikolenko2021synthetic}. 
%Numerous areas applied sim2real with varying success.
Researchers use this approach to develop cost-effective solutions for MoCap~\cite{huang2018deep, yi2021transpose, yi2022physical}, pose estimation~\cite{lin2021end}, pedestrian detection~\cite{fabbri2021motsynth,stauner2022synpeds}, action recognition~\cite{winkler2022questsim}, autonomous driving~\cite{muller2018driving, zhou2019does} and robotics~\cite{abeyruwan2023sim2real,kataoka2023bi,wu2023sim2real}. These works use generative models that make simulation look more like reality ~\cite{bousmalis2017unsupervised}, domain adaptation to mitigate domain discrepancy ~\cite{doersch2019sim2real}, or randomize the simulator to increase the distribution overlap ~\cite{horvath2022object}.

% T relevant example to our work involves IMU-based Mocap. In these studies, \textit{virtual} inertial sensors are placed on the SMPL mesh surface to generate a substantial amount of synthetic data. Although effective, real data still be needed to fine-tune the synthetic pre-trained model.

% Subsequently, open-source motion libraries (e.g., AMASS, DIP-IMU) are employed to record simulated velocity and acceleration, generating a substantial amount of synthetic data. 
\yq{To our knowledge, \citet{park2020sim} is the only work that adopted a Sim2Real approach for flexible sensor based MoCap, which simulates the interplay between flexible capacitive sensors and human motions using OpenSim \cite{delp2007opensim}. 
However, unlike their use of a distribution-based method to fine-tune models pre-trained with synthetic data using real {\it labeled} data, our SuDA achieves a high accuracy comparable to supervised learning without the need for any real labeled data and is therefore more cost-effective.
}
% \jiawei{The most relevant example to our work is \cite{park2020sim}, which simulates the relationship between flexible capacitive sensors and human motions using OpenSim \cite{delp2007opensim}. The biggest difference with our work is that they adopt a distribution-based paradigm to fine-tune the synthetic pre-trained model using real labeled data yet we use a novel \textit{Support-based }Domain Adaptation (SuDA) approach, hence achieving comparable accuracy to supervised learning without any real labeled data.
% }

% Our work explores a novel Domain Adaptation method to bridge the domain gaps between simulated and real data, \jiawei{which effectively addresses the data hungry problem}.

% and calculate the geodesic distance on the surface to simulate sensor readings. 

\subsection{Domain Adaptation}
Domain adaptation (DA) aims to mitigate the gap between the source and target domains so that models trained in the source domain(s) can be applied to the target domain(s).
% Domain adaptation has achieved a resounding success to learn the invariant representations by leveraging from a labeled data source domain to adapt to an unlabeled data target domain. 
\yq{To date, distribution-based domain adaptation (\textit{DiDA}) has dominated DA, which aims to reduce the distribution discrepancy between two domains \cite{pan2009survey}.}
% \jiawei{ The most profound techinique is distribution-based domain adaptation (\textit{DiDA}), which aims to to reduce the distribution discrepancy between the two domains\cite{pan2009survey}.}
% These methods usually perform DA by either reweighting samples from the source domain \cite{chen2019cross}, or seeking an explicit feature transformation that transforms the source and target samples into the same feature spaces \cite{wang2018stratified, wang2018visual}.
With advent of deep learning, pioneer works showed that deep neural networks could learn more transferable features for domain adaptation~\cite{yosinski2014transferable}.
Since then, various solutions have been proposed to align the distributions (or their statistics) of features extracted by deep neural networks in the source and target domains.
For example, Deep Domain Confusion (DDC)~\cite{tzeng2014deep} first proposed the use of Maximum Mean Discrepancy (MMD) loss to align the feature distributions; Deep Adaptation Networks (DAN) \cite{gretton2012optimal} extends the idea to the use of multiple-kernel MMD; Deep CORAL \cite{sun2016deep} proposed CORAL loss to align statistics of feature distributions for domain adaptation. This distribution-based formulation has become a ``standard'' for state-of-the-art domain adaptation solutions~\cite{ganin2015unsupervised,yu2019transfer,deng2021adversarial,li2022selective,lee2022confidence,garg2023rlsbench}.

\yq{Despite their effectiveness, DiDA methods struggle when there is a significant distribution disparity between the two domains
(\textit{a.k.a., Negative Transfer}), which is a life-long and challenging issue in DA~\cite{rosenstein2005transfer, wang2019characterizing, zhang2022survey}. Addressing this challenge, we propose a novel {\it support-based} domain adaption (SuDA) paradigm that aligns the supports rather than distributions between two domains. Thanks to its independence from distribution disparity, SuDA produces more accurate and generalizable results in our motion capture task.
}
% \jiawei{Unfortunately, the effectiveness of DiDA is not always guaranteed when there is a significant distribution disparity between the two domains
% (i.e.,\textit{ Negative Transfer}), which is a life-long and chanllanging issue in DA\cite{rosenstein2005transfer, wang2019characterizing, zhang2022survey}. With regard to the disadvantages of distribution-based solutions, we propose a novel {\it support-based} domain adaption method (SuDA) that aligns the supports of predictive functions instead of distributions between two domains. As independent of distribution variations, SuDA produces more accurate and generalizable results in our motion capture task. }

% , and directly optimizing the distributions with DiDA often fails to yield satisfactory results.  

\section{Real and Simulation Data Preparation}
\label{sec:hardware_simulation}

% \subsection{Real Data Collection}
\subsection{Real Data with Flexible Sensors}
\label{sec:hardware}

\begin{figure}[H]
   \centering
     % \begin{subfigure}[b]{0.31\linewidth}
     %     \centering
     %     \includegraphics[width=\linewidth]{CameraReady/LaTeX/Fig_Extension.png}

     %     \caption{Sensor Linearity}
     %     \label{fig:fcase 1}
     % \end{subfigure}
     \begin{subfigure}[b]{0.43\linewidth}
         \centering
         \includegraphics[width=\linewidth]{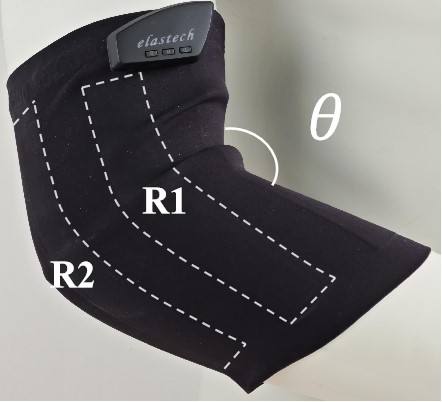}
     \end{subfigure}
     \begin{subfigure}[b]{0.55\linewidth}
         \centering
         \includegraphics[width=\linewidth]{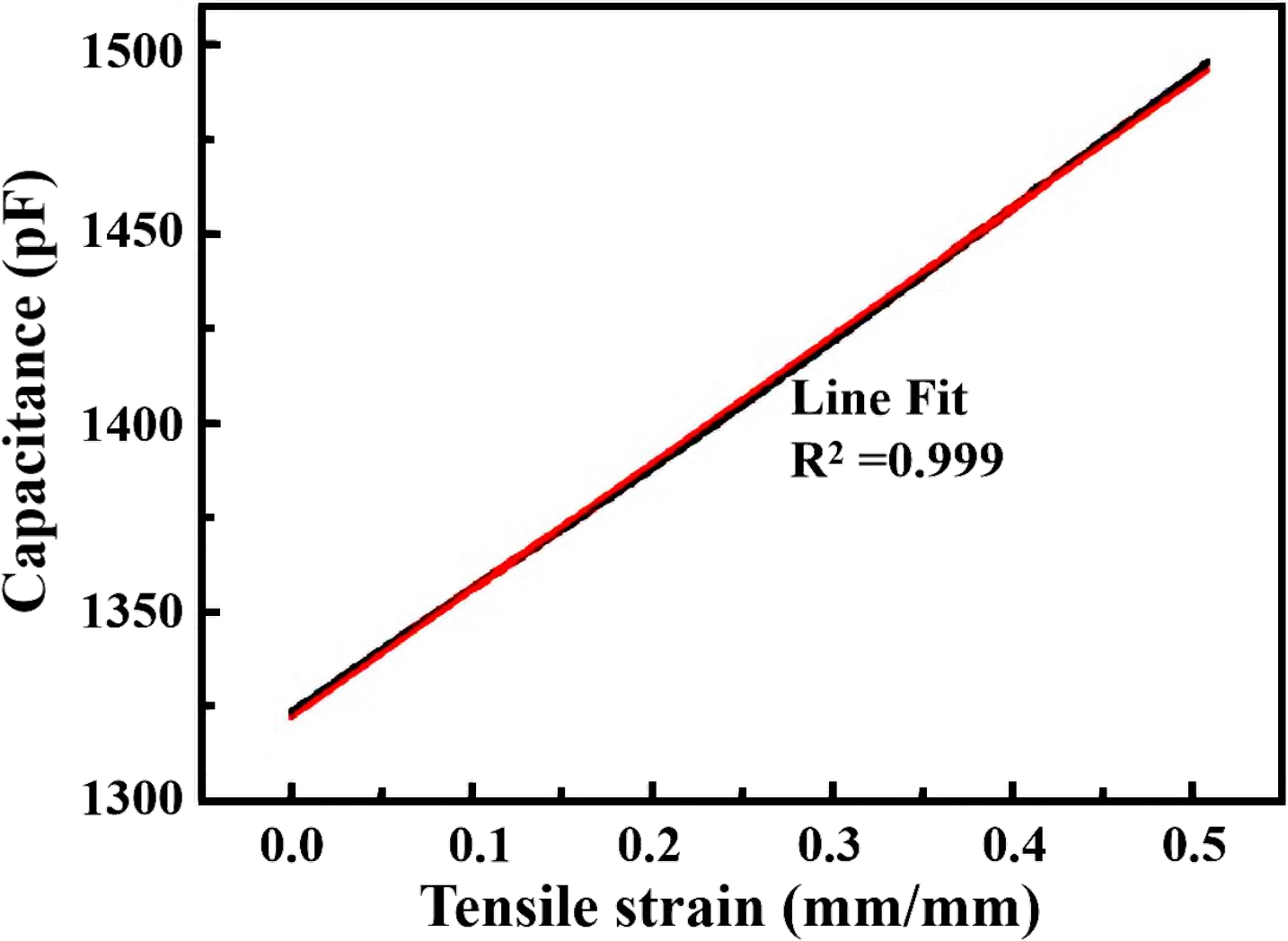}
                 \vspace{-15pt}
         \label{fig:fcase 2}
     \end{subfigure}
        \vspace{-10pt}
     \caption{\yq{(a) Our hardware. $R_1$ and $R_2$ denote the two sensors on our smart pad, and $\theta$ denotes the joint angle. (b) High linearity characteristics of capacitive strain sensors.}}
     % \caption{ (a): Our hardware. $R_1$ and $R_2$ are our two sensor readings and $\theta$ is the estimated joint angle. (b): High Linearity characters of capacitive strain sensors.}
     \label{fig:hardware_and_linearity}
\end{figure}

We develop a prototype by augmenting a standard smart pad with two capacitive strain sensors with high-linearity placed on the elbow's olecranon side. Our method aims to estimate the bending angle, $\theta$, of human joints, denoted as $y_t$, from the two sensor readings, denoted as \yq{$x_t=\{R_1, R_2\}$ (Fig.~\ref{fig:hardware_and_linearity})}. The high-linearity nature of capacitive strain sensors guarantees their capacitance (\textit{i.e.}, sensor readings) increases with the stretch caused by \yq{joint bending}
% the bending of the joint 
\cite{atalay2017batch, totaro2017soft, white2017low, wang2017balanced}. Please see Appendix A about our hardware.
    \vspace{-10pt}
     \vspace{-15pt}

% The bending angle is defined as the angle in the sagittal plane between the humerus and the central line between the radius and the ulna. 

 % The sensor readings are digitized to values in the range [0, 1023] and transmitted wirelessly via Bluetooth Low Energy at a frame rate of 50Hz. The collected sensor reading is $x$ in the target domain. 
 % \jiawei{More evidence can be seen in the Appendix}.

%More details concerning the hardware prototype can be found in the supplementary materials.

%\subsection{Overview}
%Our approach can be divided into two main components. The first part involves generating virtual angle-sensor paired data using a large-scale open-source motion library and commercial software. In the second part, we focus on presenting a domain support registration-based transfer learning method, which enhances the realism of our virtual data. This combination allows for a more authentic representation of the real-world scenarios in our study.

% \subsection{Simulation Data Generation}
\subsection{Simulation Data with Body-Fabric-Sensor Model}
The ultimate goal of our simulation approach is to mimic characteristics of the capacitive strain sensors.  

% We could obtain the equivalence of support gradients between simulation and real data with our simulation approach.

To achieve this, we here construct a body-fabric-sensor model to generate the paired sample between the simulated joint rotation and sensor reading\xiaowei{s}. We used the SMPL model to create fbx files of simulated human 3D models.
% Note that the skeleton model discussed in the following steps is a standard human skeleton obtained via MotionBuilder binding, with 68 joints instead of the 24 joints native to SMPL.
The entire process of synthesizing digital mannequins through SMPL models can be divided into \jiawei{four} major stages as follows.
%To address the effect of body sizes on sensor parameters, we used the SMPL model to create human 3D models of different body sizes. 
% To make the article more logical, 
% We briefly state the main components in the process of the simulation framework. For the detailed operations, please refer to the supplementary materials.
\subsubsection{Body Modeling} We employ the Skinned Multi-Person Linear (SMPL) \cite{loper2023smpl} model to simulate human body and movement. 
It comprises the following four steps.
% Skinned Multi-Person Linear (SMPL) ~\cite{loper2023smpl} was proposed to create realistic animations of different human bodies and motions. %We adjusted the shape parameter of SMPL to obtain 3D human bodies of different sizes. %While the pose parameter of SMPL has 24 joints, the standard rigged data of the real human body contains 68 joints. Thus, we used Autodesk Motion Builder to set the standard skeleton for the comparison effect.
% The process of synthesizing digital mannequins through SMPL models can be divided into three stages. 
\begin{itemize}
    \item \emph{Template construction}: 
    we \yq{first} defined a mean template as the foundational pose of the human body \yq{and then} represent each human pose using parameters that describe its offsets from the fundamental pose.
    % we defined a mean template to serve as the foundational pose of the human body. From there, we represent each human pose using parameters that describe its offsets from the fundamental pose. 
    % This procedure is linear and does not consider the interdependence of pose parameters.
    % we established a mean template as the basic pose of the human body. Then, we described the pose of the human body and the offset of the basic pose through parameters to form the final desired human pose. This process is linear and does not consider the influence of pose parameters.
    \item \emph{Shape variation}: we specified the shape of the simulated human body using the SMPL shape parameter and obtained a mesh with specific height and weight. 
    % The pose parameter also affects the shape of the mesh to some extent.
    \item \emph{Mesh skinning}: 
    % we performed skinning to build the actual mesh according to the movement of the skeleton joints.
    % we performed skinning to account for the movement of the skeleton joints. Skinning is a weighted linear combination of skin nodes with the change of skeletal joints.
    We only calculated the mesh in the static pose in the previous two stages. When human skeleton joints move, the ``skin'' composed of vertexes will change with the movement of the skeleton joints; this process is called skinning. 
    % The skinning process can be considered a weighted linear combination of skin nodes with changes in skeletal joints. 
    % \xiaowei{Specifically,} the closer an endpoint is to a specific bone joint, the stronger the effect of changes, such as rotation and translation, with that bone joint.
    \item \emph{Motion simulation and binding}: 
    % To address the effect of different motions on sensor parameters, we collected motion capture data from Mixamo and used the online binding function to bind the motions to the 3D human model generated based on SMPL.
    To address the effect of different motions, we collected approximately 30,000 frames of MoCap data from the free online animation library Mixamo containing eleven expressive motions (e.g., \textit{Chincken Dance, Robot Dance}).
    % Motion samples can be found in Figure. 
    % Furthermore, we used the online binding function to bind the motions to the 3D human model generated based on SMPL.
    % \item \emph{Motion} 
    % We simulate real-world human motions with the data from Mixamo~\cite{Mixamo} and use its online binding function to bind the motions to the created SMPL model.
\end{itemize}

% \begin{figure*}[ht]
% \centering
% \includegraphics[width=1.0\linewidth]{CameraReady/LaTeX/Fig_SimMotion.png}
% \caption{Sample Data from Simulation Motion}
% \label{fig:SimMotion}
% \end{figure*}

\subsubsection{Fabric Deformation}
% The creases produced by the flexible sensors when worn and the differences in wearing position are also sources of Sim2Real domain gaps.
To reduce the Sim2Real domain gaps arising from creases of real fabric, we used the cloth database built into the clothing design software Marvelous Designer. Considering the basic parameters such as elasticity, friction, and thickness of real fabric pads and visual comparison inspection of simulation results, we selected hard-jeanet from the database. 

% The simulation can produce similar wrinkles, pulling, and other phenomena as real motion scenes.
% We simulate the wearing of our elbow pad by creating a virtual fabric model using the professional software of Marvelous Designer and Blender, which can precisely simulate the different wearing positions and fabric creases during movement.
%Note that defects like tearing and threading may occur when running the fabric simulation software. 
%So it is occasionally necessary to manually fine-tune the fabric positions in the software.

% To simulate the creases produced by flexible sensors when worn and the differences in wearing position, we built a virtual fabric model using Marvelous Designer and Blender. With the marvelous designer console, we can bind simulated fabrics to 3D mannequins and simulate different wearing positions by controlling the position of the fabric. In addition, due to defects such as tearing, threading, etc. that may occur when the fabric simulation software is run, it is occasionally necessary to manually fine-tune the position of the fabric (details in appendix).
%  We calculated the length change of the sensor using Python scripts based on the 3D modeling and fabric simulation.

\subsubsection{Sensor Stretching}
\xiaowei{The} flexible sensors we used exhibit high linearity and repeatability ($R^2 = 0.999$) with low hysteresis (\yq{Fig.~\ref{fig:hardware_and_linearity}}). \xiaowei{We} can approximate its \xiaowei{readings} by its stretched length.
Specifically, we compute the changes in the geodesic distance between sensor endpoints in the elbow pad in different frames to measure sensor stretching and its reading\xiaowei{s}. The simulated sensor readings are represented as $x_s$. 

\subsubsection{Elbow Bending Angle}
We calculate the elbow angle of the SMPL model created above using the 3D coordinates of its elbow joint in different frames.
This is the $y_s$ in the source domain.
% Our task is to estimate the bending angle of the elbow joint based on two flexible sensors surrounding the arm. We calculate the elbow angle on the SMPL model. 
% We preprocessed the motion data to obtain the 3D coordinate position information of each frame of motion joint and used it to calculate the elbow bending angle.

% \begin{figure}[ht]
% \centering
% \includegraphics[width=0.8\linewidth]{CameraReady/LaTeX/Fig_Wearing position.pdf}
% \caption{The wearing postion of elbow pad}
% \label{fig:hardware}
% \setlength{\abovedisplayskip}{0pt}
% \setlength{\belowdisplayskip}{0pt}
% \end{figure}

% To simulate sensor readings, we used the geodesic distance on the corresponding wearing postion of the fabric and calculated the geodesic distance for the simulated fabric using a Dijkstra method. The soft sensors exhibited high linearity and repeatability with low hysteresis.

%\begin{figure}[ht]
%\centering
%\includegraphics[width=0.8\linewidth]{CameraReady/LaTeX/Fig_Extension.png}
%\caption{The theoretical curve between sensor readings and their extension}
%\label{fig:framework}
%\end{figure}

% Please refer to the supplement for more details about the hardware for real data collection and the process for simulation data generation.

\begin{figure*}[ht]
\centering
\includegraphics[width=1.0\linewidth]{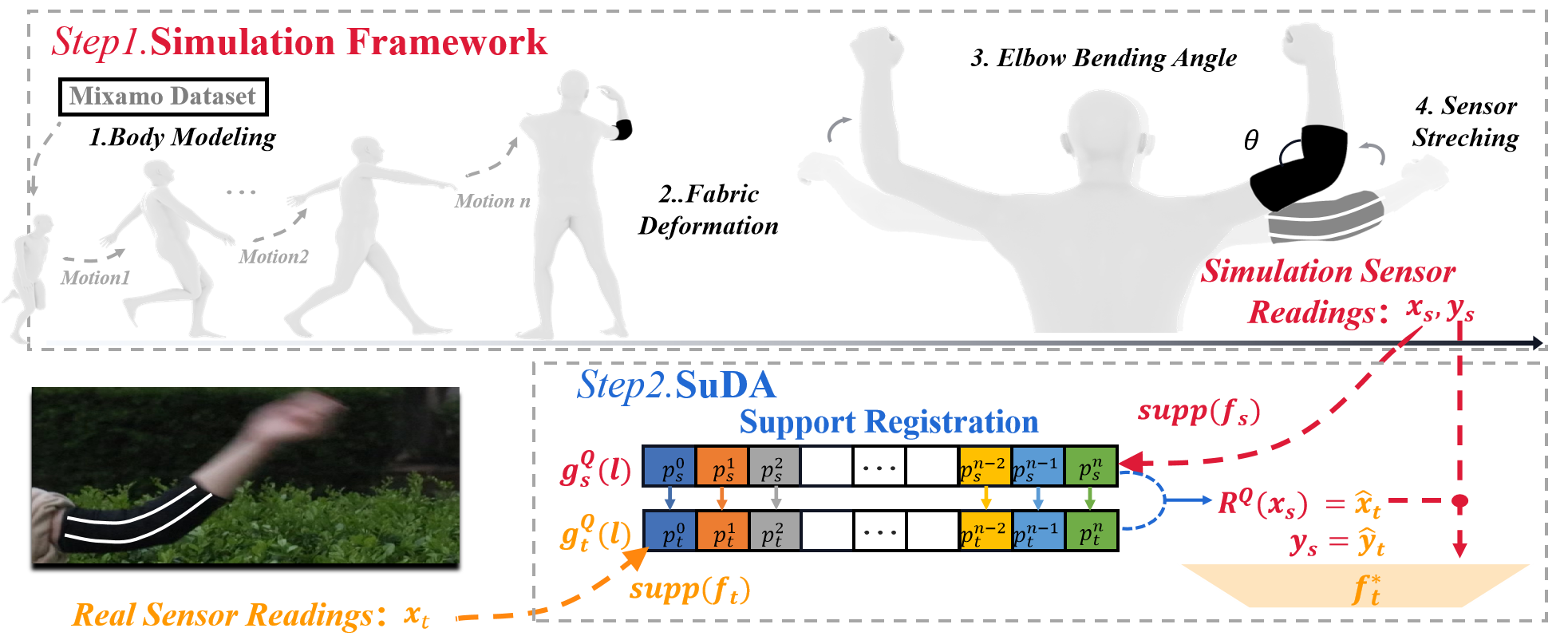}

\caption{Pipeline Overview. Top row: the generation of simulation data with our Body-Fabric-Sensor model, which consists of four main parts: {\it 1. Body Modeling 2. Fabric Deformation 3. Elbow Bending Angle 4. Sensor Streching.} Bottom row: the proposed {\it support-based} domain adaptation method (\textbf{SuDA}). SuDA first parameterizes the supports of source and target domain, denoted as {\textcolor[RGB]{222,26,55}{\textbf{$supp(f_s)$}}} and {\textcolor[RGB]{255,152,0}{\textbf{$supp(f_t)$}}}, into $n+1$ evenly-distributed proxy points respectively, denoted as {\textcolor[RGB]{222,26,55}{\textbf{$g_s^Q(l)$}}} and {\textcolor[RGB]{255,152,0}{\textbf{$g_t^Q(l)$}}}. Then, it applies a novel Support Registration $R^Q$ to map {\textcolor[RGB]{222,26,55}{\textbf{$x_s$}}} to its nearest {\textcolor[RGB]{255,152,0}{\textbf{$p_s^i$}}}. Finally, we train the predictive function {\textcolor[RGB]{255,152,0}{$f^*_t$}} on the target domain using ({\textcolor[RGB]{255,152,0}{\textbf{$\hat{x}_t$}}}, {\textcolor[RGB]{255,152,0}{\textbf{$\hat{y}_t=$}}}{\textcolor[RGB]{222,26,55}{\textbf{$y_s$}}}) and finally to {\textcolor[RGB]{255,152,0}{\textbf{$\hat{x}_t$}}}.}

\label{fig:framework}
\end{figure*}

\section{Support-based Domain Adaptation (SuDA)}
The overall objective of SuDA is to minimize the \yq{support} divergence between the two domains.
\subsection{Preliminaries}
% \paragraph{Function Support}
% \paragraph{Preliminaries}
\paragraph{Definition of Support} For a real-valued function $f: \textit{X} \mapsto \mathbb{R}$ defined on a domain $\textit{X}$, we denote the \textbf{set-theoretic support} of $f$ as $supp(f)$, {\it i.e.}, the set of points in $\textit{X}$ where $\textit{f}$ is non-zero:
\begin{equation}
supp(f) = \{x \in \textit{X}: f(x) \neq 0 \}
\end{equation}

% For flexible sensors, we define the range of support as the path of sensor readings.

% \paragraph{Unsupervised Domain Adaptation}
% Unsupervised  Domain Adaptation seeks to learn a predictive function $f_t: x_t \mapsto y_t$ on the target domain $D_t$ using labeled data from a source domain $D_s(x) = \{x_i, y_i\}_{i=1}^{N_s}$, while minimizing prediction error on the target domain~\cite{tlbook}:
% \begin{equation}
% f^*=\mathop{\arg\min}\limits_{f_s \in \mathscr{H} }\mathbb{E} _{(x,y)\in D_s}\mathbf{L}(f_s(x),y) + \lambda R((D_s), (D_t)),
% \end{equation}
% where $f_s$ is the predictive function on source domain and $R(\cdot ,\cdot )$ is the \textbf{Transfer Regularization Term}.

% \paragraph{Problem Definition}
% Support-based Domain Adaptation aims to learn predictive function by apply $R(\cdot ,\cdot)$ Term with $R(supp(f_s), supp(f_t))$:
% \begin{equation}
% f^*=\mathop{\arg\min}\limits_{f_s\in \mathscr{H}}\mathbb{E} _{(x,y)\in D_s}\mathbf{L}(f_s(x),y) + \lambda R(supp(f_s), supp(f_t))
% \end{equation}

\begin{definition}[SuDA]
Let $\mathcal{D}_s = \{(x_s, y_s) \in 
(X_s \times Y_s)\}_{s=1}^{N_s}$ be the source domain, $\mathcal{D}_t = \{x_t \in X_t\}_{t=1}^{N_t}$ be the unlabeled target domain, and let $f_s$ and $f_t$ be the motion prediction functions defined on $X_s$ and $X_t$ respectively, SuDA aims to learn an optimized $f_t^*$ by minimizing a given loss function $L$ ({\it e.g.}, MAE) computed between the prediction $f_t(\hat{x}_t)$, where $\hat{x}_t$ is estimated from $x_s$, and its corresponding label $\hat{y}_t=y_s$:
\begin{equation}
\begin{split}
f_t^*&=\mathop{\arg\min}\limits_{f_t}\mathbb{E}_{\mathcal{D}_s} L(f_t(\hat{x}_t),\hat{y}_t)\\
&=\mathop{\arg\min}\limits_{f_t}\mathbb{E}_{\mathcal{D}_s} L(f_t(R(x_s)),y_s)
\end{split}
\label{eq:SuDA_definition}
\end{equation}
where the {\bf Support Registration} function $R$ is defined as:
\begin{equation}
    \mathop{R}: supp(f_s) \mapsto  supp(f_t)
\end{equation}
\label{def:SuDA}
\end{definition}

% \paragraph{Equivalence of Support Gradient}
% The equivalence of support gradient $\frac{df_s}{dx_s}$ and $\frac{df_t}{dx_t} $ comes from the characteristics of the flexible capacitive sensors \cite{atalay2017batch, totaro2017soft, white2017low,wang2022low}. We can easily mimic this relationship through simulation.
\vspace{-10pt}
\subsection{The Rationale Underpinning SuDA}
The effectiveness of SuDA hinges on the assumption that $supp(f_s)$ and $supp(f_t)$ \yq{are bijective ({\it i.e.}, one-to-one correspondence),}
% can be registered properly, 
{\it i.e.}, $\forall x_s^a, x_s^b \in supp(f_s)$, $\forall x_t^a, x_t^b \in supp(f_t)$:
\begin{equation}
\begin{split}
x_s^a \neq x_t^a&,~x_s^b \neq x_t^b
\\
        y^b - y^a = \int_{x^a_s}^{x^b_s} \frac{df_s}{dx_s} dx_s &= \int_{x^a_t}^{x^b_t} \frac{df_t}{dx_t} dx_t
        % y^b = \int_{x^a_s}^{x^b_s} f_s'(x_s) dx_s + C_s = \int_{x^a_t}^{x^b_t}f_t'(x_t) dx_t + C_t
        % \\
% \Leftrightarrow 
    % \int_{supp(f_s)}\frac{dy}{dx_s} dx_s + C_s = \int_{supp(f_t)}\frac{dy}{dx_t} dx_t + C_t
\end{split}
\label{eq:rationale_SuDA}
\end{equation}
Eq.~\ref{eq:rationale_SuDA} guarantees the the correctness of SuDA, which ensures that $y^a$ and $y^b$ are the labels corresponding to $x_s^a, x_t^a$ and $x_s^b, x_t^b$ respectively through Support Registration:
\begin{equation}
    \begin{split}
        R: x^a_s \mapsto x^a_t, x^b_s \mapsto x^b_t
    \end{split}
    % \label{eq:7}
\end{equation}
Consequently, the success of SuDA is contingent upon the effectiveness of the Support Registration function $R$. 
% For our task, although $supp(f_s)$ and $supp(f_t)$ look different, {\it i.e.}, geodesic distances on simulated motion meshes {\it vs.} flexible sensor readings, both of them correspond to the same set of elbow angles.

% Our objective is to iteratively improve $f_s$ so that it approaches $f_t$, but directly optimizing the distance between predictive functions is infeasible due to the absence of the expression for $f_t$. As directly optimizing $f_t$ is challenging due to the absence of target domain labels, while it's easier to optimize $f_s$ as we have access to source domain data and labels :
% \begin{equation}
% f^*=\mathop{\arg\min}\limits_{f_s}\mathbb{E} _{(x,y)\in D_s}\mathbf{L}(f_s(x),y)
% \end{equation}
% Since both $supp(f_s)$ and $supp(f_t)$ are known, optimizing $R(supp(f_s), supp(f_t))$is more effective and manageable. Furthermore, given that $\frac{df_s(x)}{dx_s} \approx \frac{df_t(x)}{dx_t}$, it follows that 
% \begin{equation}
% \min R(supp(f_s),supp(f_t)) \Leftrightarrow \min R(f_s,f_t),
% \end{equation}
% which provides a simplified approach to achieve the same optimization goal.

% To minimize the distance between the supports of domain prediction functions, denoted as $R(supp(f_s),supp(f_t))$, we establish a correspondence between each element $x_{si}$ in $supp(f_s)$ and a corresponding element $x_{tj}$ in $supp(f_t)$ using a transformation function $T: x_{si} \to x_{tj}$，which ensures function support are matched together.

 % \vspace{8mm}
\subsection{Support Registration in Flexible Sensor-Based MoCap}
In our task, $y$ denotes joint angles, $supp(f_s)$ and $supp(f_t)$ denote geodesic distances on simulated motion meshes and flexible sensor readings, respectively. 
The overall object of Support Registration is to map $x_s$ and $x_t$ with the same label (\textit{i.e.,} $y_s = y_t$) properly. 

The intrinsic degree of freedom of the elbow joint $Dim=1$, therefore we parameterize the curves of $supp(f_s)$ and $supp(f_t)$ with parameter $l \in [0,1]$ using geometric curve fitting and have:
% \vspace{-10mm}
\begin{equation}
\begin{split}
    supp(f_s) &\leftrightarrow g_s(l) \\
    supp(f_t) &\leftrightarrow g_t(l)  
    \vspace{-10mm}
\end{split}
\end{equation}
where $g_s(l)$ and $g_t(l)$ denote the  parametric equations of $supp(f_s)$ and $supp(f_t)$, respectively.
After parameterization, we find data with equal parameter change (\textit{i.e.,} $l - 0$) have the same label change (\textit{i.e.,} $y^b - y^a$):
\begin{equation}
\begin{split}
        \int_{x^a_s}^{x^b_s} \frac{df_s}{dx_s} dx_s &= \int_{x^a_t}^{x^b_t} \frac{df_t}{dx_t} dx_t =\int_{0}^{l} \frac{df_l}{dl} dl
\end{split}
\label{Eq:SuDA_param_registration}
\end{equation}

which ensures the success of SuDA in our flexible sensor-based
MoCap task. Please see evidence of Eq. \ref{Eq:SuDA_param_registration} in Appendix C.

To find a proper mapping between corresponding $x_s$ and $x_t$ with the same label, we can define the Support Registration function $R$ for flexible sensor-based motion capture as:
\begin{equation}
R: x_s \mapsto l \mapsto g_t(l) = \hat{x}_t
\end{equation}

Then we simplify subsequence calculations by quantizing each of the support curves into $n+1$ evenly-distributed proxy points:
\begin{equation}
\begin{split}
    g^Q_s(l): &\{p_s^0, p_s^1, ..., p_s^{n-1}, p_s^n\},~\mathrm{where}~p_s^i = g_s(l*i/n)\\
    g^Q_t(l): &\{p_t^0, p_t^1, ..., p_t^{n-1}, p_t^n\},~\mathrm{where}~p_t^i = g_t(l*i/n)
\end{split}
\label{eq:support_quantization}
\end{equation}
This allows us to define the quantized version of our Support Registration function, denoted as $R^Q$, by mapping $x_s$ to its nearest $p_s^i$: 
\begin{equation}
    R^Q: x_s \mapsto p_s^i \mapsto i \mapsto p_t^i = \hat{x}_t
    \label{eq:R_Q}
\end{equation}
% \vspace{-5mm}

We use $R^Q$ in all our experiments. Alg.~\ref{alg:algorithm} shows the pseudo-code of our SuDA method. 
\begin{figure}[H]
   \centering
     % \begin{subfigure}[b]{0.31\linewidth}
     %     \centering
     %     \includegraphics[width=\linewidth]{CameraReady/LaTeX/Fig_Extension.png}
     %        \vspace{-15pt}
     %     \caption{Sensor Linearity}
     %     \label{fig:fcase 1}
     % \end{subfigure}
     \begin{subfigure}[b]{0.49\linewidth}
         \centering
         \includegraphics[width=\linewidth]{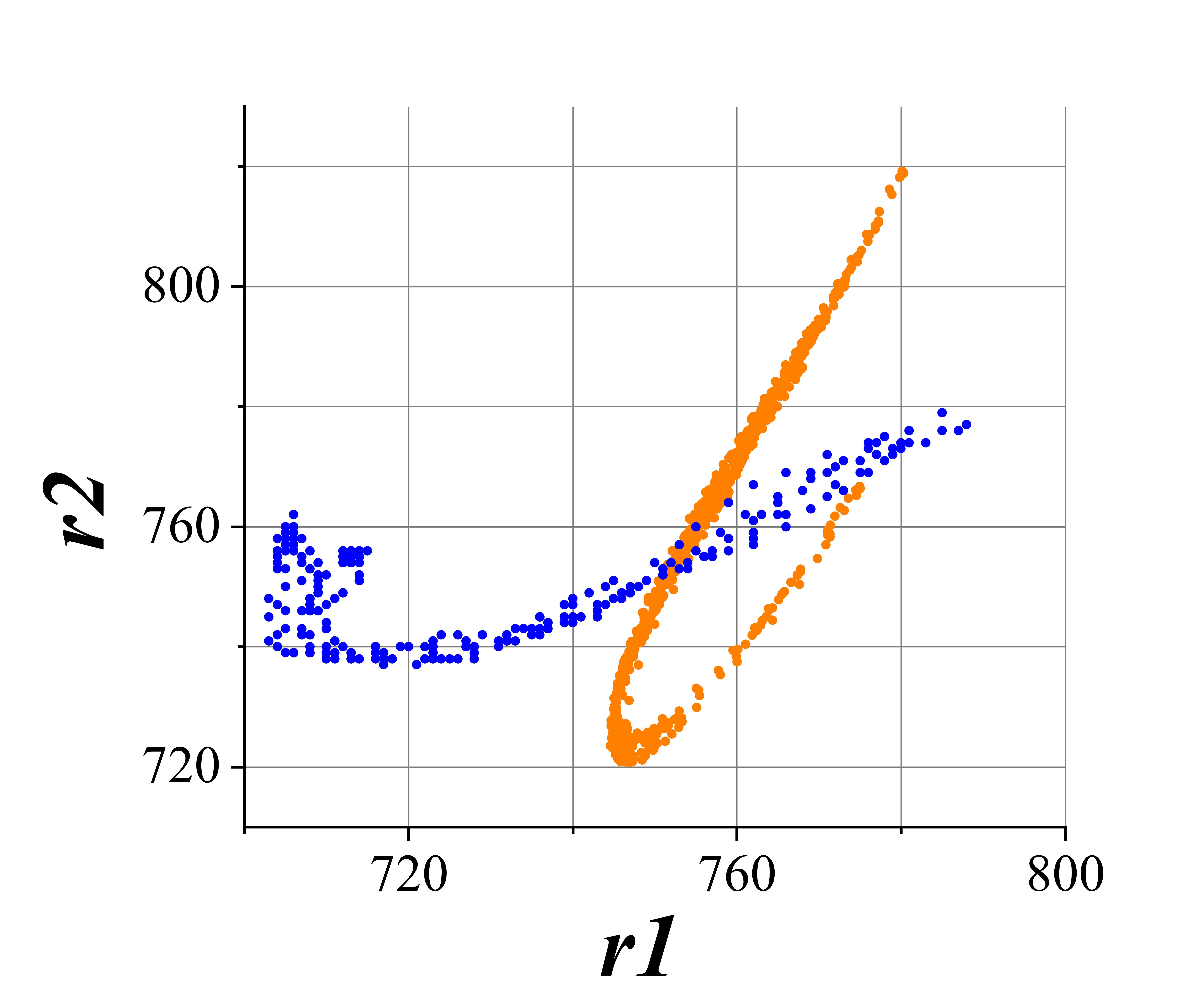}
                 \vspace{-15pt}
         \caption{Before SuDA}
         \label{fig:fcase 2}
     \end{subfigure}
     % \vspace{10pt}
     \begin{subfigure}[b]{0.49\linewidth}
         \centering
         \includegraphics[width=\linewidth]{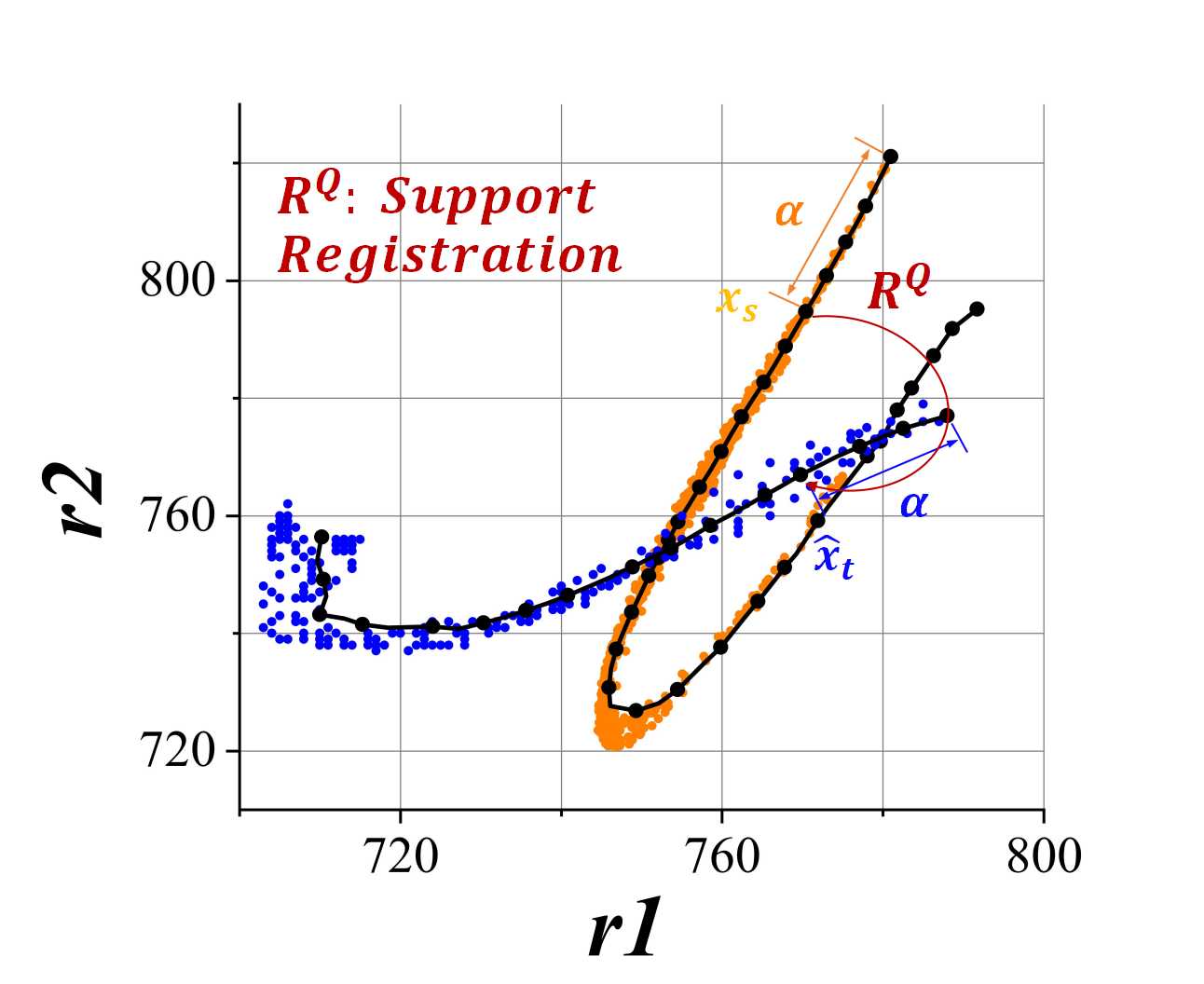}
                 \vspace{-15pt}
         \caption{After SuDA}
         \label{fig:fcase 2}
     \end{subfigure}
        \vspace{-10pt}
     \caption{  \textcolor{orange}{\textbf{Orange}} and \textcolor{blue}{\textbf{Blue}}: source and target data samples. \textbf{Black}: parameterized and quantized supports.  (a) Two scatter curve tracks are our data samples and represent function support of the two domains. (b) Then, we use curve parameterization to quantize function support into several segments. Since $\frac{df_s}{dx_s} \approx \frac{df_t}{dx_t} $, points with the same parameters have the same labels and can be registered together by $R^Q$.}
     \label{fig:suda}
\end{figure}
     \vspace{-15pt}
     
\begin{algorithm}[t]
    \caption{Support Based Domain Adaptation (SuDA)}
    \label{alg:algorithm}
    % \algsetup{linenosize=\scriptsize}
    \renewcommand{\algorithmicrequire}{\textbf{Input:}}
	\renewcommand{\algorithmicensure}{\textbf{Output:}}
    % \textbf{Input}: ;  \\
    % \textbf{Parameter}: 
    % Transfomation Matrix $T_{(3n\times2)}$; \\
    % Learning rate $\eta$; \\
    % \textbf{Output}:        
    \begin{algorithmic}[1] %[1] enables line numbers
    % \STATE Let $iter=0$.
        \REQUIRE{Source domain data $\mathcal{D}_s = \{(x_s, y_s)\}_{s=1}^{N_s}$, target domain data $\mathcal{D}_t = \{x_t \in X_t\}_{t=1}^{N_t}$}, loss function $L$
        \ENSURE{Prediction function $f^*_t$ on the target domain};
        \STATE Parameterize $supp(f_s)$ and $supp(f_t)$ with parameter $l \in [0,1]$ using $\{x_s\}^{N_s}_{s=1}$ and $\{x_t\}^{N_t}_{t=1}$, respectively;
        \STATE Quantize the parameterized supports (Eq.~\ref{eq:support_quantization})
        \STATE Conduct (quantized) support registration $R^Q$ to map $x_s$ to $\hat{x}_t$ with $\hat{y}_t = y_s$ (Eq.~\ref{eq:R_Q});
        \STATE Optimize $f_t^*=\mathop{\arg\min}\limits_{f_t}\mathbb{E}_{\mathcal{D}_s} L(f_t(\hat{x_t}),\hat{y_t})$ (Eq.~\ref{eq:SuDA_definition}) using gradient descent ({\it e.g.}, SGD);
        \STATE \textbf{Return}  $f^*_t$
    \end{algorithmic}
\end{algorithm}
\vspace{-5mm}
\section{Experimental Results}

\subsection{Experimental Setup}
\subsubsection{Implementation Details}
Following common practice, we implement the predictive function in the target domain ($f^*_t$ in Step 4 of Alg.~\ref{alg:algorithm}) with a neural network consisting of LSTM and MLP layers and train it using an SGD optimizer with a learning rate of $1e^{-3}$ and a weight decay of $5e^{-4}$, momentum= $0.9$, batch size$=32$, and training epoch $e=50$. We use a LRscheduler with $\gamma=0.0003$ and decay = $0.8$.
We employ an MAE loss as the $L$ used in Alg.~\ref{alg:algorithm}:
$\mathcal{L}_{mae}=\sum_{i=1}^{n}|y_i-f(x_i)| / n$.

To make a fair comparison, we use the same network architecture, optimization strategy, and loss function when training our Sim2Real method and the two baselines: supervised learning and Real2Real.
All experiments were conducted on a desktop PC with an Intel i7-12700KF CPU and an NVIDIA RTX 3090 GPU. \xiaowei{T}he appendix \xiaowei{includes} details on network architectures, source code (including implementations of methods in the following comparison experiment), and the dataset used in our experiments.

\subsubsection{Real Dataset $\mathcal{D}_{t^*}$}
After obtaining the ethical approval, we collected our real dataset $\mathcal{D}^{(id, m, wp)}_{t^*}$ consisting of paired motion and sensor reading\xiaowei{s}, where $id$ denotes user ID, $m$ denotes motion, and $wp$ denotes wearing position, as follows.
Specifically, we recruited 11 participants ($id = 1, 2, ...,11$) and asked them to perform four basic motions ($m=run, walk, jump, clap$) at 12 wearing positions ($wp = 30^\circ, 60^\circ, 90^\circ, ..., 360^\circ$), and collected 371,661 frames of data in total.
For each ${(id, m, wp)}$ instance, we split its data into a training set $\mathcal{D}^{(id, m, wp)}_{t, train}$ and a test set $\mathcal{D}^{(id, m, wp)}_{t^*, test}$ by $0.7:0.3$, where we remove the motion labels in the training set to satisfy the unsupervised domain adaptation conditions in real-world application scenarios.
Note that we use $\mathcal{D}_{t^*}$ rather than $\mathcal{D}_{t}$ as we need to collect ground truth motion labels to evaluate our method.

We report the mean and standard deviation of the test accuracy across all $\mathcal{D}^{(id, m, wp)}_{t^*, test}$ in a set of instances specified by different $(id, m, wp)$ over four runs in all our experiments.
Unless specified, we evaluate\xiaowei{d} all combinations of $(id, m, wp)$.

% The real world dataset preparation can be divided into four parts:
% \\
% (1) A \textbf{multi-user} dataset comprising 155, 296 frames from nine participants is utilized to validate our method through a multi-user experiment. This dataset is denoted as $D_{MU}$.
% \\
% (2) A \textbf{multi-motion} dataset encompassing 143, 552 frames involving four basic motions (run, walk, jump, clap) and complex real world motion (basketball, box, PingPong)is employed to validate our method's performance across various motion types. This dataset is labeled as $D_{MM}$.
% \\
% (3) A \textbf{multi-position} dataset comprising 219, 932 frames spanning a variety of scale displacements is utilized to validate the effectiveness of our method across different wearing positions. This dataset is denoted as $D_{MP}$.
% \\
% {\it Ethical approval} was obtained for the experiment of data collection and publicity of the collected data. 

%We will make the dataset publicly available upon acceptance.
%The whole procedure was approved by the Medical Ethics Committee of our oganizrion. 
% Before the data collection, we obtained the participants’ written consent after informing them of the experiment’s purpose and procedure.

\subsubsection{Simulation Dataset $\mathcal{D}_s$}
% 虚拟数据的生成流程

With the simulation platform introduced above, we built our simulation dataset using around 30,000 motion capture frames involving eleven complex semantic motions (e.g., swing-dancing, chicken-dancing, hybrid motions) gathered from the Mixamo online animation library~\cite{Mixamo}.

\subsection{Comparison with Supervised Learning}

% The results show that our support-based method performs better than other SOTA methods.

As Table~\ref{tab:comp_super} shows, although using {\it no} real data, the proposed method, denoted as Ours (Sim2Real), achieves a high accuracy comparable to that of supervised learning, demonstrating its effectiveness in flexible sensor-based human motion capture. 
To further justify our choice of using simulated rather than real data, we compar\xiaowei{ed} it to an additional baseline, denoted as Ours (Real2Real), which applies the proposed SuDA to a piece of real data collected with a random user, motion, and wearing position. Experimental results show that Ours (Sim2Real) achieves much higher accuracy than Ours (Real2Real), suggesting that our simulation creates data with more diverse motion features and less noise at a lower cost, which is more desirable in practice.

\begin{table}[t]
    \caption{Results of our method (Sim2Real) and two baselines: supervised learning and ours (Real2Real).}
    \label{tab:comp_super}
    \centering
    % \captionof{table}{Experimental results of the proposed method, {\it i.e.}, Ours (Sim2Real) and two baselines: Supervised Learning and Ours (Real2Real).}
    \begin{tabular*}{0.8\hsize}{@{}@{\extracolsep{\fill}}lr@{}}
        \toprule
        Method  & MAE (deg) \\
        \midrule
        Ours (Real2Real)                       & 11.91 $\pm$  4.87 \\
        {\bf Ours (Sim2Real)}            & 7.60 $\pm$  2.58 \\
        \midrule
        Supervised                        &   6.42  $\pm$  2.73 \\
        \bottomrule
    \end{tabular*}

   % \vspace{-4mm}
% \vspace{-10mm}
% \vspace{-8mm}
    % \caption{Comparison with state-of-the-art methods.}
\end{table}
       % \vspace{-20pt}

\subsection{Comparison with State-of-the-Art Methods}
 % \vspace{-0.5cm}
\begin{table}[t]
    \caption{Comparison of our method (SuDA) with SOTA distribution-based domain adaptation methods.}
    \label{tab:comp_sota}
    \centering
    % \captionof{table}{Experimental results of the proposed method, {\it i.e.}, Ours (Sim2Real) and two baselines: Supervised Learning and Ours (Real2Real).}
    \begin{tabular*}{0.9\hsize}{@{}@{\extracolsep{\fill}}lr@{}}
        \toprule
        Method  & MAE (deg) \\
        \midrule
        Source Only                 & 21.93 $\pm$  8.83 \\
DAN \cite{long2015learning}     & 22.38 $\pm$  8.34 \\  
DANN \cite{ganin2015unsupervised} & 21.51 $\pm$  8.03 \\
D-CORAL\cite{sun2016deep}         & 21.89 $\pm$  8.85 \\ 
DAAN \cite{yu2019transfer}        & 21.80 $\pm$  8.28 \\
BNM \cite{cui2020towards}       & 21.70 $\pm$  8.61 \\ 

        {\bf Ours (SuDA)}            & 7.60 $\pm$  2.58 \\
        \midrule
        Supervised                   &   6.42  $\pm$  2.73 \\
        \bottomrule
    \end{tabular*}\
    % \caption{Comparison with state-of-the-art methods.}

% \vspace{-0.9cm}
\end{table}

\begin{table*}[t]
\centering
\caption{Performance comparison over different users. Source Only: the same predictive neural network trained on the source domain (simulated) data only. We enumerate on $id$ while averaging over all possible values of $m$ and $wp$ of $D^{id, m, wp}_{t^*,test}$.}
\label{tab:users}
\scalebox{0.65}{
\renewcommand{\arraystretch}{1.1}
% \begin{tabular}{ccccccccccccc}
\begin{tabular}{lrrrrrrrrrrr}
\toprule
Method      & $id=$1           & 2           & 3           & 4           & 5           & 6           & 7           & 8           & 9           & 10           & 11          \\
\hline
Source Only & 17.94$\pm$0.34 & 32.14$\pm$1.14 & 12.94$\pm$2.28 & 24.34$\pm$0.67 & 15.34$\pm$0.98 & 16.62$\pm$1.63 & 21.26$\pm$0.69 & 33.28$\pm$5.58 & 18.34$\pm$1.12 & 14.90$\pm$3.37 & 26.15$\pm$1.93  \\
DAN         & 21.17$\pm$3.62 & 34.43$\pm$1.50 & 12.81$\pm$0.18 & 24.96$\pm$1.05 & 17.43$\pm$0.67 & 18.02$\pm$0.73 & 20.36$\pm$0.33 & 33.95$\pm$4.14 & 16.80$\pm$2.48 & 14.37$\pm$3.03   & 24.37$\pm$0.82  \\
DANN        & 17.62$\pm$1.48 & 31.60$\pm$1.24 & 12.02$\pm$0.70 & 24.37$\pm$0.98 & 14.72$\pm$2.30 & 16.31$\pm$1.22 & 19.65$\pm$0.67 & 33.43$\pm$5.69 & 18.96$\pm$1.59 & 15.33$\pm$3.53   & 24.75$\pm$0.39  \\
D-CORAL     & 17.46$\pm$0.87 & 31.82$\pm$1.25 & 12.23$\pm$0.77 & 25.20$\pm$1.12 & 13.29$\pm$2.18 & 16.21$\pm$2.24 & 21.09$\pm$0.69 & 33.56$\pm$3.73 & 17.11$\pm$2.20 & 15.13$\pm$3.16   & 26.18$\pm$1.17  \\
DAAN        & 18.43$\pm$0.35 & 33.16$\pm$1.39 & 12.53$\pm$0.68 & 25.36$\pm$0.71 & 14.12$\pm$1.25 & 16.40$\pm$2.48 & 21.03$\pm$0.20 & 31.67$\pm$5.01 & 16.46$\pm$3.24 & 15.51$\pm$2.62   & 25.36$\pm$0.77  \\
BNM         & 18.41$\pm$0.44 & 32.14$\pm$1.14 & 12.15$\pm$0.53 & 24.34$\pm$0.68 & 13.20$\pm$2.00 & 16.11$\pm$2.10 & 21.08$\pm$0.77 & 33.53$\pm$4.02 & 16.99$\pm$2.02 & 14.99$\pm$3.29   & 26.53$\pm$2.17  \\

\textbf{SuDA (Ours)}     & \textbf{2.57$\pm$0.13}  & \textbf{2.55$\pm$0.09}   & \textbf{6.72$\pm$0.36} & \textbf{7.37$\pm$0.75}  & \textbf{5.76$\pm$0.08}  & \textbf{9.65$\pm$0.16}  & \textbf{7.59$\pm$0.16}  & \textbf{7.43$\pm$0.29}  & \textbf{10.16$\pm$0.72} & \textbf{12.87$\pm$3.16}  & \textbf{14.11$\pm$1.31}\\
\hline
Supervised  &  1.41$\pm$0.04 & 1.96$\pm$0.07  & 3.06$\pm$0.13  & 2.54$\pm$0.16  & 6.51$\pm$0.23  & 7.58$\pm$0.31 & 9.18$\pm$0.98  & 3.33$\pm$0.36  & 7.67$\pm$0.38  & 15.12$\pm$1.00   & 17.01$\pm$0.27 \\
\bottomrule
\end{tabular}

}
\vspace{-5mm}
\end{table*}

\begin{table*}[t]
\centering
\caption{Performance comparison over different wearing positions. Source Only: the same predictive neural network trained on the source
domain (simulated) data only. We enumerate on $wp$ while average over all possible values of $id$ and $m$ of $D^{id, m, wp}_{t^*,test}$.}
\label{tab:position}
\scalebox{0.6}{
\renewcommand{\arraystretch}{1.1}
% \begin{tabular}{ccccccccccccc}
\begin{tabular}{lrrrrrrrrrrrr}
\toprule
Method      & $wp=30^\circ$         & $60^\circ$        & $90^\circ$           & $120^\circ$          & $150^\circ$        & $180^\circ$        & $210^\circ$        & $240^\circ$        & $270^\circ$        & $300^\circ$        & $330^\circ$        & $360^\circ$        \\ \hline
Source Only & 17.94$\pm$0.34 & 13.57$\pm$0.67 & 12.11$\pm$1.38 & 17.21$\pm$3.90 & 21.75$\pm$1.98 & 27.04$\pm$1.76 & 46.70$\pm$1.64 & 24.07$\pm$1.45 & 16.60$\pm$0.57 & 12.96$\pm$0.49 & 19.78$\pm$2.15 & 11.34$\pm$0.53 \\
DAN         & 21.17$\pm$3.62 & 16.24$\pm$0.68 & 12.44$\pm$1.08 & 19.59$\pm$2.79 & 25.44$\pm$1.62 & 25.41$\pm$0.49 & 45.01$\pm$3.26 & 24.04$\pm$3.53 & 16.28$\pm$0.67 & 13.02$\pm$0.57 & 20.34$\pm$0.69 & 19.85$\pm$0.91 \\
DANN        & 17.62$\pm$1.48 & 13.96$\pm$0.44 & 12.30$\pm$1.00 & 19.20$\pm$2.10 & 22.00$\pm$2.13 & 28.15$\pm$2.60 & 42.92$\pm$3.64 & 24.03$\pm$0.62 & 17.04$\pm$1.02 & 12.77$\pm$0.22 & 19.37$\pm$1.62 & 20.24$\pm$0.31 \\
D-Coral     & 17.46$\pm$0.87 & 14.04$\pm$1.00 & 12.51$\pm$1.08 & 19.74$\pm$4.73 & 24.34$\pm$2.53 & 28.50$\pm$1.49 & 46.75$\pm$1.38 & 24.52$\pm$0.58 & 16.65$\pm$0.22 & 13.09$\pm$0.25 & 19.50$\pm$1.39 & 19.95$\pm$0.87 \\
DAAN        & 18.43$\pm$0.35 & 13.76$\pm$0.28 & 12.81$\pm$1.45 & 20.92$\pm$2.70 & 24.76$\pm$2.41 & 26.92$\pm$2.19 & 44.70$\pm$3.09 & 24.85$\pm$0.66 & 15.83$\pm$0.93 & 12.67$\pm$0.18 & 19.52$\pm$1.69 & 19.65$\pm$0.44 \\
BNM         & 18.41$\pm$0.44 & 13.88$\pm$0.59 & 12.10$\pm$1.38 & 17.21$\pm$3.90 & 24.34$\pm$2.53 & 27.47$\pm$1.76 & 46.70$\pm$1.64 & 25.04$\pm$1.22 & 16.37$\pm$0.57 & 12.96$\pm$0.49 & 19.50$\pm$1.39 & 19.89$\pm$0.88 \\

{\bf SuDA (Ours)}        & \textbf{2.57$\pm$0.13}  & \textbf{5.55$\pm$0.36} & \textbf{5.41$\pm$0.23}  &\textbf{ 4.05$\pm$0.77}  & \textbf{9.39$\pm$0.14}  & \textbf{6.79$\pm$0.14}  & \textbf{8.18$\pm$0.25}  & \textbf{7.99$\pm$0.31}  & \textbf{10.59$\pm$0.42} & \textbf{11.69$\pm$0.24} &\textbf{ 13.45$\pm$0.19} & \textbf{5.83$\pm$0.23} 
\\ \hline
Supervised  & 1.41$\pm$0.04  & 6.52$\pm$0.45  & 8.05$\pm$0.86  & 2.55$\pm$0.11  & 6.88$\pm$0.29  & 6.25$\pm$0.30  & 8.49$\pm$0.39  & 4.83$\pm$0.19  & 6.61$\pm$0.09  & 7.61$\pm$0.13  & 10.86$\pm$0.23 & 6.59$\pm$0.24  \\
\bottomrule
\end{tabular}
}
\vspace{-5mm}
\end{table*}

As Table~\ref{tab:comp_sota} shows, we compare\xiaowei{d} our method with state-of-the-art (SOTA) distribution-based domain adaptation methods.
For a fair comparison, we have adapted the official code provided by the authors to share the same input and output format as ours. Please see the Appendix for more details on implementations. It can be observed that SuDA outperforms all SOTA DiDA methods, which demonstrates its superiority in flexible sensor-based human motion capture. 
% Our solution relies on a novel support-based domain adaptation method that aligns the supports of the predictive functions rather than the instance-dependent distributions between the source and target domains. This makes our adaptation method applicable to unpaired simulated and real data with inherently different body shapes, motions, sensor position.
% \subsection{Results on Different Users, Positions \& Motions}
\subsection{Robustness against Different Users, Motions \& Wearing Positions}
% \subsubsection{Performance Comparison on Different Users}
\subsubsection{Robustness against Different Users}
As Table~\ref{tab:users} shows, our method consistently achieves high accuracy comparable to supervised learning and \yq{demonstrates} significant improvements over competing methods for all 11 users, demonstrating the robustness of our method \xiaowei{across different users}.

% \subsubsection{Performance Comparison on Different Wearing Positions}
\subsubsection{Robustness against Different Wearing Positions}
Similar to the above, our method achieves \yq{high} accuracy comparable to supervised learning and demonstrates its superiority over competing methods across different wearing positions (Table~\ref{tab:position}).

\subsubsection{Robustness against Different Motions}
The same conclusions hold for different motions: Ours is comparable to supervised learning while significantly outperforming competing methods (Table~\ref{tab:motion}).
% We conducted a comparison between our method and state-of-the-art distribution-based methods on four fundamental single-person actions (jump, run, clap, walk), as detailed in Table~\ref{tab:motion}. The result shows that distribution-based method can handle the gap.

In addition, we observed that: i) the accuracy of both our SuDA method and supervised learning change with different users, wearing positions and motions, showing that the inherent variations among instances still account for significant differences in performance; ii) in most cases, most state-of-the-art (SOTA) domain adaptation methods work comparably to Source Only, that is, applying models trained on the source domain (simulated) data only, indicating that they failed at our task and did not perform much meaningful adaptation at all. We ascribe their failure to the inherently different distributions between the two domains and extremely low dimensionality of our sensor data.

\begin{table}[t]
\centering
\caption{Performance comparison over different motions. Source Only: the same predictive neural network trained on the source domain (simulated) data only. We enumerate on $m$ while average over all possible values of $id$ and $wp$ of $D^{id, m, wp}_{t^*,test}$.}
\label{tab:motion}

\scalebox{0.75}{
\renewcommand{\arraystretch}{1.1}
% \begin{tabular}{ccccccccccccc}
    \begin{tabular}{lrrrr}
    \toprule
    Method      & $m=$Jump        & $m=$Run          & $m=$Clap         & $m=$Walk     \\    \hline
    Source Only & 27.49$\pm$0.49 & 13.76$\pm$0.62 & 36.19$\pm$2.40 &  31.34$\pm$3.08 \\
    DAN         & 27.29$\pm$1.51 & 12.79$\pm$0.63 & 37.62$\pm$0.70 & 17.15$\pm$1.87 \\
    DANN        & 26.04$\pm$0.44 & 13.50$\pm$0.27 & 37.29$\pm$1.53 & 17.86$\pm$0.90 \\
    D-Coral     & 27.79$\pm$0.64 & 13.78$\pm$0.63 & 35.85$\pm$1.55 & 17.72$\pm$1.78 \\
    DAAN        & 28.04$\pm$0.30 & 13.86$\pm$0.79 & 36.78$\pm$1.91 & 17.45$\pm$2.05 \\
    BNM         & 27.62$\pm$0.50 & 13.63$\pm$0.64 & 36.19$\pm$2.40 & 17.54$\pm$1.65 \\
    
    {\bf SuDA (Ours)}     & \textbf{7.63$\pm$0.07}  & \textbf{11.00$\pm$0.10} & \textbf{3.98$\pm$0.17} & \textbf{8.59$\pm$0.23} \\
    \hline
    Supervised  & 6.51$\pm$0.05  & 10.55$\pm$0.23 & 4.30$\pm$0.13  & 7.87$\pm$0.08  \\
    \bottomrule
    \end{tabular}
       % \vspace{-10mm}
}
\vspace{-5mm}
\end{table}

\section{Applications}
\yq{{\bf Performance in real-world scenarios.}} We recruited five new participants of varying body shapes to investigate how our proposed method works in real-world applications. We asked them to wear our elbow pads in three different arbitrary positions and perform three distinct real-world activities respectively: Table Tennis, Basketball, and Boxing.
In total, we collected 45 (5 participants $\times$ 3 wearing positions $\times$ 3 real-world activities) unique data segments, comprising 81, 848 frames.
As Table~\ref{tab:appliactions} shows, our method achieves a high accuracy comparable to supervised learning. It significantly outperforms SOTA distribution-based domain adaptation methods, demonstrating that our SuDA method generalizes well to real-world scenarios.

\begin{table}[t]
\centering
% \captionof{table}{Applications in real-world scenarios. The results are averaged over 5 new participants with varying body shapes who wore the elbow pad in three arbitrary positions.}
\caption{Applications in real-world scenarios. The results are averaged over 5 new participants with varying body shapes who wore the elbow pad in three arbitrary positions.}
\label{tab:appliactions}
\scalebox{0.75}{
\renewcommand{\arraystretch}{1.1}
% \begin{tabular}{ccccccccccccc}
    \begin{tabular}{lrrrr}
    \toprule
    Method      & \textbf{Table Tennis}           & \textbf{Basketball}         & \textbf{Boxing}             & \textbf{Average}                 \\ \hline
    Source Only & 21.53$\pm$6.40          & 19.39$\pm$4.52          & 20.89$\pm$8.71          & 20.61$\pm$0.90          \\ 
    DAN         & 21.60$\pm$5.74          & 18.80$\pm$3.81          & 19.42$\pm$6.41          & 19.94$\pm$1.20          \\
    DANN        & 21.73$\pm$6.39          & 18.79$\pm$3.82          & 20.10$\pm$8.41          & 20.47$\pm$1.24          \\ 
    D-Coral     & 22.00$\pm$6.04          & 18.90$\pm$4.81          & 21.14$\pm$8.31          & 20.68$\pm$1.31          \\
    DAAN        & 21.40$\pm$6.07          & 17.78$\pm$4.34          & 20.85$\pm$8.33          & 20.01$\pm$1.60          \\
    BNM         & 21.74$\pm$6.06          & 19.26$\pm$5.08          & 21.26$\pm$8.73          & 20.76$\pm$1.08          \\

        {\bf SuDA (Ours)}    & \textbf{10.56$\pm$2.30} & \textbf{11.62$\pm$3.53} & \textbf{10.02$\pm$3.18} & \textbf{10.74$\pm$0.67} \\ 
        \hline
        Supervised  & 12.28$\pm$3.50          & 11.90$\pm$3.06          & 11.96$\pm$3.75          & 12.05$\pm$0.17          \\
        \bottomrule
    \end{tabular}
}
\vspace{-5mm}
\end{table}

\yq{{\bf Generalization across joints.}} \jiawei{Besides, we recruited another three participants to investigate how SuDA works in different joints. In total, we collected 12 (3 participants $\times$ 4 joints) unique data segments, comprising 22, 577 frames. As Table~\ref{tab:scenario} shows, our method achieves a high accuracy comparable to supervised learning. It significantly outperforms SOTA distribution-based domain adaptation methods, demonstrating that our SuDA method generalizes well to different joints. Notice that we only use right elbow simulation data as our source domain.}

\begin{table}[t]
\centering
\caption{Applications in four different joints. The results are averaged over 3 new participants with varying body shapes who wore the elbow pad and knee pad.}
\label{tab:scenario}
\scalebox{0.75}{
\renewcommand{\arraystretch}{1.1}
% \begin{tabular}{ccccccccccccc}
    \begin{tabular}{lrrrr}
    \toprule
    Method      & \textbf{Left Elbow}       & \textbf{Right Elbow}        & \textbf{Left Knee }       & \textbf{Right Knee  } \\    \hline
    Source Only & 39.29$\pm$3.78 & 40.85$\pm$5.13 & 43.45$\pm$6.28 &  42.88$\pm$5.80 \\
    DAN         & 38.01$\pm$4.75 & 37.98$\pm$5.80 & 41.92$\pm$5.81 & 40.32$\pm$6.28 \\
    DANN        & 40.22$\pm$3.50 & 40.89$\pm$4.83 & 42.92$\pm$5.35 & 42.01$\pm$6.05 \\
    D-Coral     & 39.23$\pm$5.26 & 41.25$\pm$5.91 & 43.18$\pm$5.27 & 40.14$\pm$5.10 \\
    DAAN        & 39.65$\pm$3.58 & 40.59$\pm$7.92 & 42.92$\pm$5.35 & 42.01$\pm$6.05 \\
    BNM         & 38.78$\pm$4.60 & 40.76$\pm$5.54 & 42.01$\pm$6.46 & 42.28$\pm$5.82 \\
    
    {\bf SuDA (Ours)}     & \textbf{9.09$\pm$1.89}   & \textbf{9.86$\pm$2.55}& \textbf{12.13$\pm$2.02} & \textbf{10.84$\pm$1.84} \\
    \hline
    Supervised  & 6.07$\pm$0.84  & 7.50$\pm$2.84 & 8.20$\pm$0.40  & 9.12$\pm$3.06  \\
    \bottomrule
    \end{tabular}
       \vspace{-4mm}
}
\vspace{-5mm}
\end{table}

% \jiawei{Due to SuDA only relying on matching to the support scope, it just requires one joint bending for calibration. We have users perform calibration at different speeds (six stages) and test the mocap effects in daily wearing after calibration. This aims to demonstrate that SuDA achieves comparable accuracy with a very short amount of data.}

\section{Discussions}
\subsection{MAE vs. Size of Simulation Dataset}
\begin{figure}[t]
\centering
\includegraphics[width=0.8\linewidth]{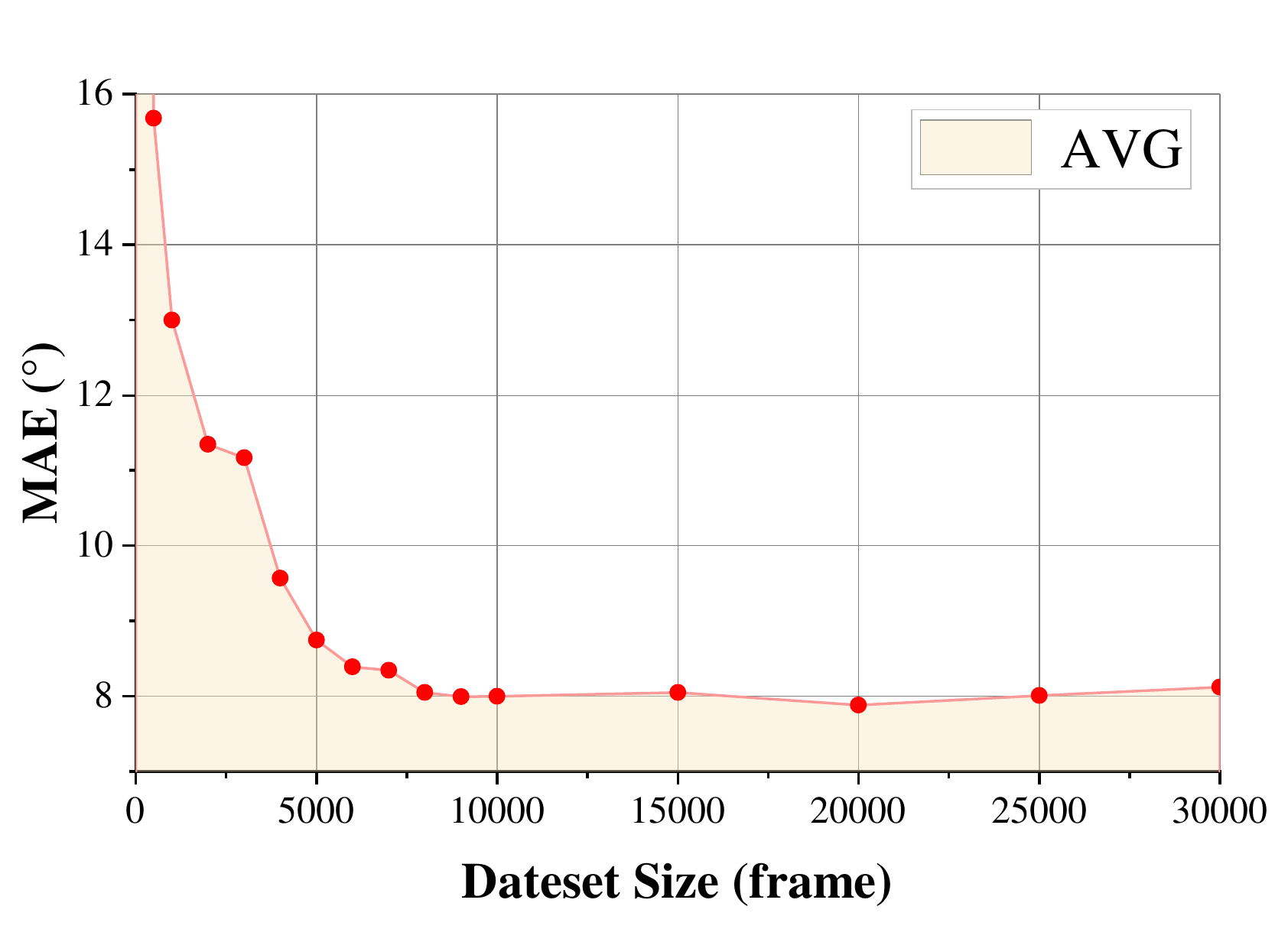}
\vspace{-10pt}
\caption{MAE vs. size of simulation dataset. MAE drops quickly between 0 and 5,000 frames with the increase in the size of the simulation dataset. Between 5,000 and \jiawei{10,000} frames, the MAE drops slightly towards convergence. With more than 10,000 frames, MAE becomes stable against the increase in dataset size.
}
\label{fig:dataset}
\vspace{-5mm}
\end{figure}

As Fig.~\ref{fig:dataset} shows, MAE decreases quickly  \xiaowei{when the size of the simulation dataset increases from 0 to 5,000 frames}, followed by a slower decline, and eventually converges around 10,000 frames.
This indicates that the support of the source domain (simulated) can be well approximated with more than 10,000 motion frames.
In all our experiments, we use a safer option of 30,000 frames as the size of our simulation dataset.
% We also conducted an analysis of the impact of virtual data scale on the final error. 
% As shown in the Figure~\ref{fig:dataset}, with an increase in virtual data scale, the error gradually decreases. It experiences a rapid decrease in the initial 5000 frames, followed by a slower decline, and eventually converges around 20000 frames. This behavior can be attributed to the fact that the support-based method aims to find the optimal match between the virtual domain and the real domain's supports. With a larger amount of virtual data, the number of supports also increases, resulting in a finer match with the target domain. This enables the learning of a more comprehensive $f_t$, leading to higher tracking accuracy.

% \paragraph{Support-based Analysis}

% \begin{figure}[ht]
% \centering
% \includegraphics[width=1.2\linewidth]{CameraReady/LaTeX/Fig_Timeindex.jpg}
% \caption{Time sequence diagrams for two distinct actions}
% \label{fig:hardware}
% \setlength{\abovedisplayskip}{0pt}
% \setlength{\belowdisplayskip}{0pt}
% \end{figure}

%\section{Discussions}
%We propose a support-based domain adaptation method to address this issue. Extensive experimentation has demonstrated that our approach can mitigate the challenges arising from the diversity in motion, human characteristics, and positions of the data, achieving accuracy levels approaching, and sometimes even surpassing, those achieved by supervised methods.

\subsection{Distribution-based vs. Support-based Domain Adaptation Methods (DiDA vs. SuDA)}

% Comparison between two different domain adaptation methods: Distribution-based Domain Adaptation (DiDA) and Support-based Domain Adaptation (SuDA).

\paragraph{Objective} DiDA methods have the same goal as SuDA - aligning the supports of the predictive functions between the source and target domains, {\it i.e.}, $x_s \leftrightarrow x_t$ so that the predictive function that works on the source domain $f_s$ can also be applied to $x_t$. However, they take an indirect approach to achieve this goal which aligns the feature distributions of the source and target domains.
% \vspace{-5mm}
% as it's hard to find supports registration of high dimensional features.

% The effectiveness of this approach is based on three assumptions: i) the same underlying feature is shared between domains ({\it e.g.}, cat and dog face both contain features like eyes, nose, etc.); ii) the data collected from the source and target domains share the same distribution in the feature mentioned above space ({\it e.g.}, the source and target domain datasets contain the same proportion of cat faces and dog faces respectively); iii) the feature is adequate for the final task ({\it e.g.}, face detection).

% When these three assumptions are satisfied, the feature extractor implicitly aligns the support of both domains ({\it i.e.}, cat and dog faces) together into the exact proxy representation in the space of some practical feature and achieves the goal of domain adaptation.

\begin{figure}[H]
\vspace{-5pt}
   \centering
     \begin{subfigure}[b]{0.32\linewidth}
         \centering
         \includegraphics[width=\linewidth]{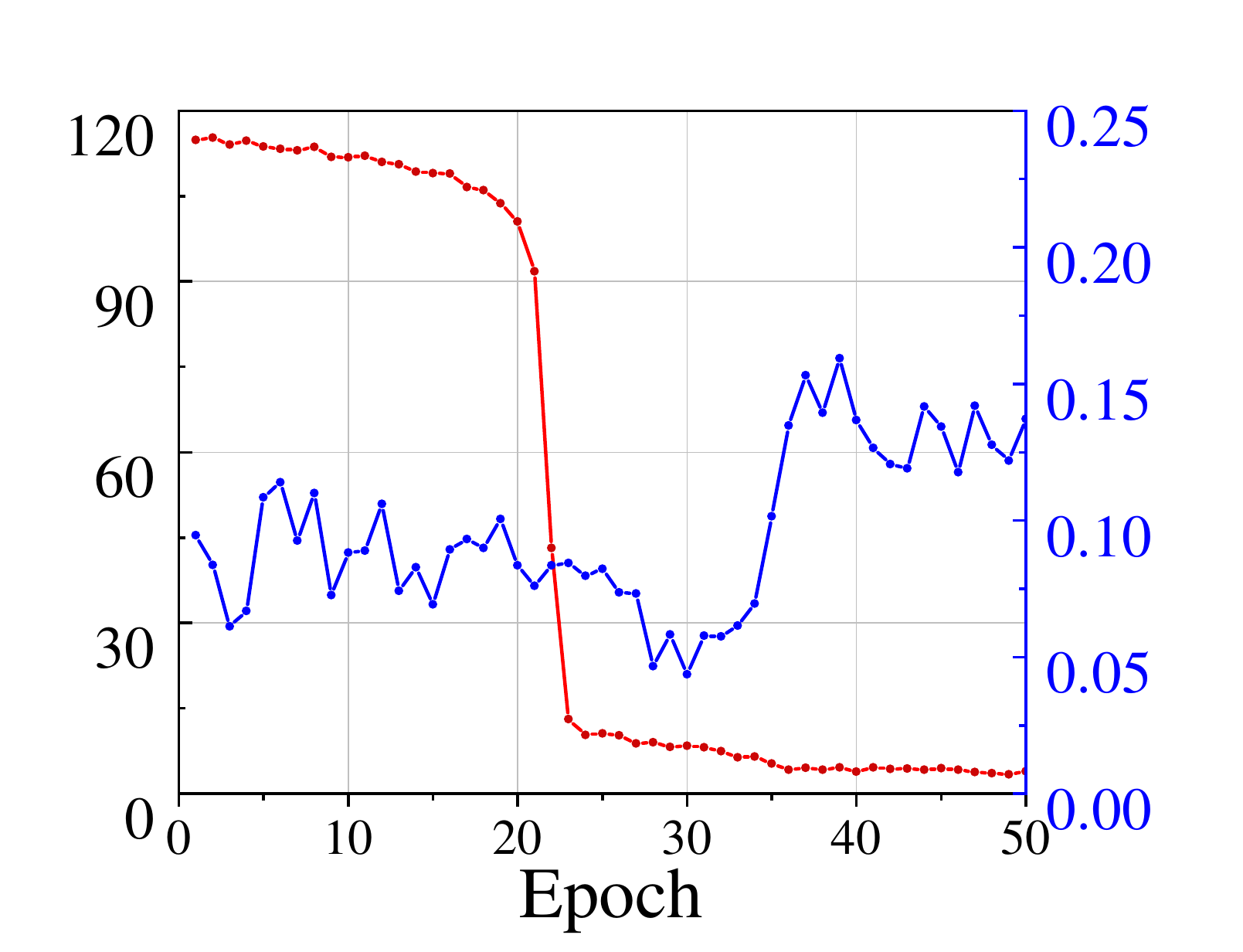}
              \vspace{-15pt}
         \caption{DAN}
         \label{fig:fcase 1}
     \end{subfigure}
     \begin{subfigure}[b]{0.32\linewidth}
         \centering
         \includegraphics[width=\linewidth]{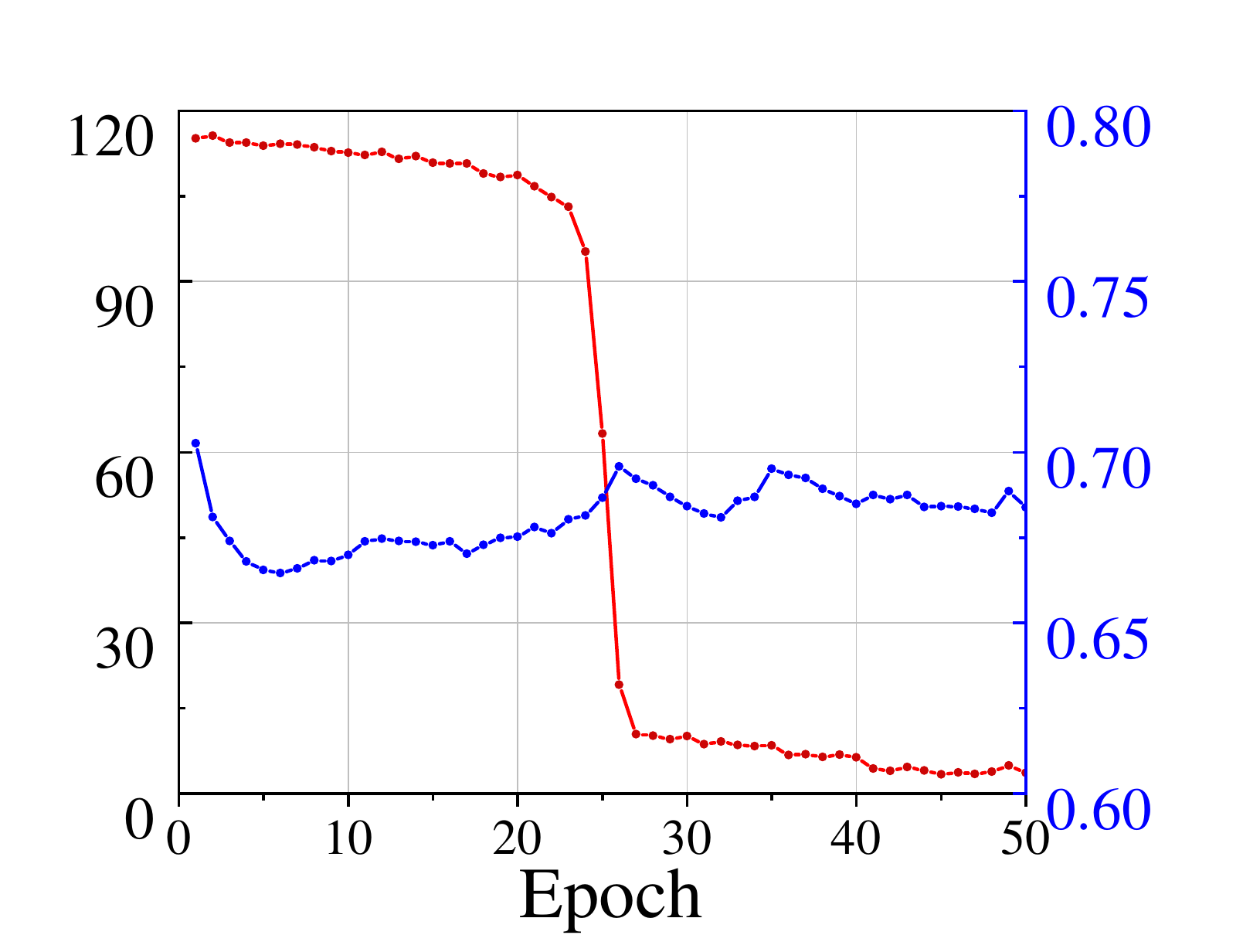}
              \vspace{-15pt}
         \caption{DANN}
         \label{fig:fcase 2}
     \end{subfigure}
     % \vspace{10pt}
     \begin{subfigure}[b]{0.32\linewidth}
         \centering
         \includegraphics[width=\linewidth]{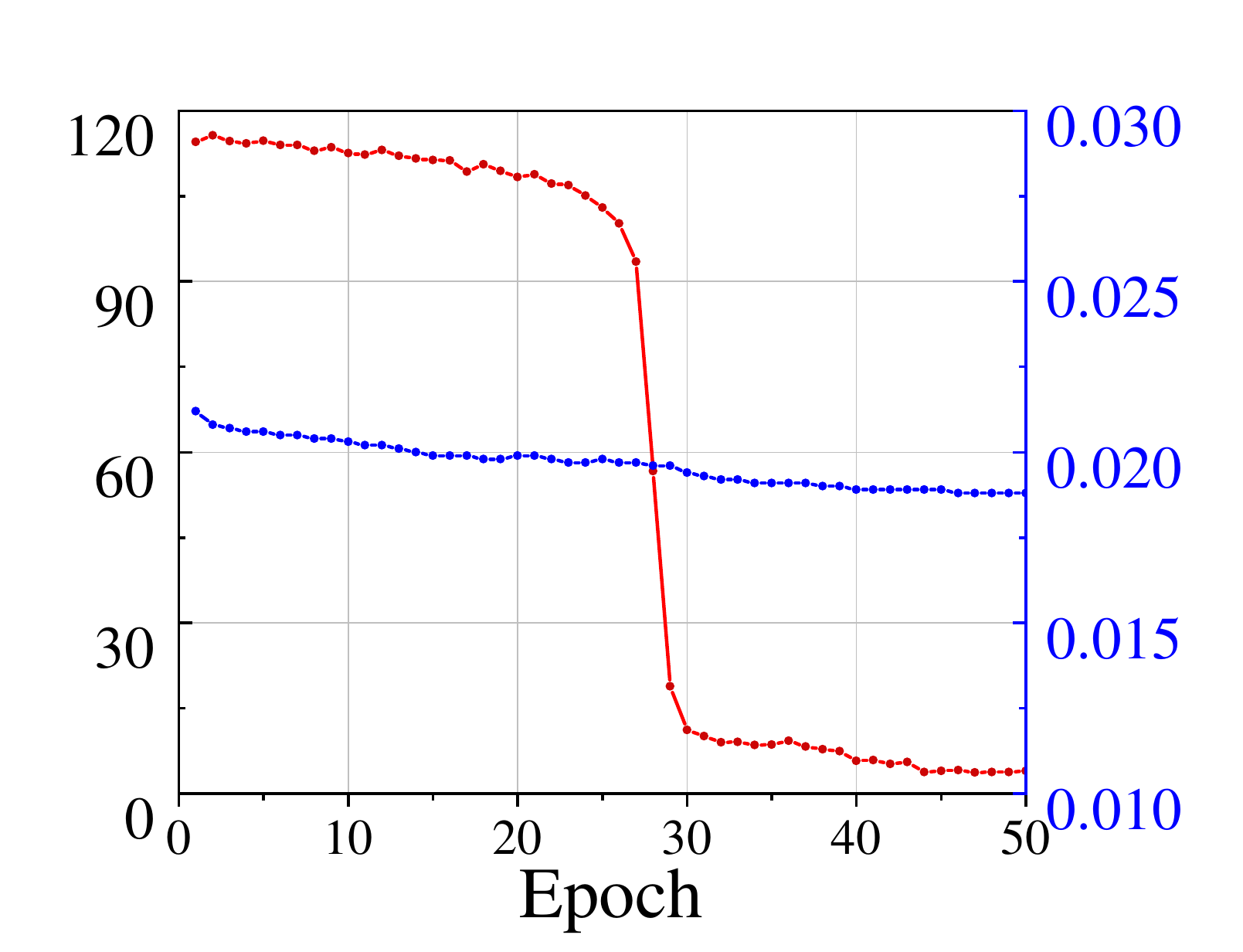}
              \vspace{-15pt}
         \caption{DAAN}
         \label{fig:fcase 2}
     \end{subfigure}
     \vspace{-4mm}
     \caption{ Training loss of distribution-based methods. \textcolor{red}{Red Line}: Supervised Loss; \textcolor{blue}{Blue Line}: Transfer Loss.}
     \label{fig:transfer_loss}
     \vspace{-10pt}
\label{fig:loss}
\end{figure}

% \paragraph{Data Quantity} 
% SuDA generally does not rely on large data quantity too much yet DiDA requires a huge number of data to estimate distribution. 
% In our joint-angle task, gathering full-scope support is easy, requiring only one full joint bending, which ensures the quick calibration to sensor displacements.
\vspace{-5mm}
\jiawei{\paragraph{Data Distribution} 
SuDA generally relies on function support yet DiDA relies more on data distribution. In our Sim2Real task, the distribution gap between the two domains (\textit{e.g., real running and simulated dancing, real knee and simulated elbow}) varies remarkably, leading to failure on all DiDA methods. Please see Appendix D about Distribution Variations of our data.
\vspace{-5mm}
\paragraph{Data Dimension}
SuDA is more applicable to low-dimensional data yet DiDA is suitable to high-dimensional one. Low-dimensional data provides significant advantages for directly understanding the function supports. In our task, the extremely low (readings from two flexible sensors) dimension of our data makes us easy to find support registration methods, which motivates this work and leads to success. 
Despite the success of DiDA methods ({\it e.g.}, in computer vision), they have a hidden limiting assumption - that the data being handled must be complex and high-dimensional to provide enough space for extracting lower-dimensional features. This reliance on high-dimensional data restricts the applicability of distribution-based methods to low-dimensional datasets like the one used in this paper. 
}
Hence, the impact of diverse data distributions and low dimensionality of sensor data makes DiDA methods have no advantage over naive source-only methods. Empirically, we verify this in Fig.~\ref{fig:loss}, where the “transfer loss” measuring distributional distances does not reduce during training, yielding similar performance as Source-only.

% However,  DiDA are hard to generalize as significant gap between the source and target low-dimensional data distributions (\textit{Sim v.s. Real}). Thus, the SuDA approach meaningfully complements DiDA for the domain adaptation of low-dimensional data, which makes a significant contribution to the field.

% Similarly, SuDA is difficult to apply to high-dimensional data because it is difficult to find support registration for high-dimensional data. 

\vspace{-13pt}

\subsection{Limitations}
Although effective, the proposed method is based on the assumption that function support in the source and target domain can be registered properly.
Specifically, for flexible sensor-based MoCap, the source and target data must have similar ranges of joint angles during motions.
Observing that the effective range of joint angles is similar for most people performing typical motions (around $40^\circ$ to $160^\circ$), leading to matching to the support scope, so we find that our method works well in most real-world scenarios after calibration.
However, this assumption becomes problematic in some extreme cases when the effective range of elbow joint angles differs between domains (please see failure cases in the Appendix).
% As our method is support-based, it necessitates knowledge about the support ranges in both the source and target domains. When the support ranges in the two domains align perfectly, our method can achieve higher accuracy. Yet, this issue is not exclusive to support-based methods; distribution-based transfer methods also face this challenge and require consistent feature scales between source and target domains for optimal performance.
% Another limitation is that support-based methods are sensitive to the results of position parameterization. 
% Optimal parameterization guides the source and target domains to find better support matches, resulting in higher tracking accuracy.

\vspace{-3mm}
\section{Conclusion}
To address the costly real data collection process in motion capture using flexible sensors, we propose a novel \emph{Sim2Real} solution based on domain adaptation, eliminating the need for labeled data yet achieving comparable accuracy.
Our solution relies on a novel \emph{support-based} domain adaptation method that aligns the \emph{supports} of the predictive functions rather than the instance-dependent distributions between the source and target domains. This makes our adaptation method applicable to unpaired simulated and real data with inherently different body shapes, motions, sensor positions, etc., as long as they share similar ranges of joint motion. 
Experimental results demonstrate that the effectiveness of our unsupervised support-based method is comparable with supervised learning and its superiority over state-of-the-art distribution-based domain adaptation methods.
In future work, we plan to explore applying the proposed method to other applications, especially those involving low-dimensional data.

\bibliographystyle{ACM-Reference-Format}
\bibliography{example_paper.bib}

%%%%%%%%%%%%%%%%%%%%%%%%%%%%%%%%%%%%%%%%%%%%%%%%%%%%%%%%%%%%%%%%%%%%%%%%%%%%%%%
%%%%%%%%%%%%%%%%%%%%%%%%%%%%%%%%%%%%%%%%%%%%%%%%%%%%%%%%%%%%%%%%%%%%%%%%%%%%%%%
% APPENDIX
%%%%%%%%%%%%%%%%%%%%%%%%%%%%%%%%%%%%%%%%%%%%%%%%%%%%%%%%%%%%%%%%%%%%%%%%%%%%%%%
%%%%%%%%%%%%%%%%%%%%%%%%%%%%%%%%%%%%%%%%%%%%%%%%%%%%%%%%%%%%%%%%%%%%%%%%%%%%%%%
\newpage
\appendix
\onecolumn
\section{Hardware}

We design and develop a prototype by adding two capacitive strain sensors to the standard elbow and knee pads, which are placed on the ulnar side (elbow) and tibial side (knee). 

\begin{figure}[ht]
\centering
\includegraphics[width=1.0\linewidth]{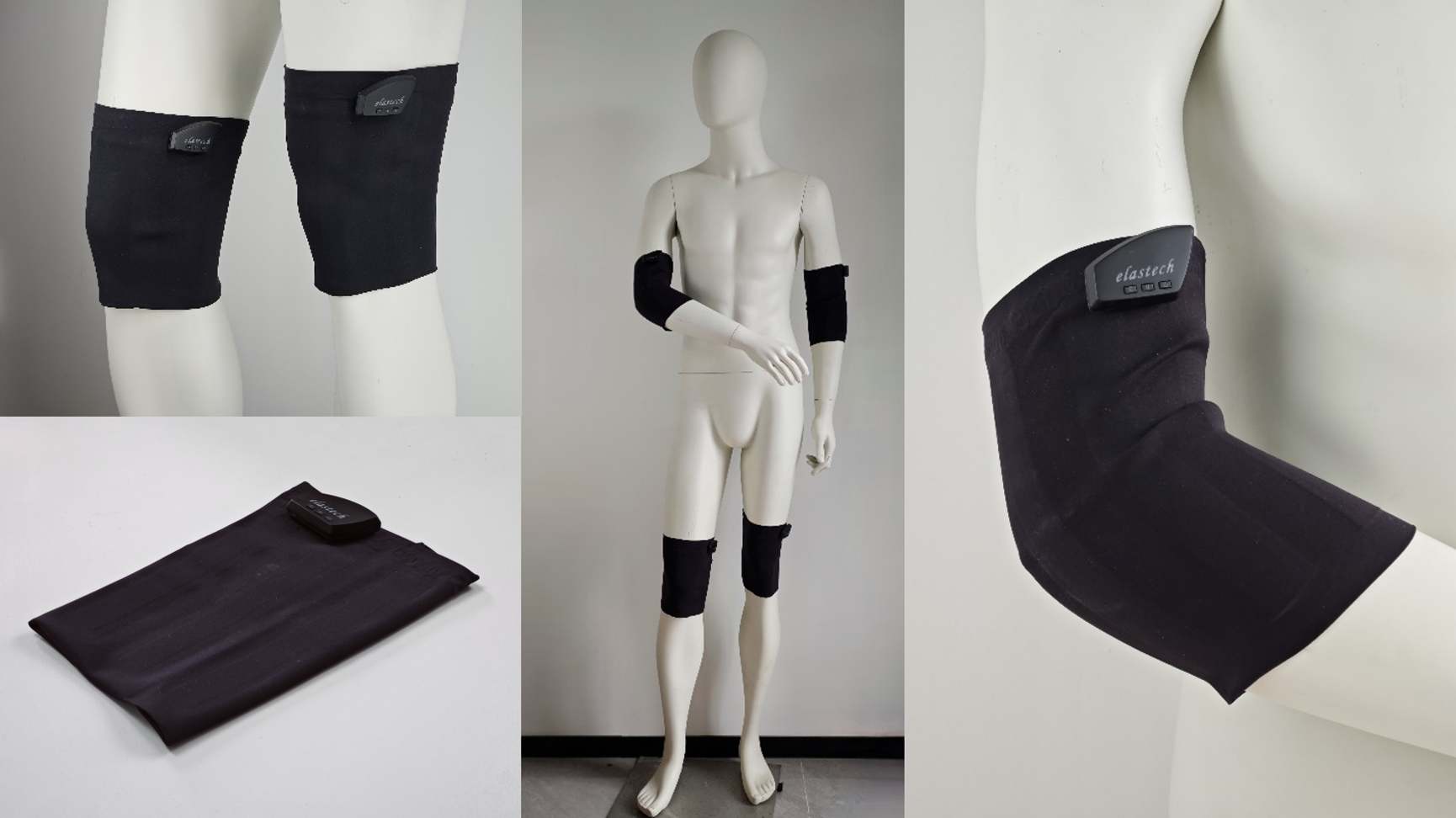}
\caption{Our smart elbow and knee pad (hardware).}
\label{fig:hardware}
\end{figure}

% which are distributed on the side of the olecranon of the elbow.
The length and circumference of the pad are 20 and 25 cm (elbow pad), and 20 and 40cm (knee pad).
The two sensors are placed 2 cm apart.
Our method aims to estimate the bending angle, $\theta$, of an elbow joint (Fig.~\ref{fig:hardware}) from the sensor readings of the two flexible sensors. 
The elbow bending angle is defined as the angle in the sagittal plane between the humerus and the central line between the radius and the ulna. In contrast, the knee bending angle is defined as the angle in the sagittal plane between the femur and the central line between the tibia and the fibula.

Fabric sensors are purchased as off-shelf products from ElasTech.
They are capacitive, {\it i.e.}, and their capacitance increase with the stretch caused by the bending of the arm.
% their capacitance increases as the arm bends and stretches the sensor.
The sensor readings are digitized to values in the range [0, 1023] and transmitted wirelessly via Bluetooth Low Energy at a frame rate of 50Hz. 
The collected sensor reading\xiaowei{s} are $x$ in the target domain.
% \section{Implementation of SOTA Methods}

For capacitive strain sensors, due to their high stretchability, lightness, and ease of attachment near joints, are suitable for monitoring human body movements in unrestricted wearable devices. As flexible materials inherently possess stickiness and elasticity, and joint movements involve twisting, bending, and other deformations, hysteresis and low repeatability may occur. To address these issues, we have taken the following measures:

\begin{enumerate}
    \item We selected the Elas Tech ESSB-01 commercial capacitive sensor, which features high reliability (over 300,000 bending cycles), thinness (as thin as 1mm), and high linearity.
    \item In order to improve repeatability and suppress hysteresis, we used a blended fabric with high spandex content and stretchable wires to assemble the sensor and the entire system. We pre-stretched the assembled system to eliminate its internal stress, and measured a linearity of $R^2 = 0.999$ after pre-stretching.
\end{enumerate}

% \begin{figure}[H]
%    \centering
%      \begin{subfigure}[b]{0.30\linewidth}
%          \centering
%          \vspace{-10mm}
%          \includegraphics[width=0.9\linewidth]{CameraReady/LaTeX/Fig_Sensors.jpg}
%           \caption{Our capacitive strain sensor with 16.5 cm long and 2.5 cm wide.}
%          \label{fig:Sensors}
%      \end{subfigure}
%      \begin{subfigure}[b]{0.33\linewidth}
%          \centering
%          \includegraphics[width=\linewidth]{CameraReady/LaTeX/Fig_Strain.png}
%          \caption{strain-versus-capacitance characterization of a soft sensor.}
%          \label{fig:Strain}
%      \end{subfigure}
%         \begin{subfigure}[b]{0.33\linewidth}
%          \centering
%          \includegraphics[width=\linewidth]{CameraReady/LaTeX/Fig_Hysterisis.png}
%          \caption{strain-versus-capacitance characterization of a soft sensor.}
%          \label{fig:Hysterisis}
%      \end{subfigure}  
%      \caption{points with the same parameter have the same label and are registered together by $R^Q$.}
%      \label{fig:suda}
% \end{figure}

\section{Implementation Details}

\begin{figure}[ht]
\centering
\includegraphics[width=1.0\linewidth]{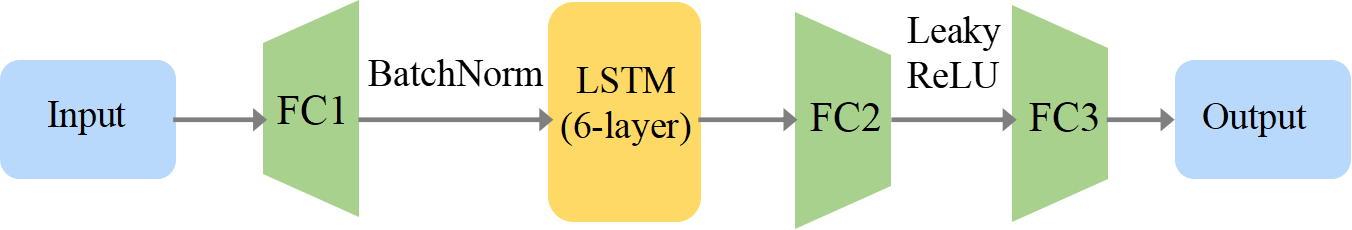}
\caption{Detailed network architecture. LSTM: Long Short-Term Memory. FC: Fully-Connected.}
\label{fig:LSTM}
\end{figure}

% \begin{figure}[t!]
%       \centering  
%       % \vspace{-2pt}
%      \begin{subfigure}[b]{0.49\linewidth}
%          \centering
%         \includegraphics[width=1.2\linewidth]{CameraReady/LaTeX/Fig_LSTM.png}
%          \caption{Detailed network architecture. LSTM: Long Short-Term Memory. FC: Fully-Connected.}
%          \label{fig:LSTM}
%      \end{subfigure}  

%     % \vspace{-15pt}
% \end{figure}

Fig.~\ref{fig:LSTM} shows the proposed network's architecture details. Specifically, we use a 3-layer LSTM network with a hidden layer size of 256 for sequence feature extraction and feed both its hidden state and cell state into FC2 for subsequent computation. 
% Note that the ten regressors (1 main + 9 auxillary) implemented by FC3 share the same feature vector. 
The output sizes of network layers are listed in Table~\ref{tab:net components}.

\begin{table}[!ht]
    \centering
    \begin{tabular}{lc}
        \hline
        Layer                  & Output Size \\
        \hline
        FC1                        & 256      \\
        LSTM                       & $6\times256\times2$      \\
        FC2                        & 128       \\
        FC3                        & 1       \\
        \hline
    \end{tabular}
    \caption{Details of network layers. }
    \label{tab:net components}
\end{table}

To make a fair comparison, we used the same architecture shown in Table~\ref{tab:net components} and the same hyper-parameters.

In addition, we used the same loss $\mathcal{L}_{mae}$ (please see the main paper) for the supervised pre-training but made the following modifications for different competitors:
\begin{itemize}
    % \item {\bf FFE}: Remove FFE from the model.
    % \item {\bf MMD} \cite{long2015learning}: We used MMD as $\mathcal{L}_{d}$. The output of FC2 is used for $\mathcal{L}_d$ computation. 
    \item {\bf MMD} \cite{long2015learning}: We applied a Max Mean Discrepancy (MMD) loss at the output of FC2 for the adaptation. 
    % \item {\bf D-CORAL} \cite{sun2016deep}: Use CORAL as $\mathcal{L}_{d}$. The output of FC2 is used for $\mathcal{L}_d$ computation. 
    \item {\bf D-CORAL} \cite{sun2016deep}: We applied a CORAL loss at the output of FC2 for the adaptation. 
    \item {\bf DANN} \cite{ganin2015unsupervised}: We applied a gradient reverse layer and a domain classifier (a 2-layer fully connected network) at the output of FC2 for the adaptation.
    % \item {\bf DAAN} \cite{yu2019transfer}:  We applied a DAAN loss at the softmax output of FC2 for the adaptation and updated its dynamic factor every epoch.
    \item {\bf DAAN} \cite{yu2019transfer}:  We added a softmax layer after FC2 and applied a DAAN loss at its output for the adaptation. The dynamic factor was updated every epoch.
    \item {\bf BNM} \cite{cui2020towards}: We applied a Batch Nuclear-norm Maximization  (BNM) loss at the output of FC2 for the adaptation.
\end{itemize}

% \begin{figure} 

% \begin{minipage}{.5\linewidth} 
% \vspace{-5mm}
% \centering
%  \begin{tabular}{lc}
%         \hline
%         Layer                  & Output Size \\
%         \hline
%         FC1                        & 256      \\
%         LSTM                       & $6\times256\times2$      \\
%         FC2                        & 128       \\
%         FC3                        & 1       \\
%         \hline
%     \end{tabular}
% \captionof{table}{Details of network layers. } 
% \end{minipage} 
% \begin{minipage}[b]{.5\linewidth} \centering \includegraphics[width=10cm]{Fig_Evidence.jpg} \caption{Image} 
% \end{minipage}% 

% \end{figure}

\section{Evidence of Eq.~\ref{Eq:SuDA_param_registration}}
\begin{figure}[ht]
\centering
\includegraphics[width=0.6\linewidth]{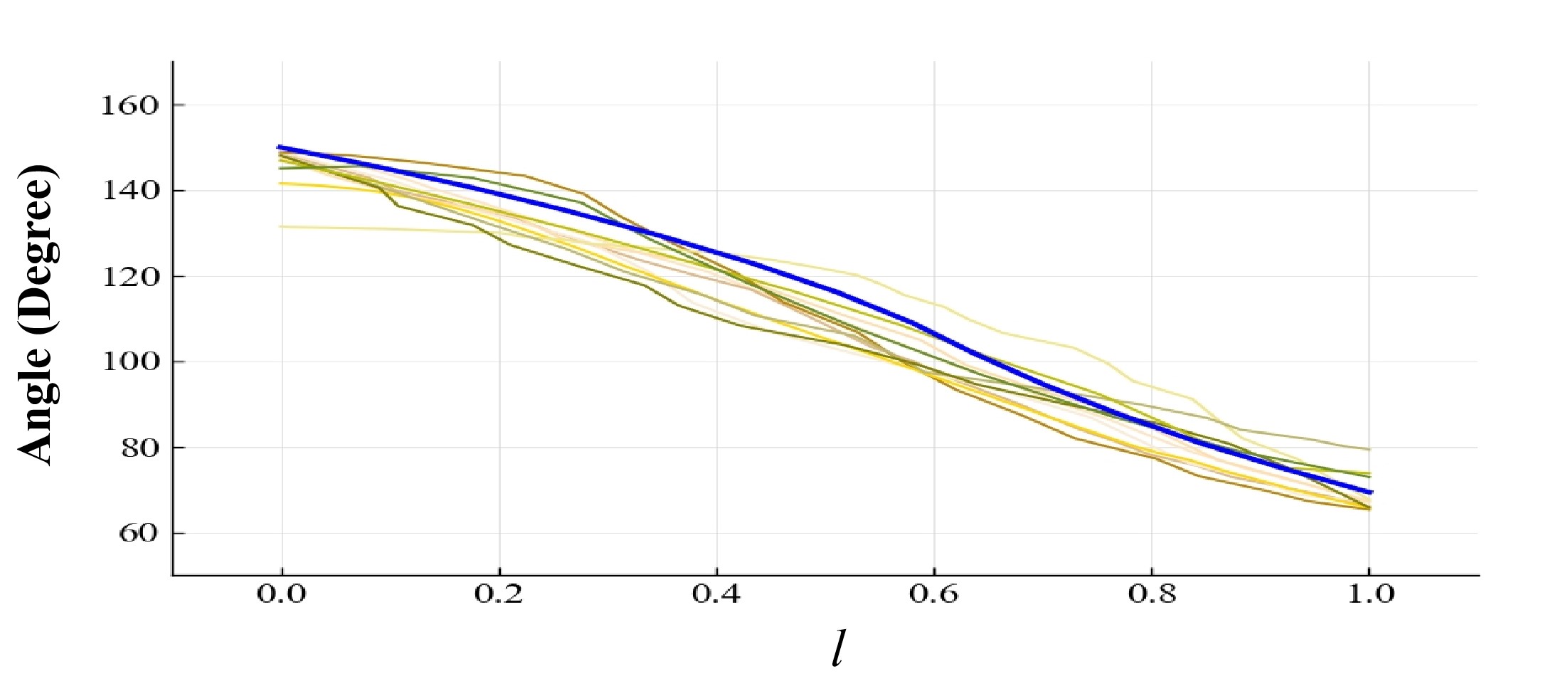}
\caption{The relationship between parameters $l$ and label angle. The blue one is the simulation data and the others are real data samples.}
\label{fig:evidence}
\end{figure}

\begin{equation}
\begin{split}
        \int_{x^a_s}^{x^b_s} \frac{df_s}{dx_s} dx_s &= \int_{x^a_t}^{x^b_t} \frac{df_t}{dx_t} dx_t =\int_{0}^{l} \frac{df_l}{dl} dl
\end{split}
\label{Eq:SuDA_param_registration}
\end{equation}
suggests that data with equal parameter change (\textit{i.e.,} $l - 0$) have the same label change (\textit{i.e.,} $y^b - y^a$), we empirically evaluate
this by conducting parameterization to both simulation (source) and real (target) function supports (Fig.\ref{fig:evidence}).

\section{Illustration of Distribution Variations}
\begin{figure}[t!]
      \centering  
      % \vspace{2pt}
     \begin{subfigure}[b]{0.19\linewidth}
         \centering
        \includegraphics[width=1.2\linewidth]{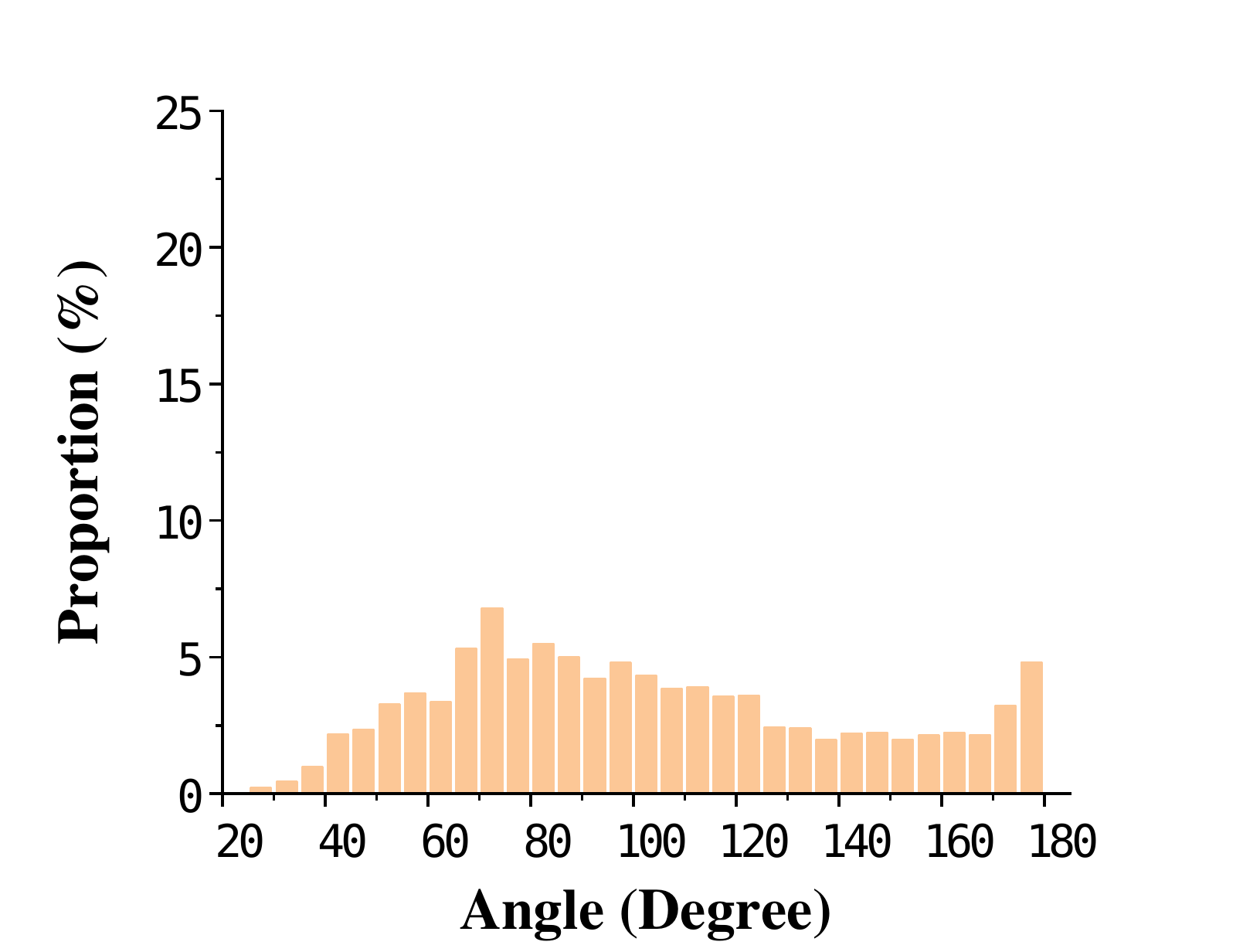}
         \caption{$\mathcal{D}(Simulation)$}
         \label{fig:AngleSim}
     \end{subfigure}
    \begin{subfigure}{0.19\linewidth}
        \centering
        \includegraphics[width=1.2\linewidth]{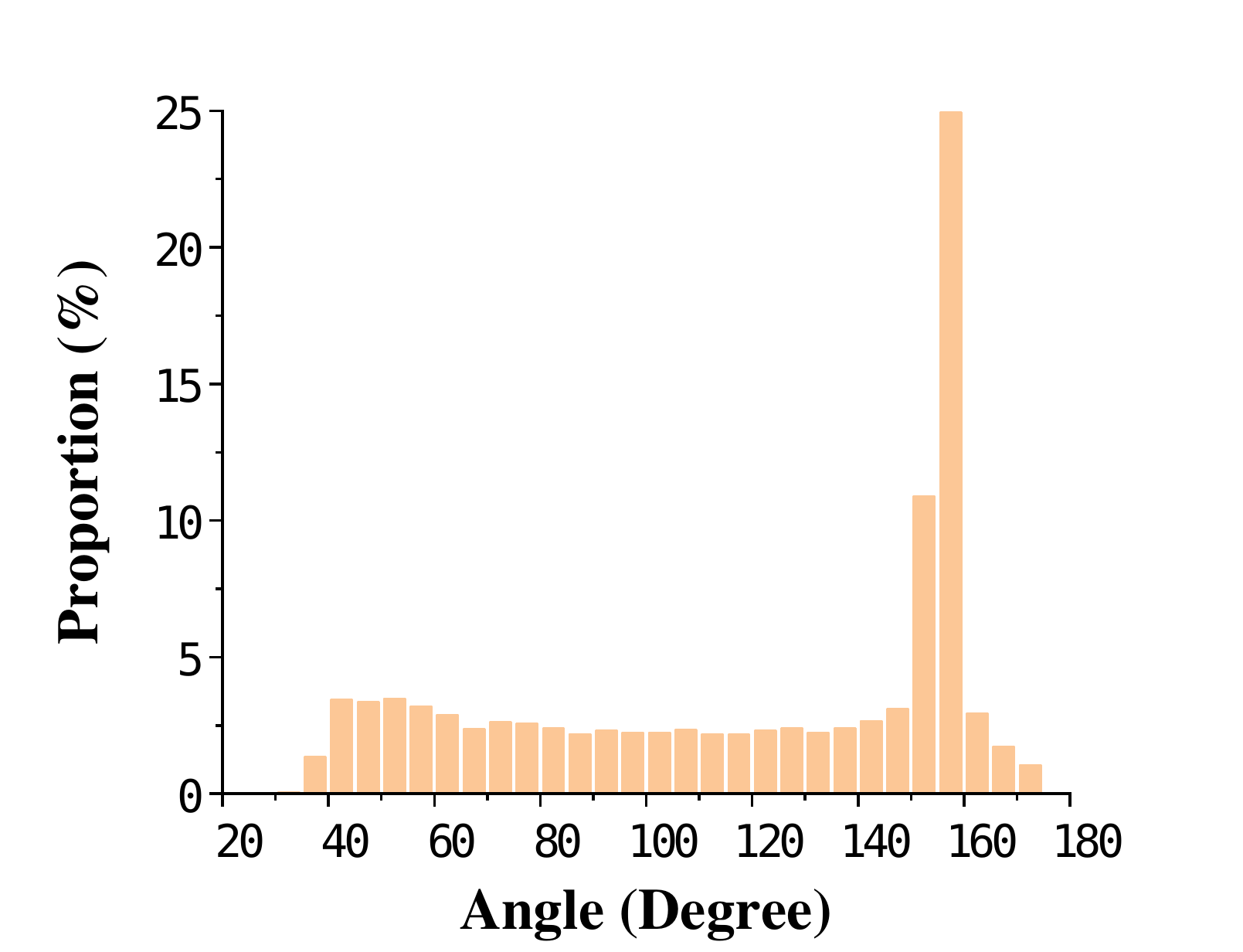}
            % \vspace{-0.57cm} %
        \caption{$\mathcal{D}(id=1)$}
        \label{fig:AngleUser}
    \end{subfigure}
    \begin{subfigure}{0.19\linewidth}
		\centering
		\includegraphics[width=1.2\linewidth]{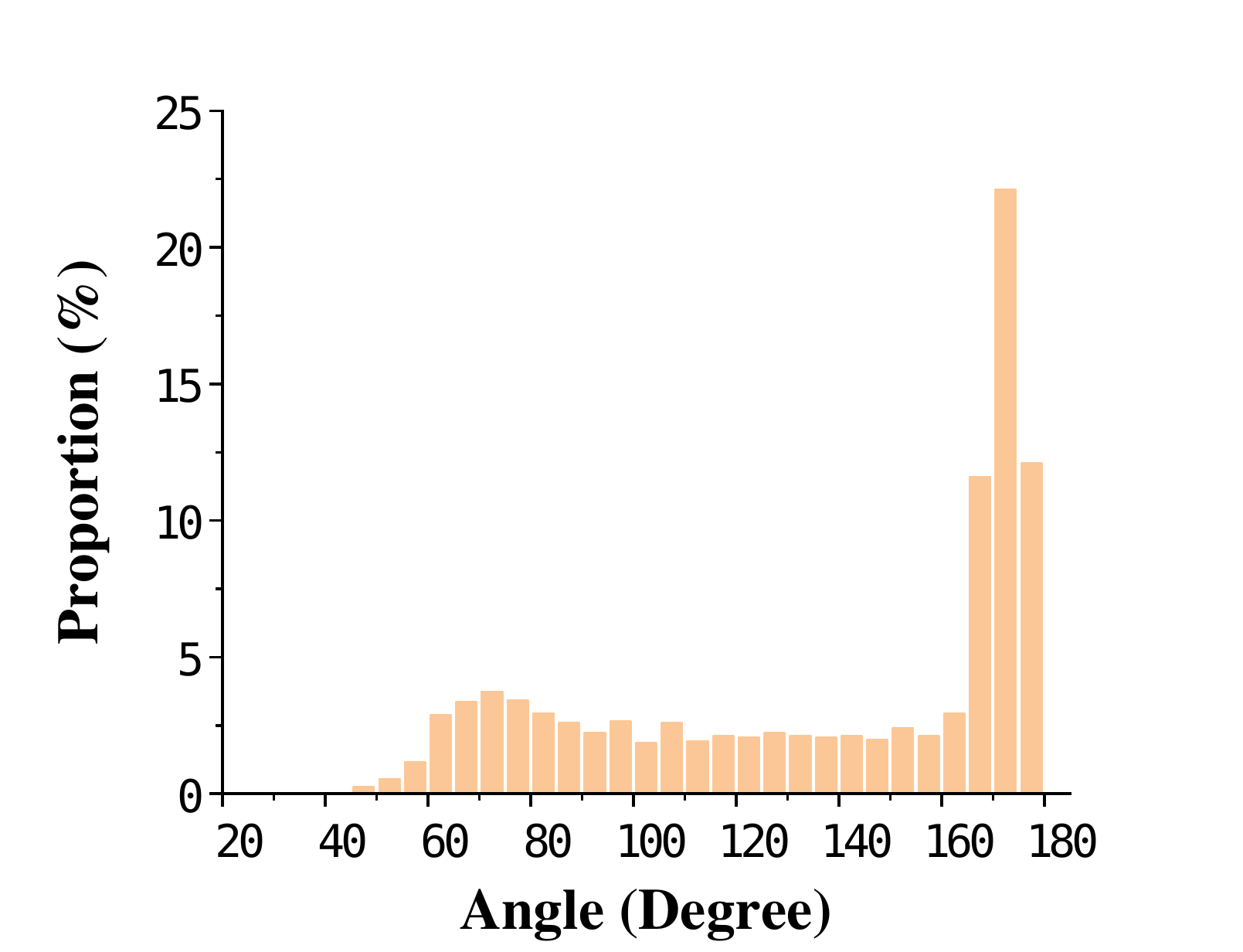}
        \caption{ $\mathcal{D}(wp=180^\circ)$}
		\label{fig:AnglePosition}%文中引用该图片代号
	\end{subfigure}
    \begin{subfigure}{0.19\linewidth}
		\centering
		\includegraphics[width=1.2\linewidth]{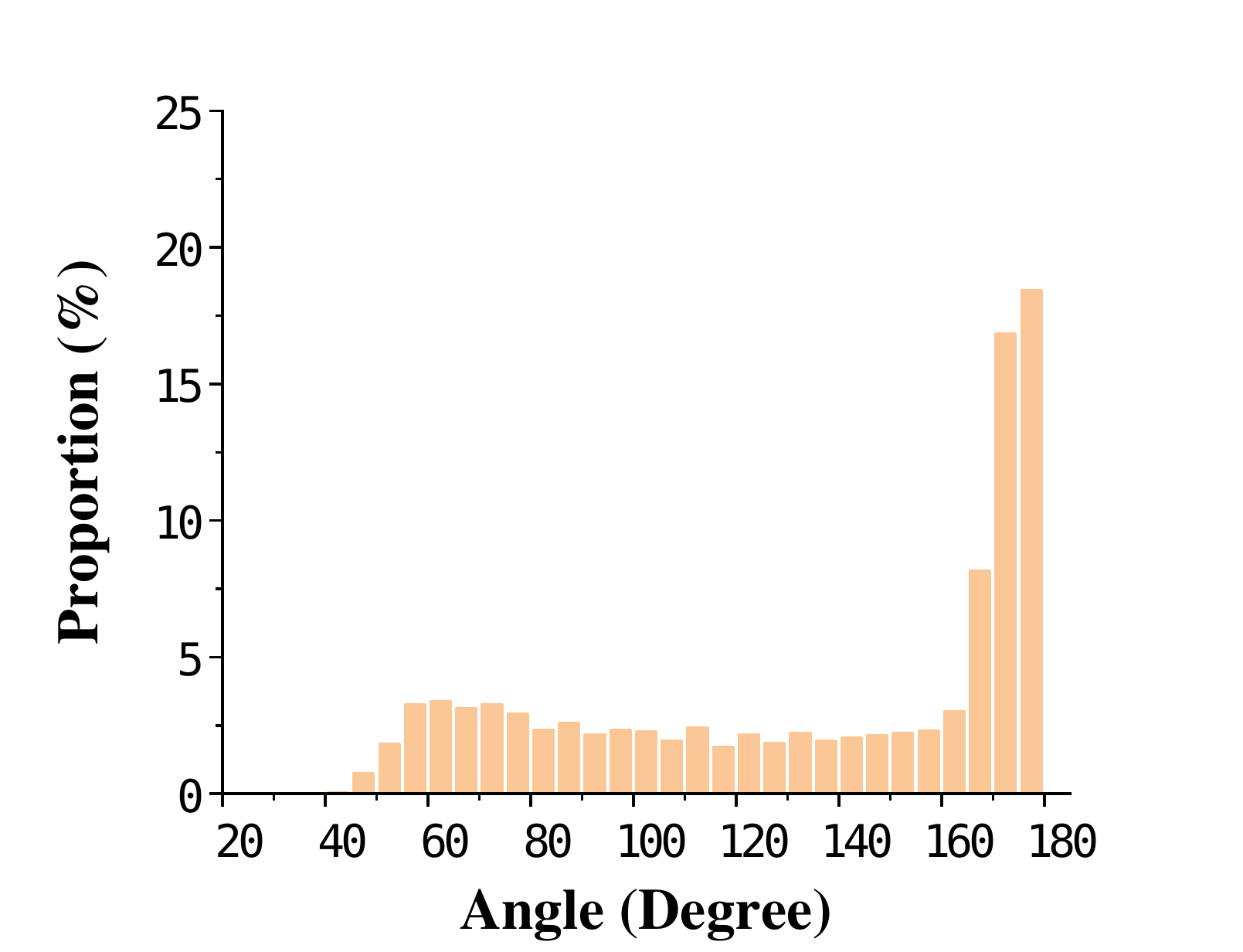}
        \caption{ $\mathcal{D}(m = walk)$}
		\label{fig:AngleMotion}%文中引用该图片代号
	\end{subfigure}
      \begin{subfigure}[b]{0.19\linewidth}
         \centering
        \includegraphics[width=1.2\linewidth]{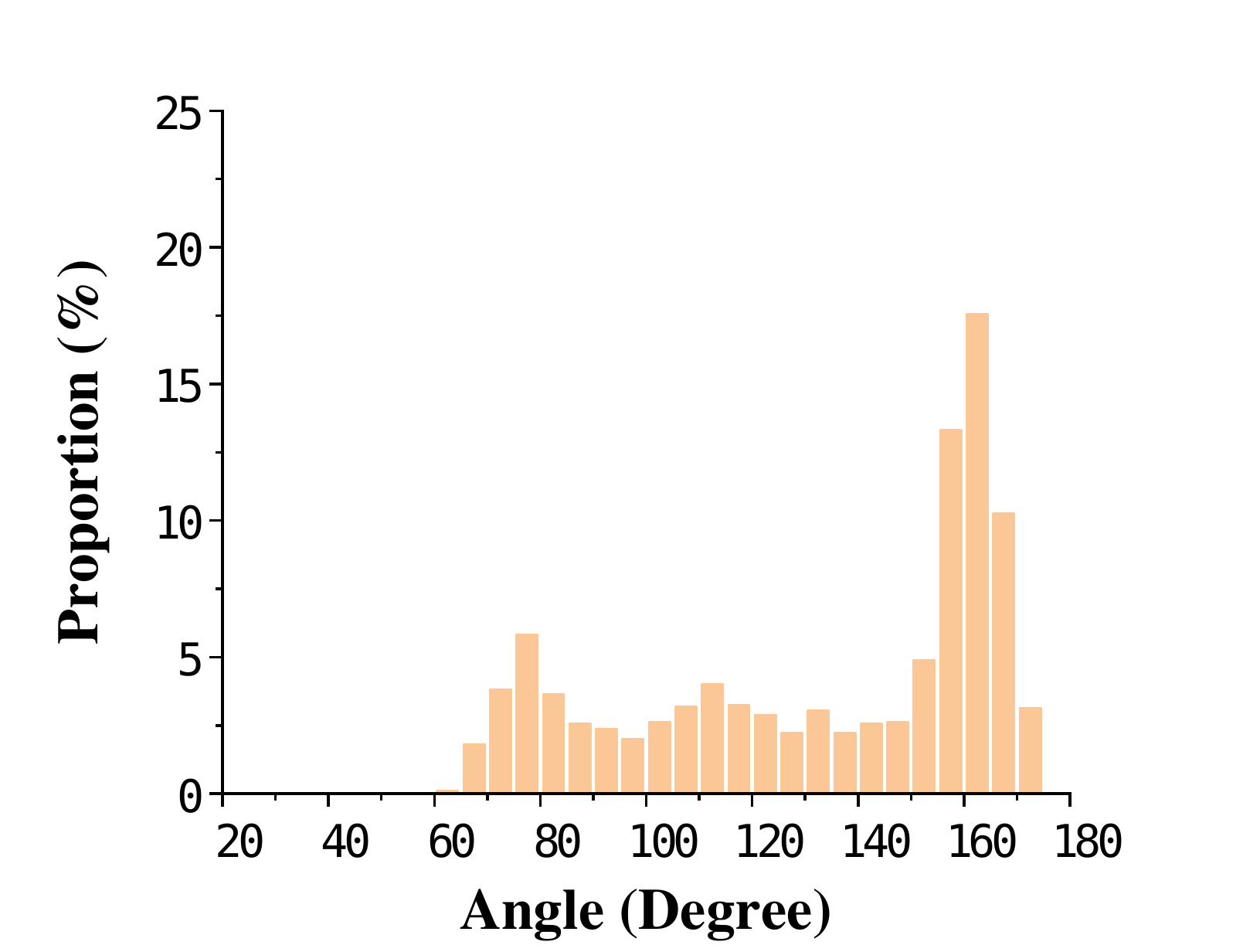}
         % \caption{$D(Right Knee)$}
                          \caption{$\mathcal{D}(Right Knee)$}
         \label{fig:AngleJoints}
     \end{subfigure}
     
    \begin{subfigure}{0.19\linewidth}
        \centering
        \includegraphics[width=1.2\linewidth]{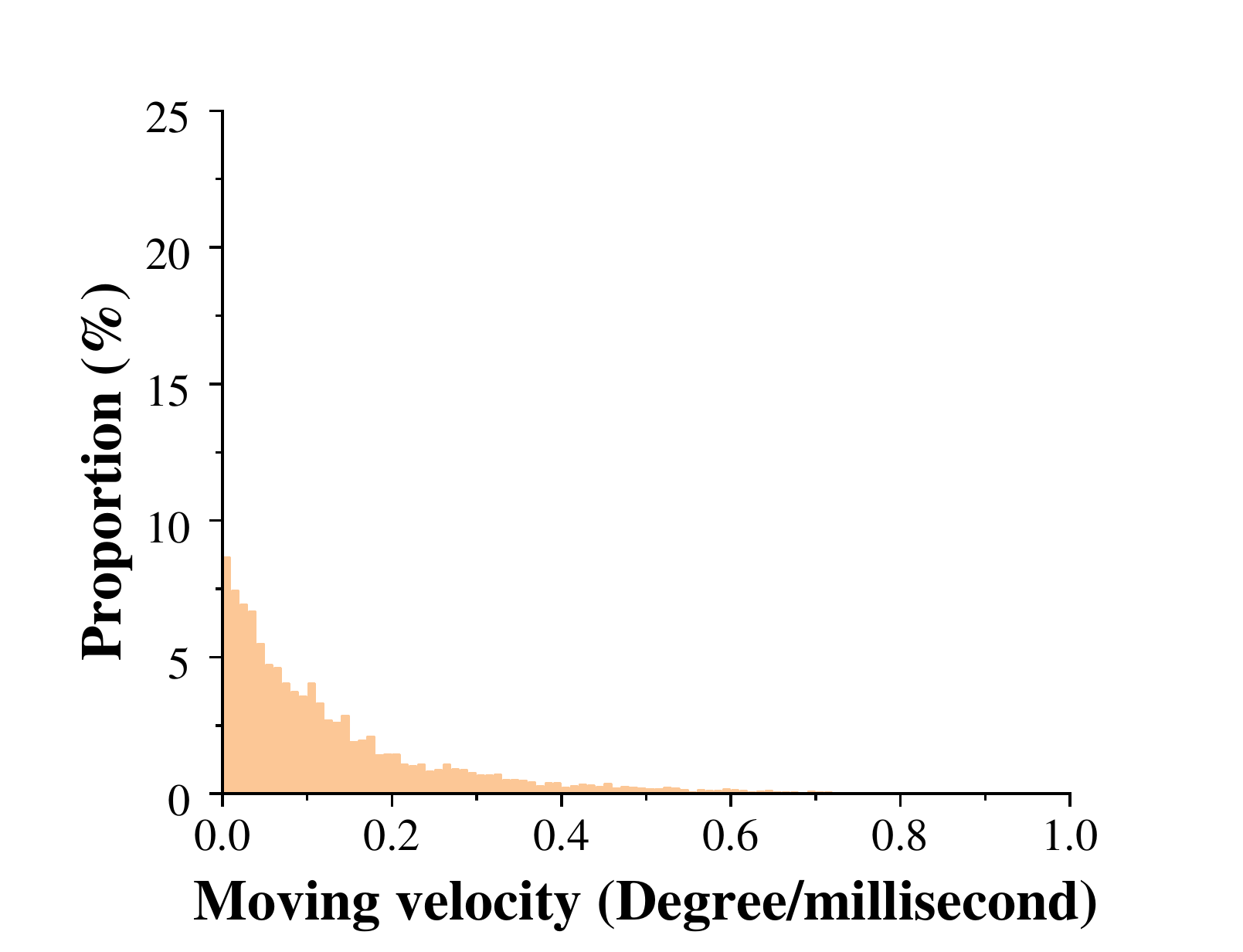}
            % \vspace{-0.57cm} %
         \caption{$\mathcal{D}(Simulation)$}
        \label{fig:SpeedSim}
    \end{subfigure}
    \begin{subfigure}{0.19\linewidth}
		\centering
		\includegraphics[width=1.2\linewidth]{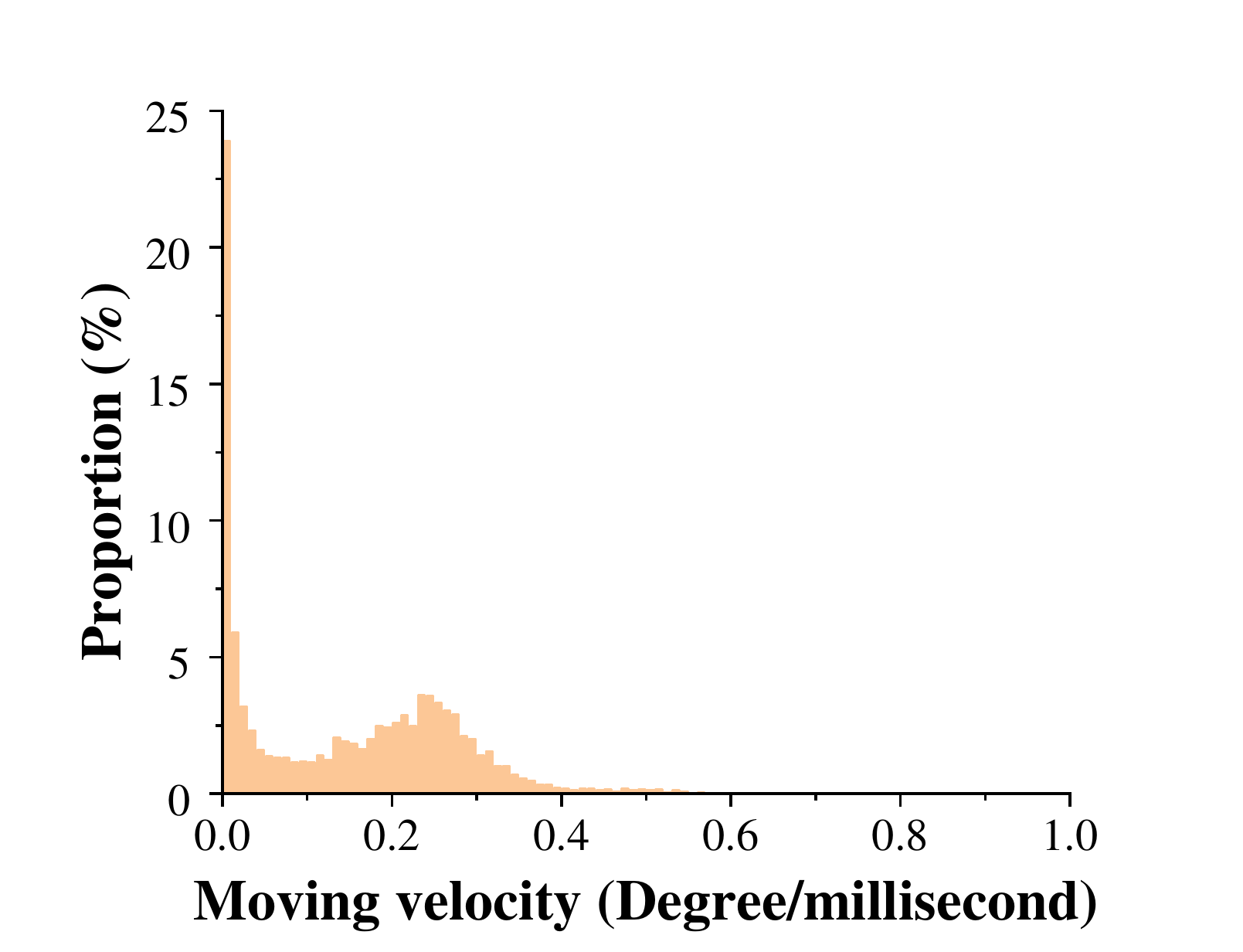}
        \caption{$\mathcal{D}(id=1)$}
		\label{fig:SpeedUser}%文中引用该图片代号
	\end{subfigure}
    \begin{subfigure}{0.19\linewidth}
		\centering
		\includegraphics[width=1.2\linewidth]{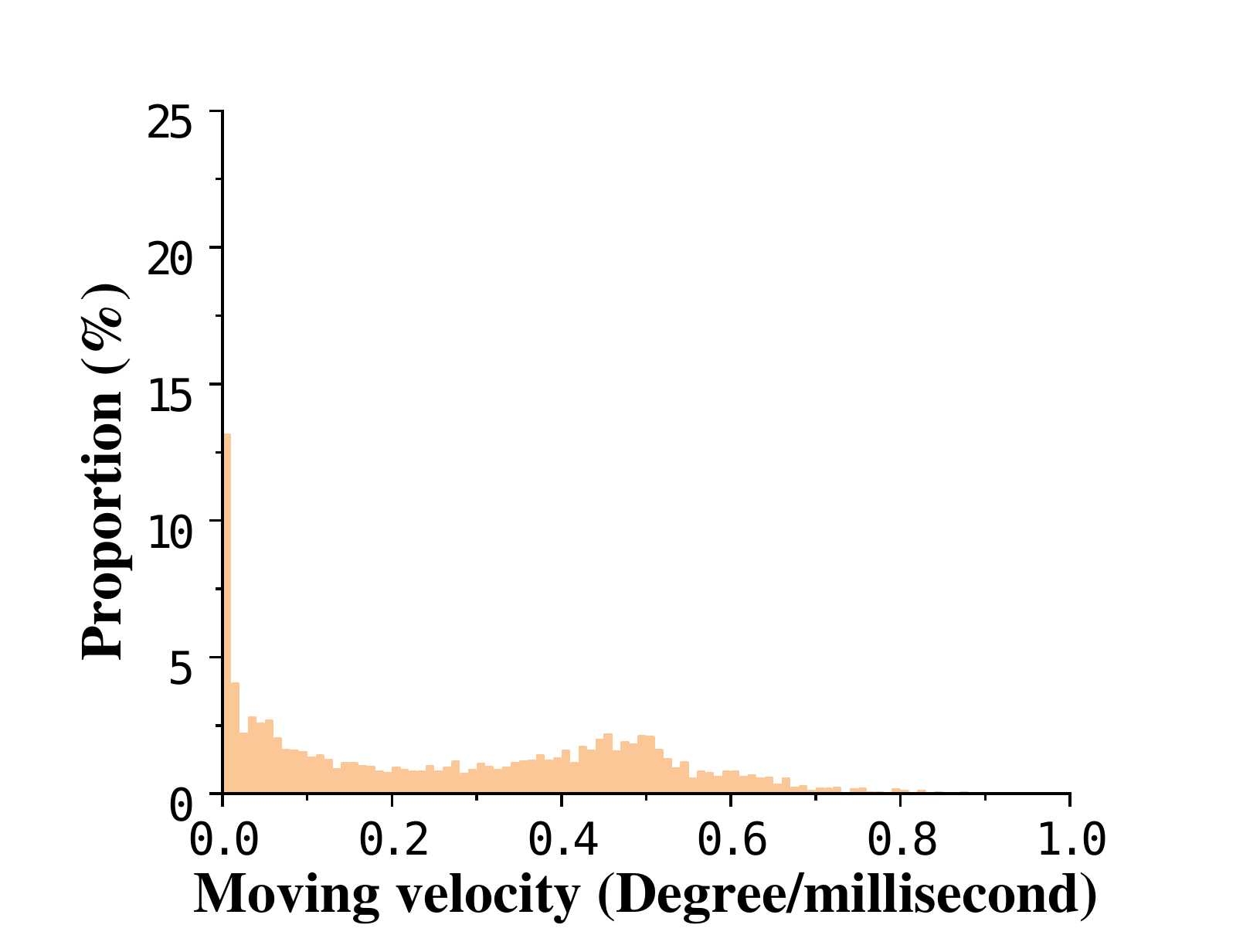}
        \caption{ $\mathcal{D}(wp=180^\circ)$}
		\label{fig:SpeedPosition}%文中引用该图片代号
	\end{subfigure}
     \begin{subfigure}{0.19\linewidth}
		\centering
		\includegraphics[width=1.2\linewidth]{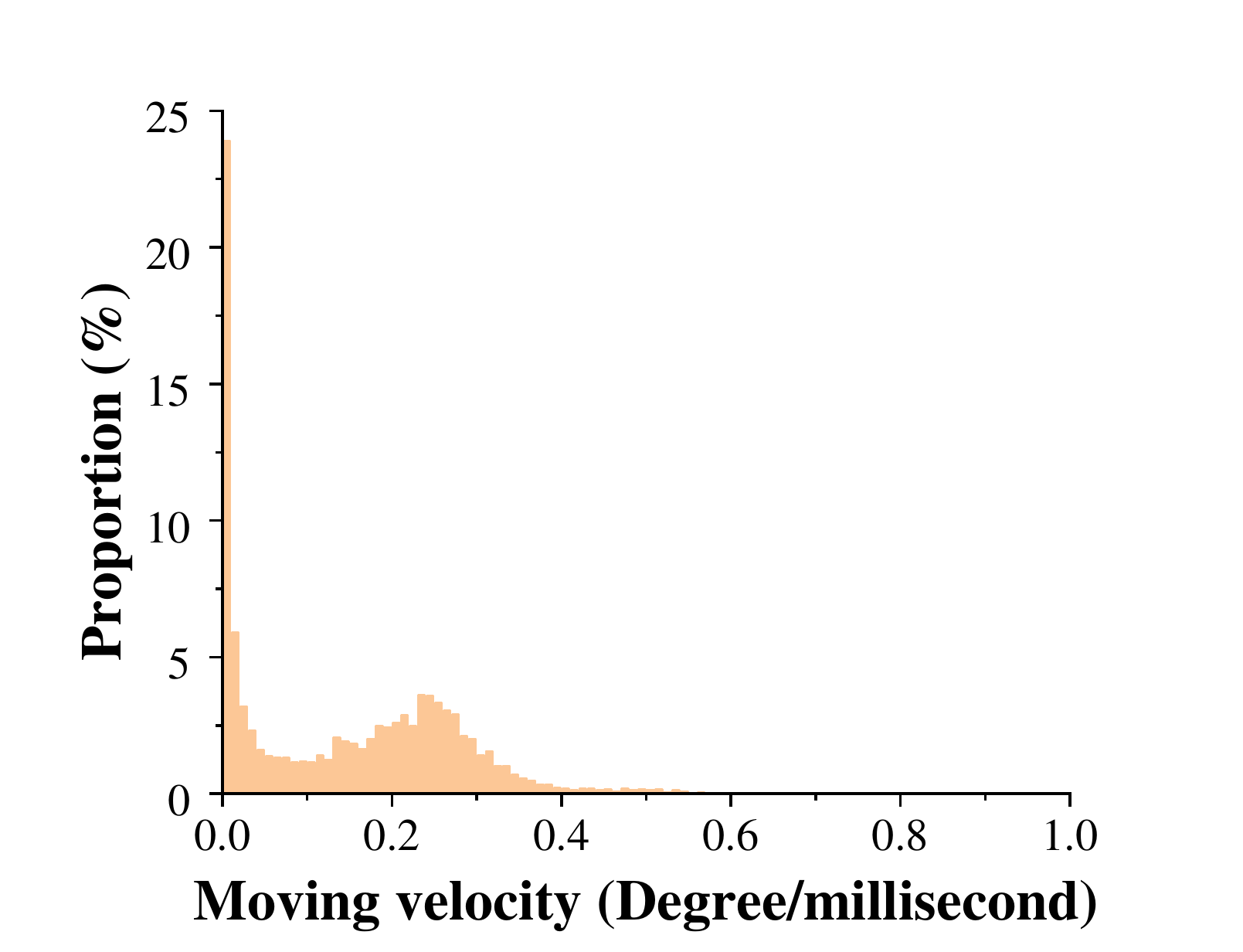}
        \caption{ $\mathcal{D}(m = walk)$}
		\label{fig:SpeedMotion}%文中引用该图片代号
	\end{subfigure}
    \begin{subfigure}{0.19\linewidth}
		\centering
		\includegraphics[width=1.2\linewidth]{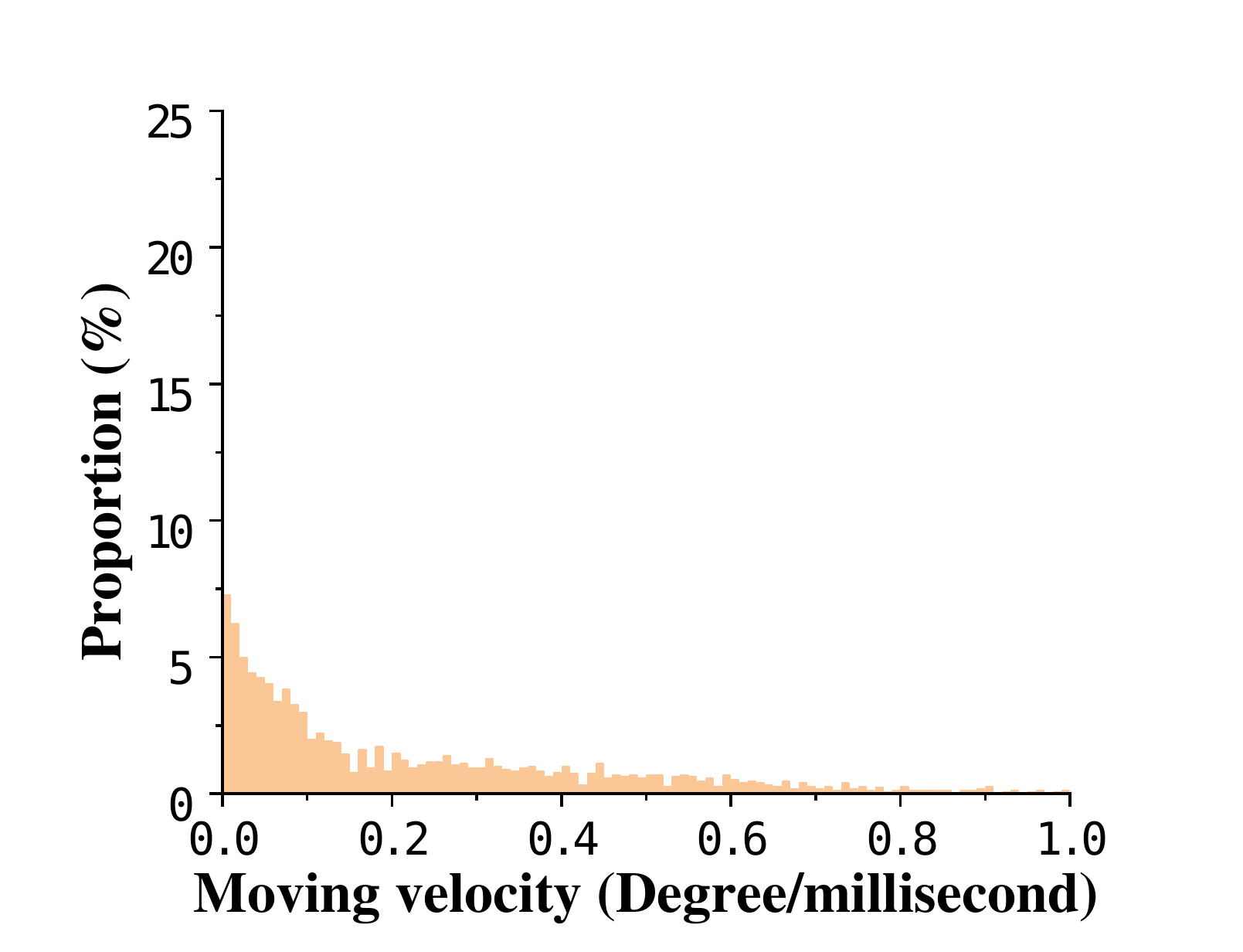}
        % \caption{$D(Right Knee)$}
                         \caption{$\mathcal{D}(Right Knee)$}
		\label{fig:SpeedJoints}%文中引用该图片代号
	\end{subfigure}
 
     \begin{subfigure}{0.19\linewidth}
		\centering
		\includegraphics[width=1.2\linewidth]{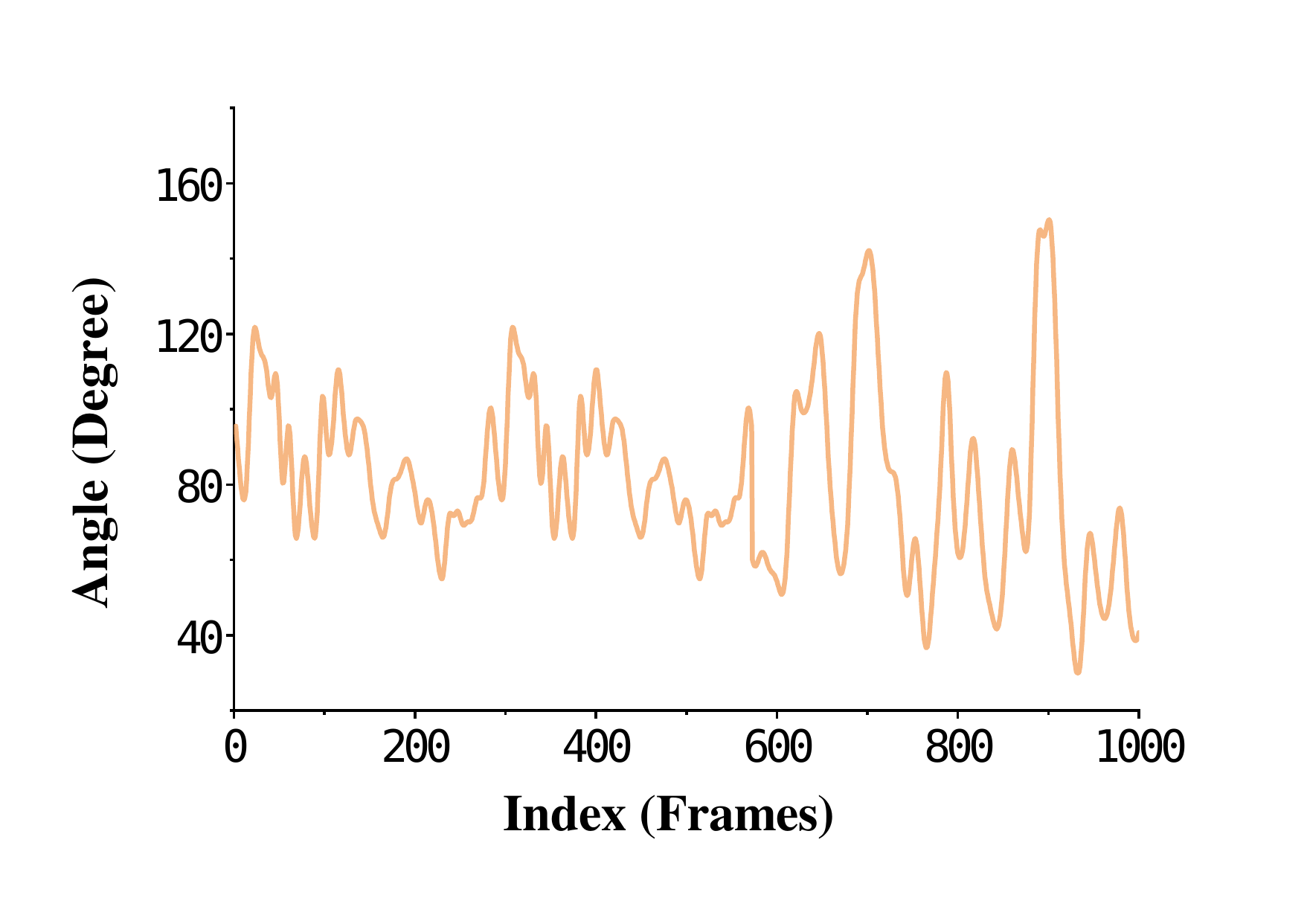}
         \caption{$\mathcal{D}(Simulation)$}
		\label{fig:TimeSim}%文中引用该图片代号
	\end{subfigure}
    \begin{subfigure}{0.19\linewidth}
		\centering
		\includegraphics[width=1.2\linewidth]{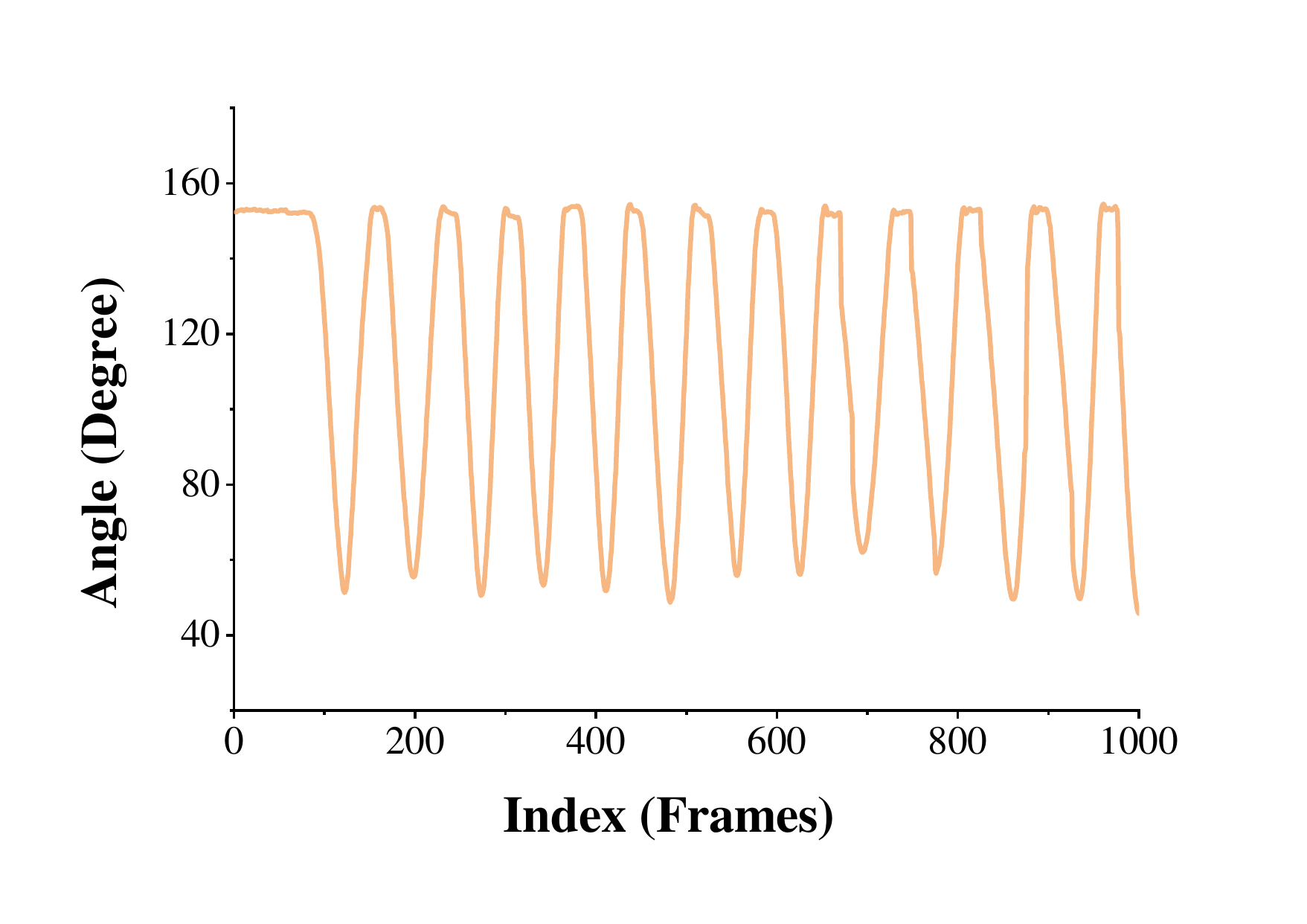}
        \caption{$\mathcal{D}(id=1)$}
		\label{fig: TimeUser}%文中引用该图片代号
	\end{subfigure}
     \begin{subfigure}{0.19\linewidth}
		\centering
		\includegraphics[width=1.2\linewidth]{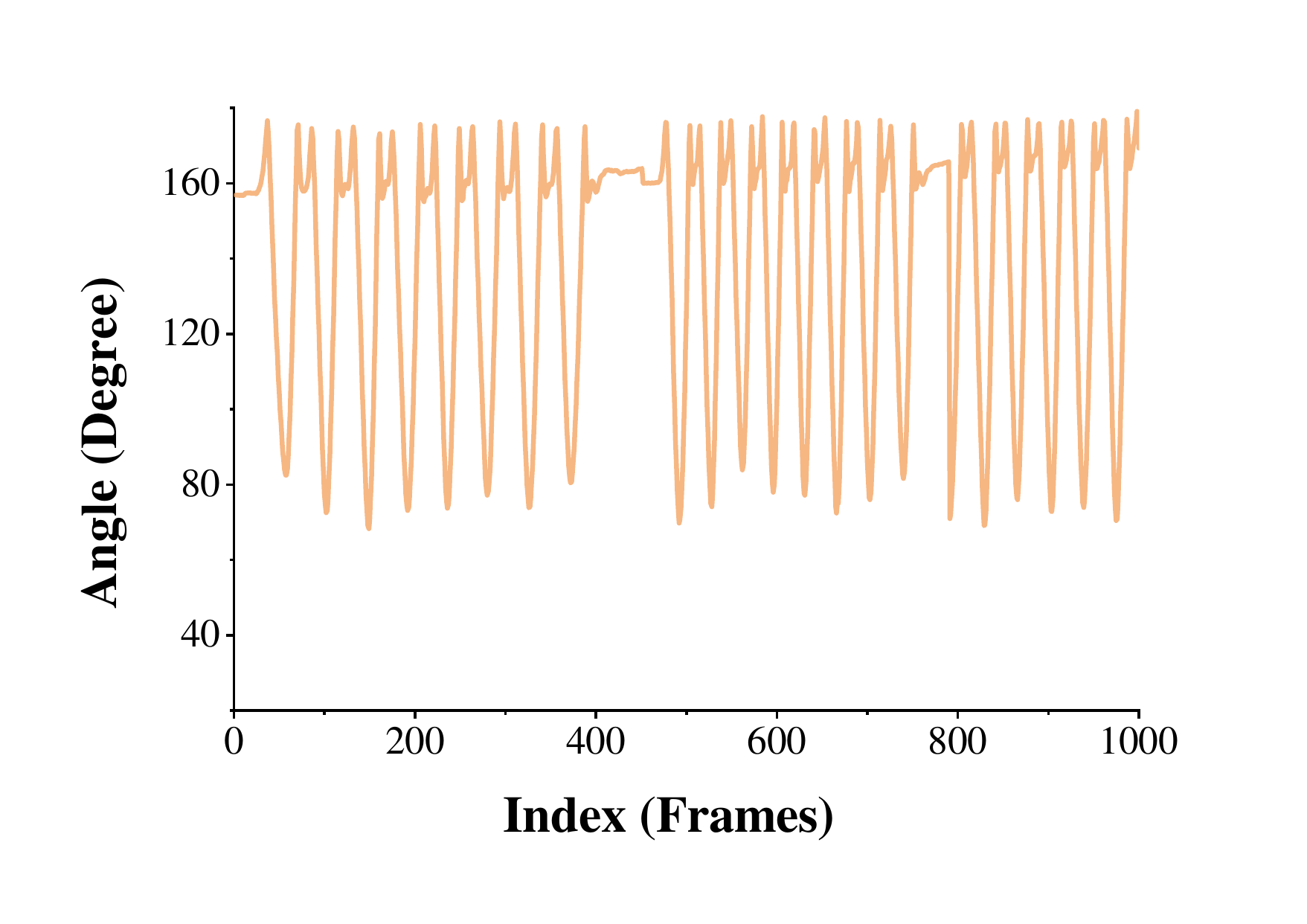}
        \caption{$\mathcal{D}(wp=180^\circ)$}
		\label{fig:TimePosition}%文中引用该图片代号
	\end{subfigure}
    \begin{subfigure}{0.19\linewidth}
		\centering
		\includegraphics[width=1.2\linewidth]{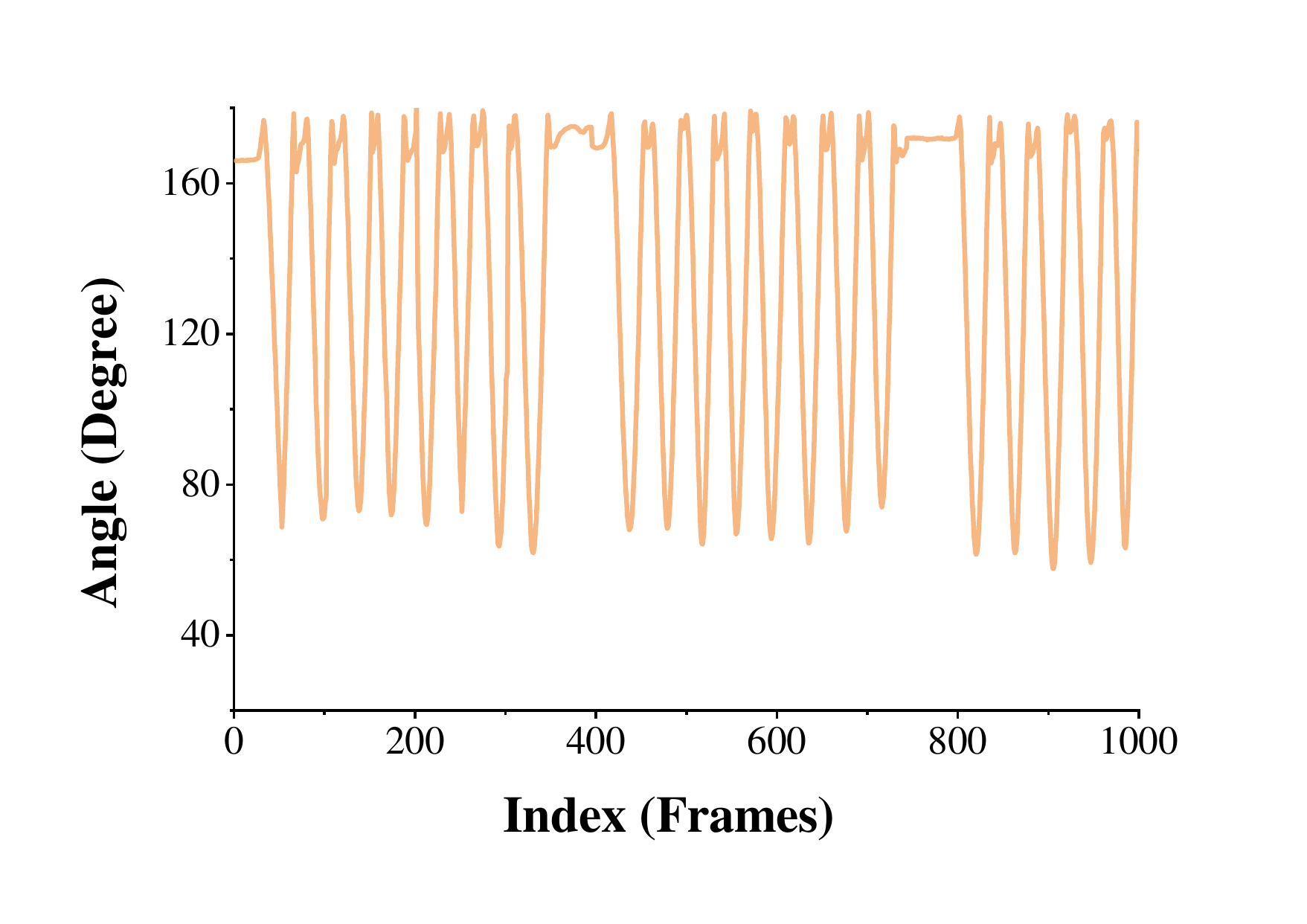}
        \caption{$\mathcal{D}(m = walk)$}
		\label{fig:TimeMotion}%文中引用该图片代号
	\end{subfigure}
     \begin{subfigure}{0.19\linewidth}
		\centering
		\includegraphics[width=1.2\linewidth]{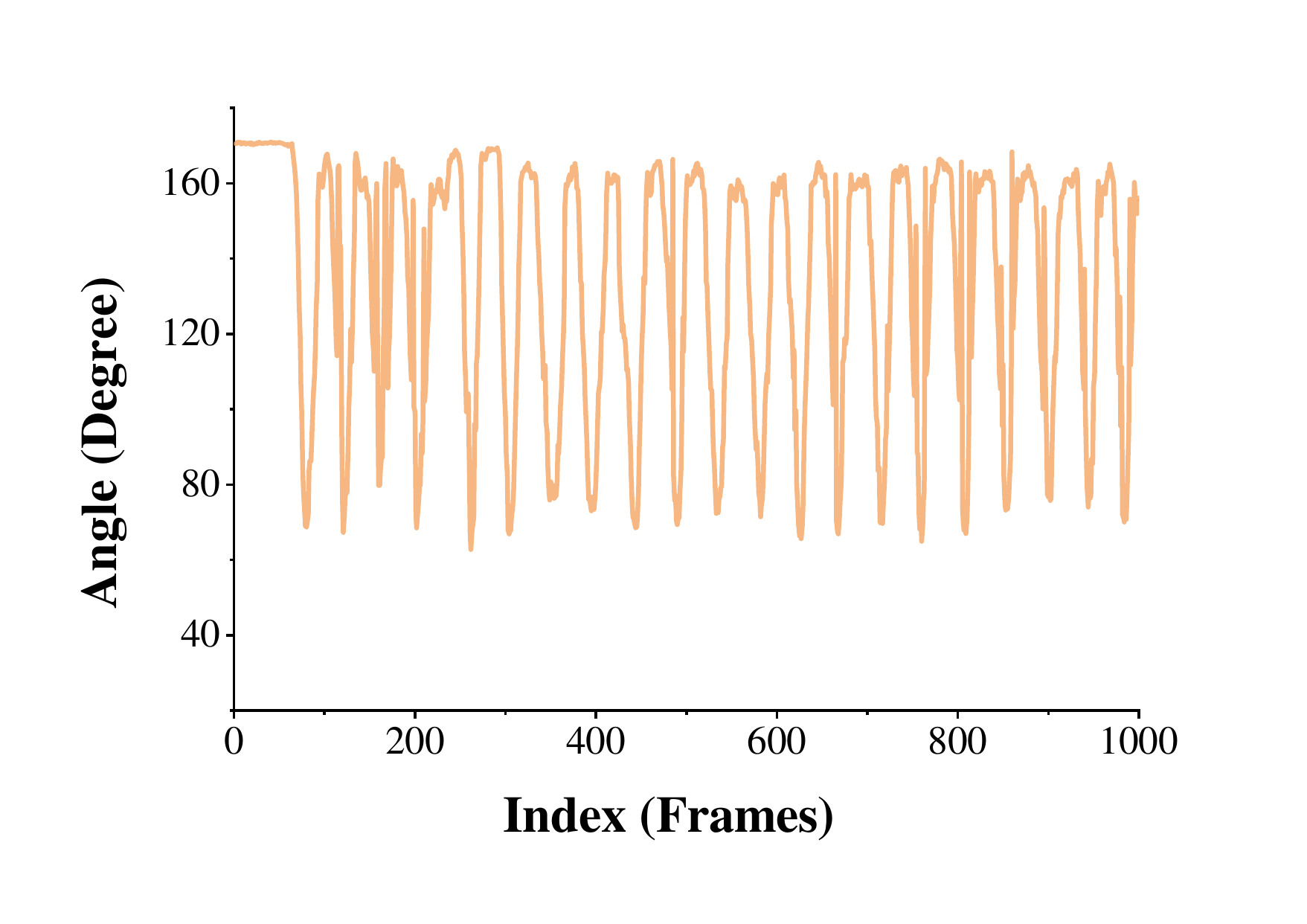}
        % \caption{$D(Right Knee)$}
                         \caption{$\mathcal{D}(Right Knee)$}
		\label{fig:TimeJoints}%文中引用该图片代号
	\end{subfigure}

     \begin{subfigure}{0.19\linewidth}
		\centering
		\includegraphics[width=1.2\linewidth]{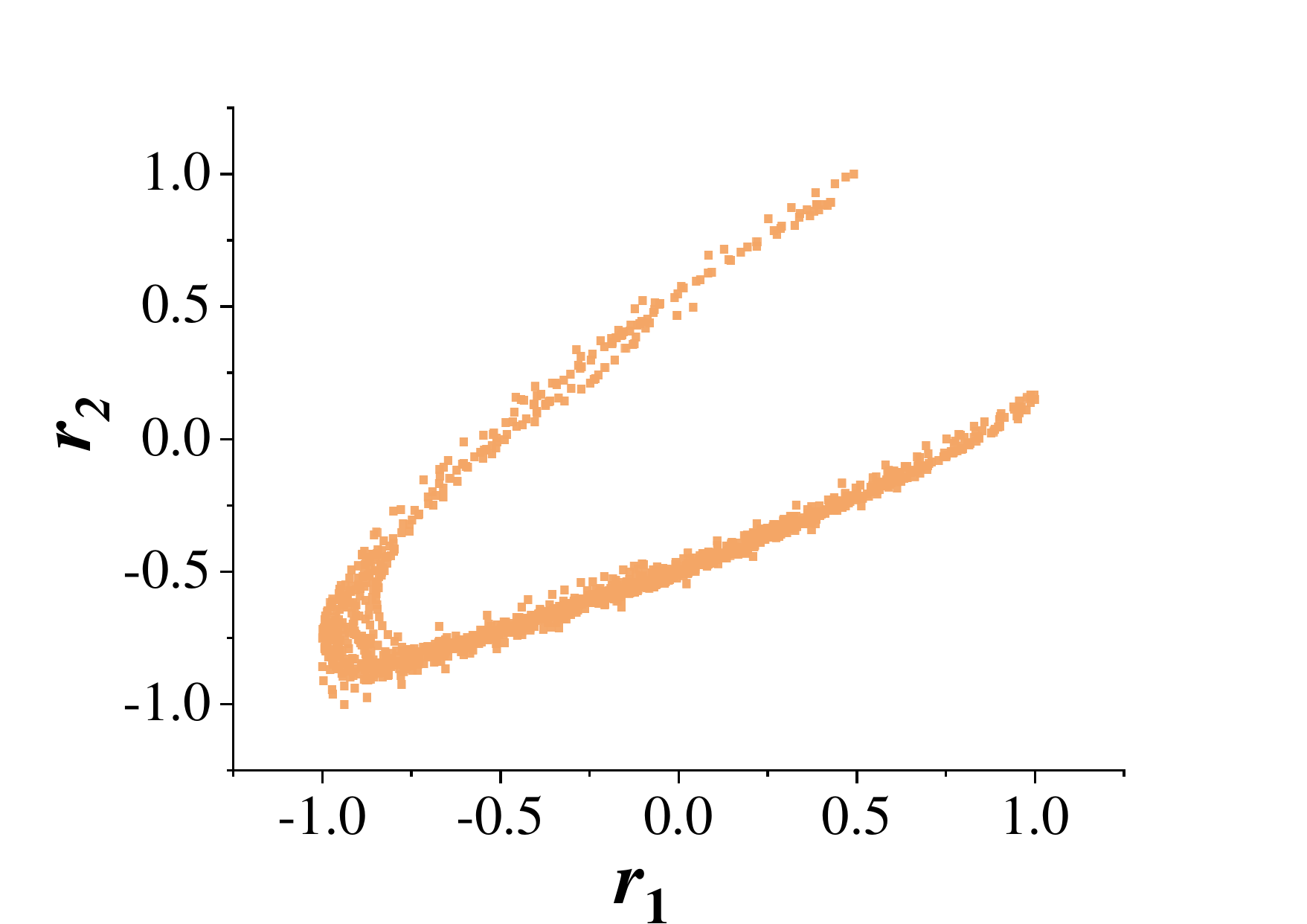}
         \caption{$\mathcal{D}(Simulation)$}
		\label{fig:SupportSim}%文中引用该图片代号
	\end{subfigure}
    \begin{subfigure}{0.19\linewidth}
		\centering
		\includegraphics[width=1.2\linewidth]{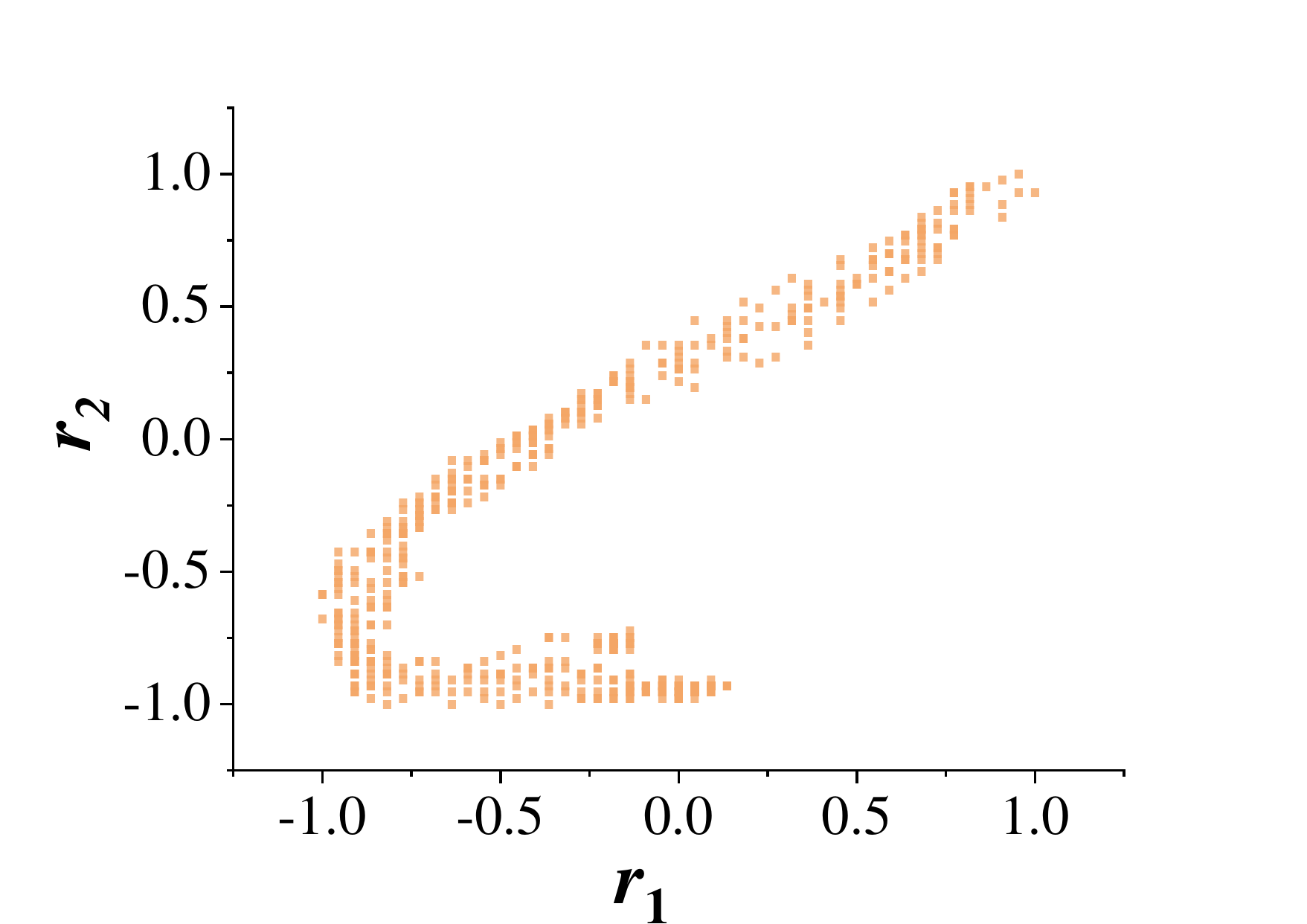}
        \caption{$\mathcal{D}(id=1)$}
		\label{fig:SupportUser}%文中引用该图片代号
	\end{subfigure}
     \begin{subfigure}{0.19\linewidth}
		\centering
		\includegraphics[width=1.2\linewidth]{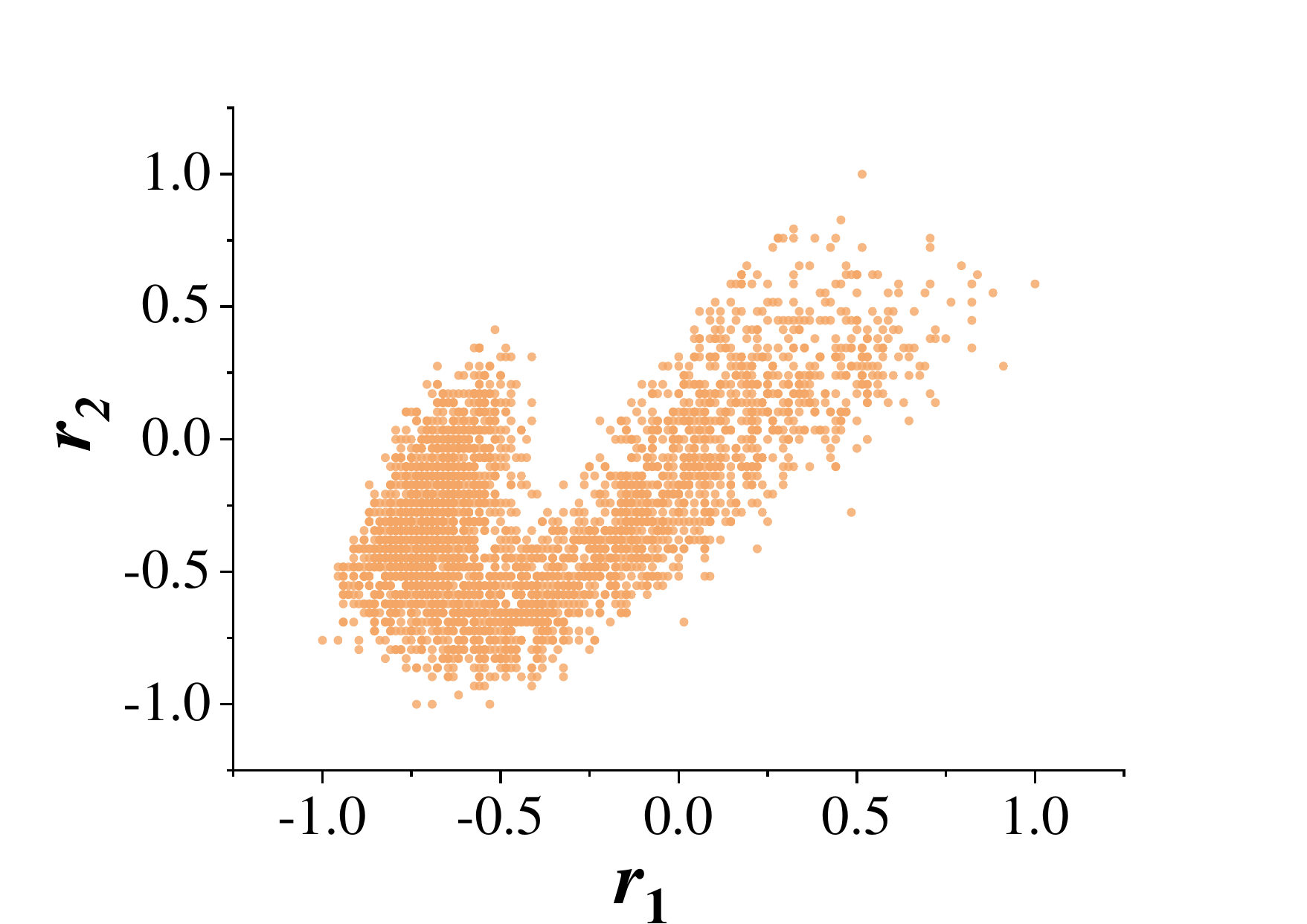}
        \caption{$\mathcal{D}(wp=180^\circ)$}
		\label{fig:SupportPosition}%文中引用该图片代号
	\end{subfigure}
    \begin{subfigure}{0.19\linewidth}
		\centering
		\includegraphics[width=1.2\linewidth]{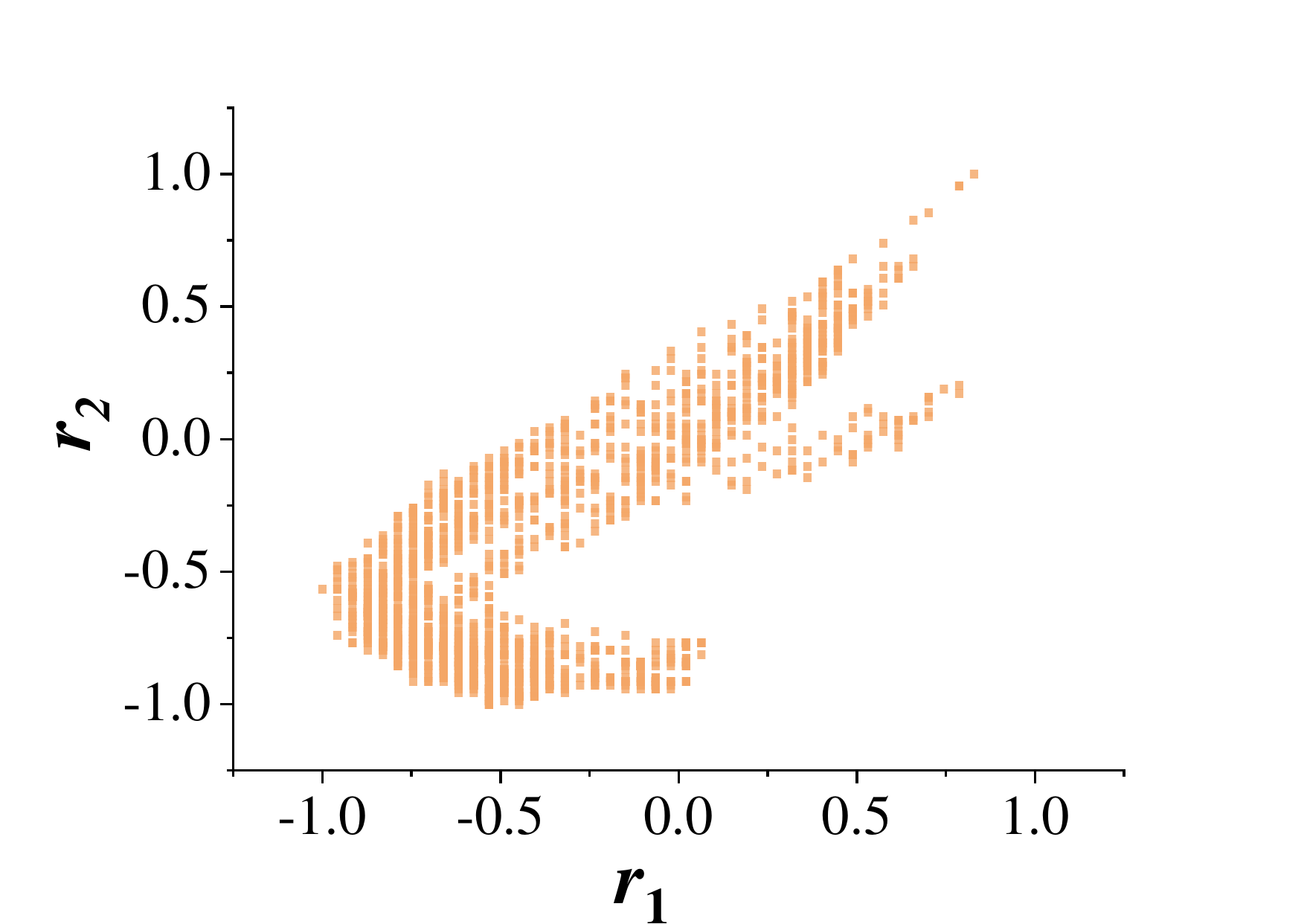}
        \caption{$\mathcal{D}(m = walk)$}
		\label{fig:SupportMotion}%文中引用该图片代号
	\end{subfigure}
     \begin{subfigure}{0.19\linewidth}
		\centering
		\includegraphics[width=1.2\linewidth]{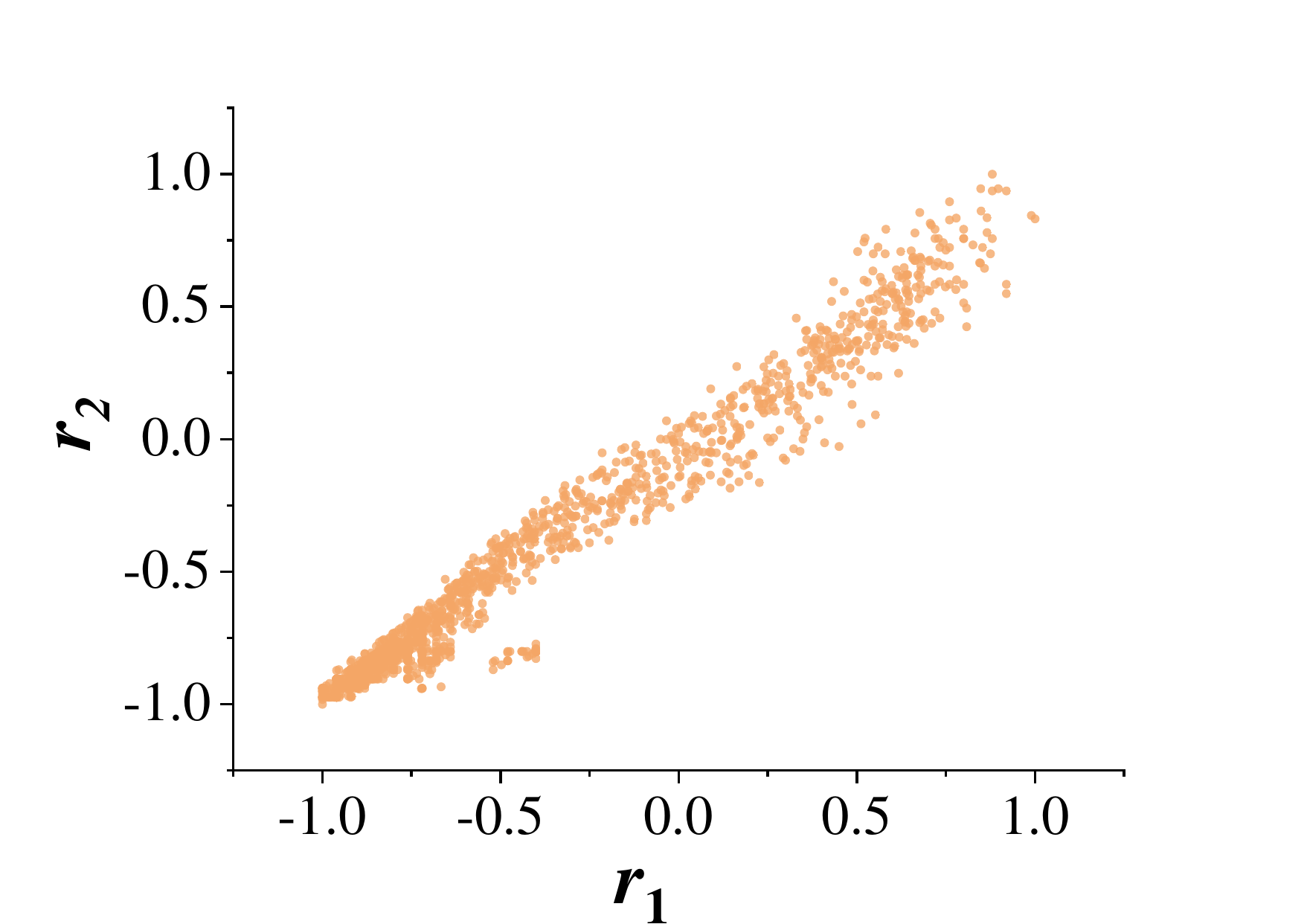}
        % \caption{$D(Right Knee)$}
                 \caption{$\mathcal{D}(Right Knee)$}
		\label{fig:SupportJoints}%文中引用该图片代号
	\end{subfigure}
    \caption{ (a) (b) (c) (d) (e) \textit{Label (Angle) proportion} of Simulation, User, Position, Motion, and Joint data Samples;
    (f) (g) (h) (i) (j) \textit{Moving Velocity} of Simulation, User, Position, Motion, and Joint data Samples; (k) (l) (m) (n) (o) \textit{Time Sequence} of Simulation, User, Position, Motion, and Joint data Samples;  (p) (q) (r) (s) (t) Function Supports (sensor readings) of Simulation, User, Position, Motion, and Joint data Samples;  
    }
    % \vspace{-15pt}
\end{figure}

From (a) to (e), we observe commonalities in simulation data: they all share similar label ranges. However, the distinction lies in the distribution of labels. Simulation data exhibits a relatively uniform distribution across labels, with a higher concentration within a smaller label range (60 degrees to 100 degrees). In contrast, real users, positional motions, and joint movements tend to be more prevalent in a broader label range (160 degrees to 180 degrees), constituting over 30$\%$. This discrepancy may be attributed to the increased occurrence of elbow extension movements in real-life scenarios.

When examining the moving velocity (from (f) to (j)), it is evident that the proportion of simulation data velocities gradually decreases from (0-0.4) (from 10 $\%$ to 0 $\%$). In contrast, real data for users, positions, and motions experiences a decline followed by a sudden increase. This phenomenon may be attributed to the fact that the range of 0.2 to 0.6 (degrees/microsecond) corresponds to the natural movement speed of the human body, thus constituting a higher proportion.

The time-sequence plots from (k) to (o) reveal a notable mismatch between simulation data and all real data. In fact, in practical usage scenarios, it is challenging to identify two action sequences with entirely matching time sequences between virtual and real data. This substantial disparity in distribution between virtual and real data is a crucial factor contributing to this discrepancy.

In (p) through (t), the support distribution plots are presented. 1. Real data exhibits a more pronounced dispersion compared to simulation data, attributed to inherent biases or noise introduced during the sensor-based real data acquisition process. 2. The shapes of the support distributions for simulation and real data differ. SuDA assumes identical $ \frac{df}{dx}$ , aligning supports to achieve successful migration, offering a mechanism to address the dissimilarity between the support distributions of simulation and real data.

In conclusion, the data presented above underscores significant disparities in the distribution between simulation data and real data. Moreover, the failure of the DiDA method can be attributed to the low-dimensionality of the data.

% \clearpage
\section{Evaluations of SuDA}
\begin{figure}[h]
      \centering  
      % \vspace{2pt}
     \begin{subfigure}{0.33\linewidth}
         \centering
        \includegraphics[width=0.9\linewidth]{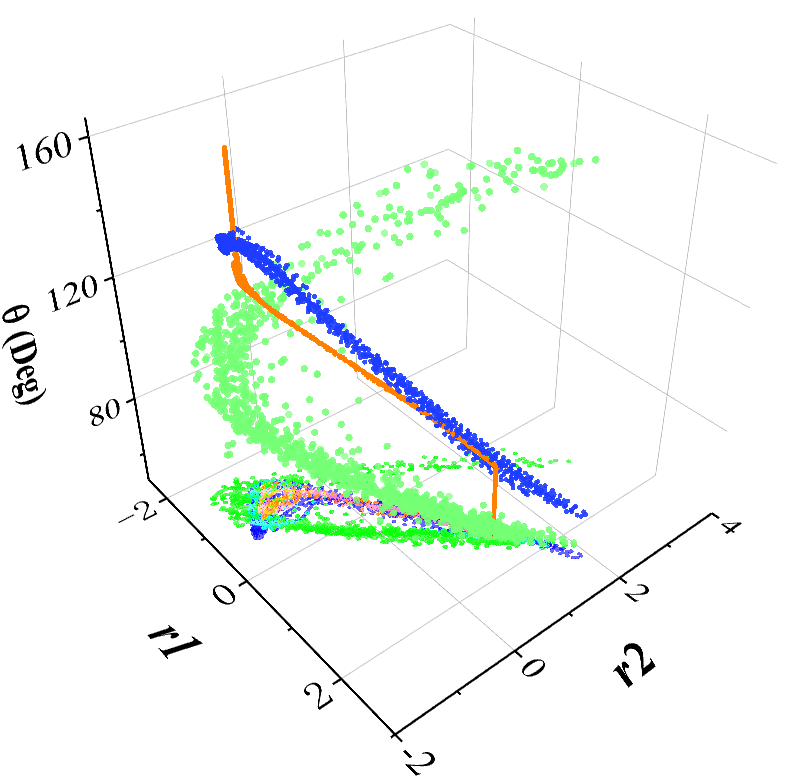}
         \caption{$\mathcal{D}_s$ and $\mathcal{D}(id=1)$ Visualization}
         % \label{fig:AngleSim}
     \end{subfigure}
    \begin{subfigure}{0.33\linewidth}
        \centering
        \includegraphics[width=0.9\linewidth]{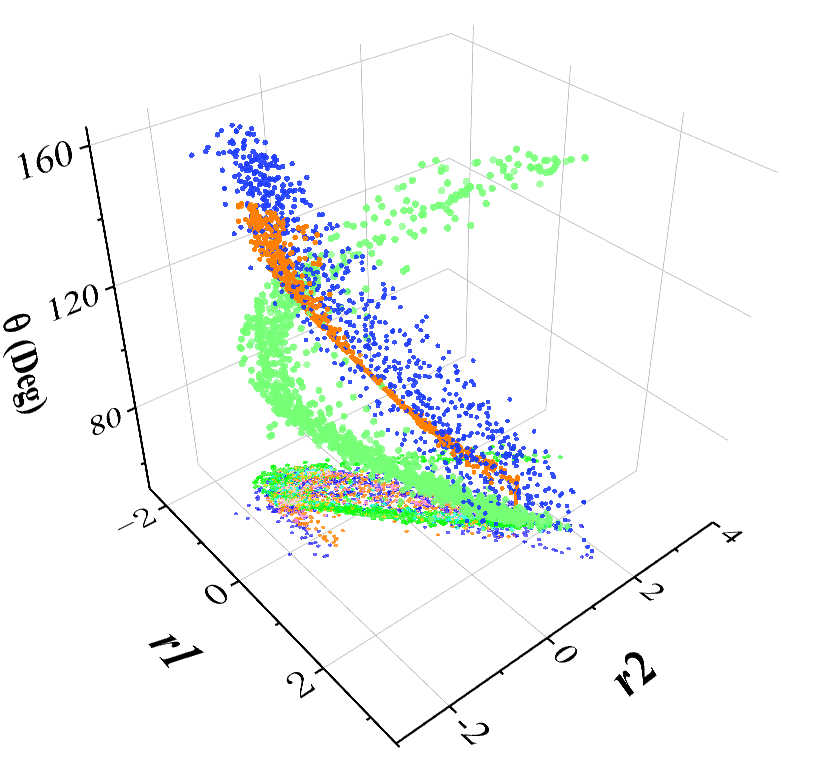}
            % \vspace{-0.57cm} %
        \caption{$\mathcal{D}_s$ and $\mathcal{D}(wp=180^\circ)$ Visualization}
        % \label{fig:AngleUser}
    \end{subfigure}
    \begin{subfigure}{0.33\linewidth}
		\centering
		\includegraphics[width=0.9\linewidth]{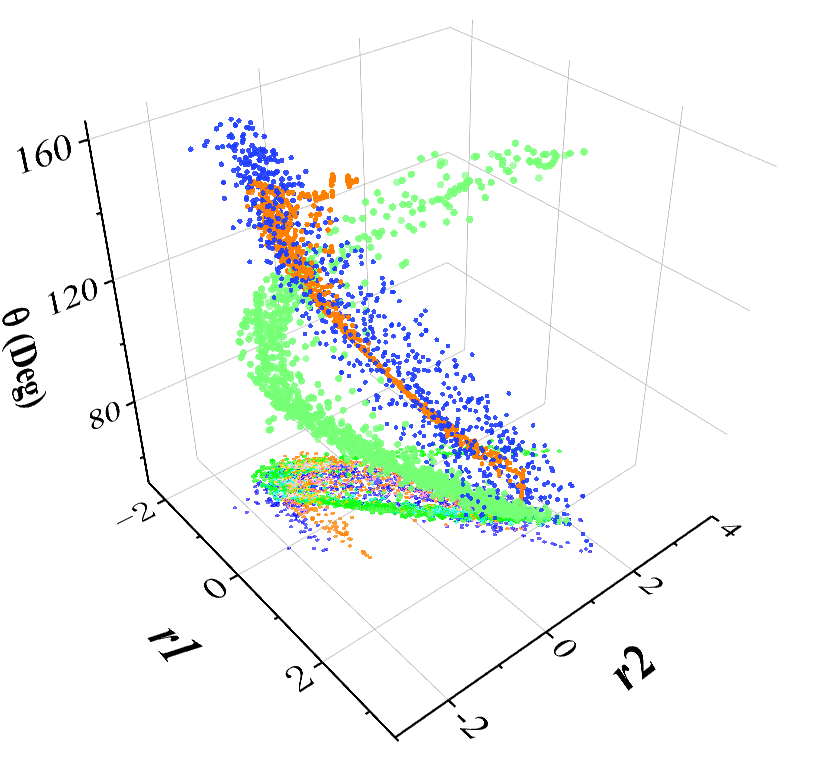}
        \caption{$\mathcal{D}_s$ and $\mathcal{D}(m=run)$ Visualization}
		% \label{fig:AnglePosition}%文中引用该图片代号
	\end{subfigure}
     % \vspace{-5pt}
     \vspace{-8mm}
    \caption{
    Performance Visualization of our method {\it \textbf{SuDA}}. Note that the green points represent source data (Simulation), the blue points represent  target data (Real World), and the orange points represent {\it \textbf{SuDA}} registered simulation data. It is evident that {\it \textbf{SuDA}} is capable of aligning source and target data in the feature space by aligning the {\it supports}.
        % \caption{yes}
    }
    % \vspace{-15pt}
    \label{fig:performance visualization}
\end{figure}

\begin{figure*}[h]
\centering
\includegraphics[width=0.9\linewidth]{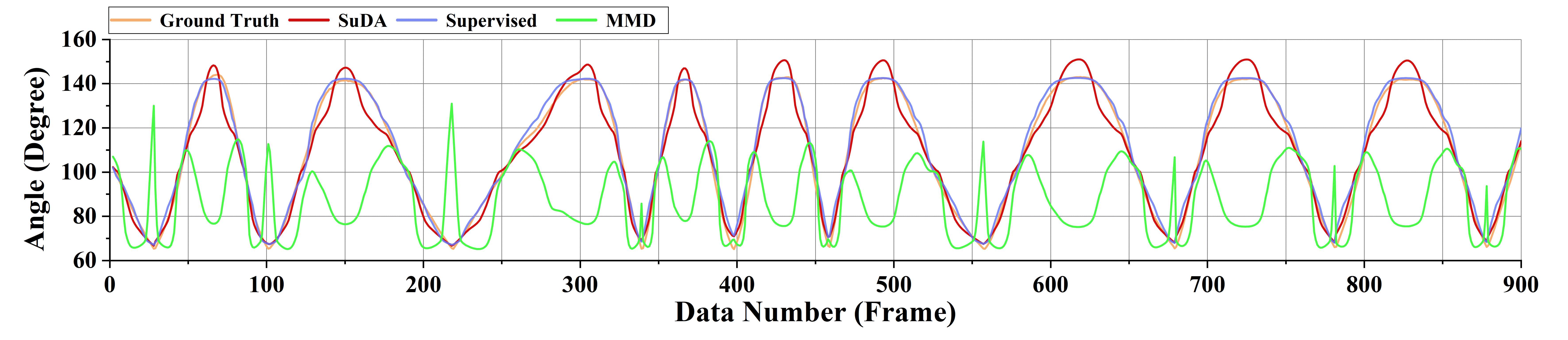}
\vspace{-5mm}
\caption{A plot of the detailed predict results on $\mathcal{D}(id=1)$}
\label{fig:User}
\end{figure*}

\begin{figure*}[h]
\centering
\includegraphics[width=0.9\linewidth]{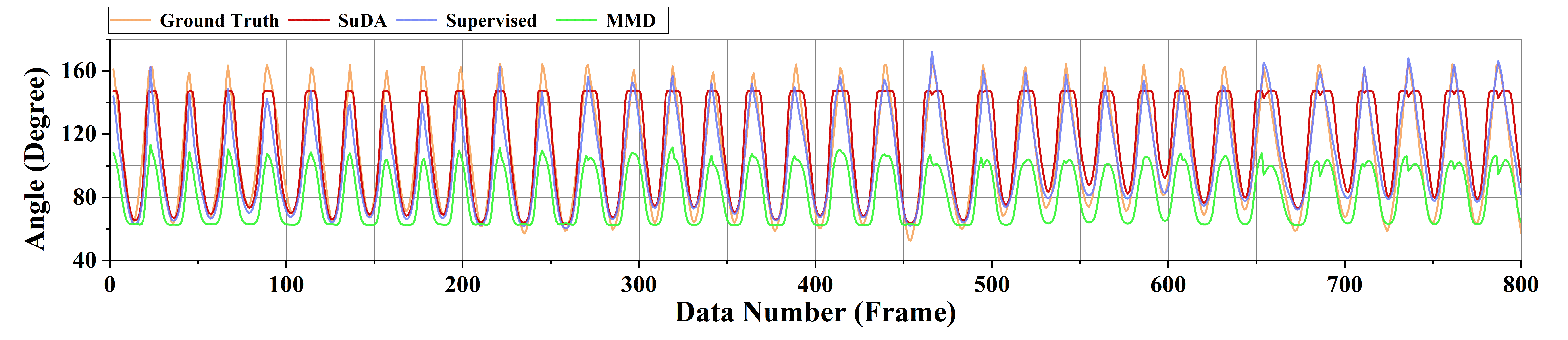}
\vspace{-5mm}
\caption{A plot of the detailed predict results on $\mathcal{D}(wp=180^\circ)$ }
\label{fig:Position}
\end{figure*}

\begin{figure*}[ht!]
\centering
\includegraphics[width=0.9\linewidth]{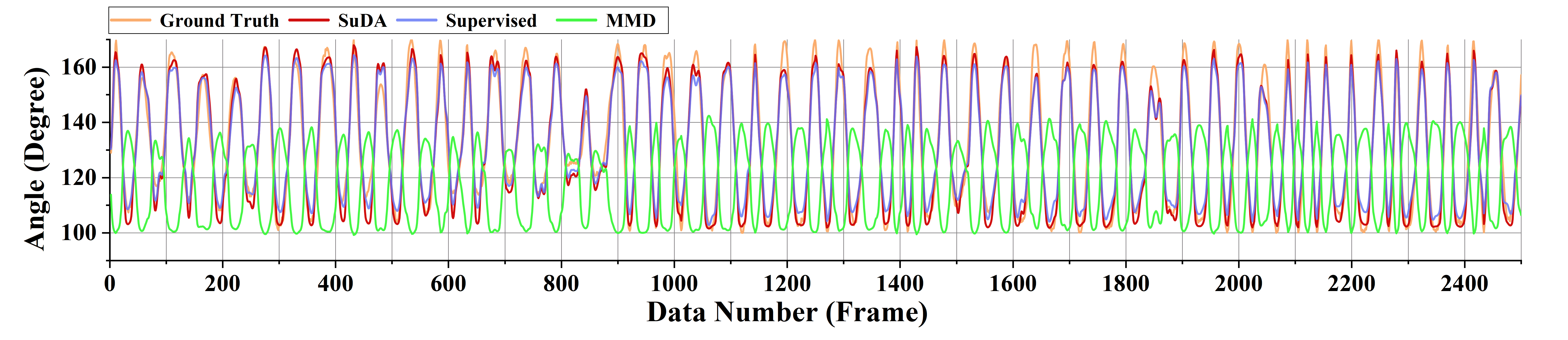}
\vspace{-5mm}
\caption{A plot of the detailed predict results on $\mathcal{D}(m=run)$}
\label{fig:Motion}
\end{figure*}

Figure \ref{fig:performance visualization}shows the Performance Visualization of our method SuDA. Figure \ref{fig:User}, \ref{fig:Position}, and \ref{fig:Motion} show that {\it \textbf{SuDA}} achieves comparable motion capture predictive performance in different Users, Wearing Positions and motions, although the source data (Simulation) is quite different from target data (Real world).
  % \vspace{-15pt}
\vspace{-5mm}
\vspace{-5mm}

\newpage
\section{Real World Dataset Description}
The real-world data collected was calculated using the three-dimensional positions of markers worn on the volunteers' elbow joints to determine real-time angle labels. The MARS 2H motion capture system, produced by NOKOV 3 company, tracked and calculated the three-dimensional positions of the markers at a rate of 60 frames per second. The motion capture system's three-dimensional tracking error is ±15mm, and the estimated rotation tracking error is a maximum of 0.88. 

\begin{table*}[t]
\centering
\begin{tabular}{lccccc}
\\
\hline
User ID   & upperc(cm) & lowerc (cm) & height (cm) & weight (kg) & Gender \\
\hline
ID1  & 24.6       & 24          & 186         & 78.3        & Female \\
ID2  & 24         & 23.5        & 174         & 61          & Female \\
ID3  & 25.8       & 24          & 171         & 69          & Male   \\
ID4  & 24.3       & 23.8        & 164         & 54.6        & Famale \\
ID5  & 25         & 24.5        & 170         & 60.2        & Male   \\
ID6  & 28         & 27          & 160         & 72.2        & Male   \\
ID7  & 23         & 22          & 158         & 50.2        & Female \\
ID8  & 25         & 24.2        & 164         & 61          & Male   \\
ID9  & 26.3       & 27.5        & 170         & 65.4        & Male   \\
ID10 & 30         & 28.5        & 175         & 86          & Male   \\
ID11 & 29         & 27.6        & 177         & 75          & Male  \\
\hline
\end{tabular}
\caption{Body profile details of the eleven experiment participants}
\label{table:User}
\end{table*}
\begin{table*}[t]
    \centering
    % \scalebox{0.8}{
    \begin{tabular}{c c c c c c c}
        \toprule
        User ID   & Gender & Height (cm) &Weight (kg) & Arm circumference (cm) \\
         \midrule
        1       &Male    & 179 & 84 &  30.2\\
        2       &Male    & 175 & 86 &  28.5\\
        3       &Male    & 177 & 75 &  27.6\\
        4       &Female  & 172 & 67 &  27.0\\
        5       &Female  & 168 & 65 &  26.6\\
        \bottomrule
    \end{tabular}
    \caption{Body profile details of the five real world scenarios experiment participants}
    \label{tab:body profile}
    % }
\end{table*}
% \begin{table*}[t]
%     \centering
%     % \scalebox{0.8}{
%     \begin{tabular}{c c c c c c c c}
%         \toprule
%         User ID   & Gender & Height (cm) &Weight (kg) & Left Arm & Right Arm  &Left Leg &Right Leg\\
%          \midrule
%         1       &Male    & 181 & 75 &  30.2 & 1 &1 &1  \\
%         2       &Male    & 178 & 68 &  30.2 & 1 &1 &1 \\
%         3       &Male    & 176 & 70 &  30.2 & 1 &37 &1 \\
%         \bottomrule
%     \end{tabular}
%     \caption{Body profile details of the three joints experiment participants}
%     \label{tab:joint profile}
%     % }
% \end{table*}
\begin{table}[t]
\centering
\begin{tabular}{cc}
\hline
\textbf{Class}            & \textbf{Dataset Size} \\ \hline
Chicken Dance       & 571          \\ 
House Dance         & 2376         \\ 
House Dance2        & 3142         \\ 
Macarena Dance      & 988          \\ 
Robot Hip Hop Dance & 1852         \\ 
Robot Hip Hop Dance & 2798         \\ 
Swing Dancing2      & 297          \\ 
Swing Dancing3      & 624          \\ 
Swing Dancing4      & 2965         \\ 
Swing Dancing5      & 2510         \\ 
Tut Hip Hop Dance   & 2032         \\ 
\textbf{Sum}                 & 20155        \\ 
\hline
\end{tabular}
\caption{Motion types of simulation datasets.}
\end{table}

\paragraph{Users}
We collected 11 volunteers as our User data, which can be denoted as $D(id = 1, 2, ...,11)$. Their body shape details can be seen in Table ~\ref{table:User}. Our User dataset consists of a total of 219, 932 frames.
\paragraph{Positions}
Considering that users of elbow pads may have different wearing habits, we took into account different wearing positions when collecting data. We explore the effect caused by circular types of offsets. We moved the elbow pad in 30-degree increments for circular offset to collect data, denoted as $D(wp = 30 ^\circ, 60 ^\circ, ..., 360 ^\circ)$. Our position dataset consists of a total of 155, 296 frames.
\paragraph{Motions}
Motion data of four basic motions, which can be denoted as $D(m = run, walk, jump, clap)$. Our Motion dataset consists of a total of 143, 575 frames.
\paragraph{Sum}
After data cleansing and calibration, we finally collected 371, 611 frames as our aggregate data, which can be denoted as $\mathcal{D}^{(id, m, wp)}_{t^*}$.
\paragraph{Applications of Real-world scenarios}
The real-world scenarios ((\textit{table tennis, basketball, football, boxing})) participants can be seen in Table \ref{tab:body profile}. Our Real world dataset consists of a total of 81, 848 frames. Our different joints applications ((\textit{Left elbow, right elbow, left knee, right knee})) datasets comprises 22, 577 frames.

\section{Simulation Framework}
% To address the effect of different body sizes on sensor parameters, we used the SMPL model to create fbx files of simulated human 3D models of different body sizes.

We used the SMPL model to create fbx files of simulated human 3D models.
Note that the skeleton model discussed in the following steps is a standard human skeleton obtained via MotionBuilder binding, with 68 joints instead of the 24 joints native to SMPL.
The entire process of synthesizing digital mannequins through SMPL models can be divided into three major stages as follows.

\subsection{SMPL Establishment}
\paragraph{Shape Blend Shapes}  Firstly, we established a mean template $\overline{T}$ as the basic pose of the whole human body, which was obtained statistically by using N=6,890 vertexes to represent the whole mesh. Then we described the pose of the human body and the offset of the basic pose through the parameters and superimposed them to form the final desired human pose. The process described above is linear, which contains a matrix multiplication procedure for parametric linear matrices, and we will continue to discuss it next. The pose of the human mesh obtained here is called the T-pose because it does not consider the influence of the pose parameters.

\paragraph{Pose Blend Shapes } When we specified the shape of the human mesh according to the specified $\beta$ parameter, we got a mesh with a specific height and weight. Nevertheless, we knew that a specific action might affect the specific shape of the local human body changes; in other words, the pose parameter $\theta$ will also affect the silent pose mesh shape to some extent.

\paragraph{Skinning} We only calculated the mesh in the static pose in the previous two stages. When human skeleton joints move, the "skin" composed of vertexes will change with the movement of the skeleton joints; this process is called skinning. The skinning process can be considered a weighted linear combination of skin nodes with changes in skeletal joints. However, the closer an endpoint is to a specific bone joint, the stronger the effect of changes, such as rotation and translation, with that bone joint.

\subsection{Motion Binding}
To address the effect of different motions on sensor parameters, we collected approximately 30,000 frames (some samples are increased to balance the weight) of motion capture data from the free online animation library Mixamo, which contains motion capture files for walking, running, clapping, and hybrid motions. Furthermore, we used the online binding function to bind the motions to the 3D human model generated based on SMPL.
 % Motion samples can be found in Figure \ref{fig:SimMotion}.
\subsection{Fabric Simulation}
The creases produced by the flexible sensors when worn and the differences in wearing position are also sources of Sim2Real domain gaps.
In order to solve these problems, we used the cloth database built into the clothing design software Marvelous Designer. Considering the basic parameters such as elasticity, friction, and thickness of real elbow pads and visual comparison inspection of simulation results, we selected hard-jeanet from the database. The simulation can produce similar wrinkles, pulling, and other phenomena as real motion scenes.
% The creases produced by the flexible sensors when worn and the differences in wearing position are also sources of errors.
% In order to solve these problems. We took the following steps.
% \begin{itemize}
%   \item [1)] 
%   We used the cloth database built into the clothing design software Marvelous Designer. Considering the basic parameters such as elasticity, friction, and thickness of real elbow pads and visual comparison inspection of simulation results, we selected hard-jeanet from the database. The simulation can produce similar wrinkles, pulling, and other phenomena as real motion scenes.      
%   \item [2)]
%   We exported the model file bound to the fabric and motion frame by frame. We used pygeodesic to calculate the length variation of the sensor virtualized by 3D modeling and fabric simulation.
%   \item [3)]
%   Based on the mapping between the real sensor shape variable and the resistance value, the virtual sensor signal is derived.
% \end{itemize}

\subsection{Elbow Bending Angle and Sensor Readings Simulation}
Our task is to estimate the bending angle of the elbow joint based on two flexible sensors surrounding the arm. Previously, it usually required professional motion capture (mocap) equipment to complete, {\it i.e.}, the wearer had to wear the mark points of the mocap equipment, and the data needed to be collected in a specific professional environment (as shown in Figure (b)). As soon as the data collection was completed, professional staff had to preprocess the collected data (Figure (c)). There is no doubt that data collection is highly labor-intensive and time-consuming.
To address this issue, our SuDA method employs a Sim2Real paradigm and no longer requires real data, thus significantly reducing the costs for flexible sensor-based motion capture.
 
\paragraph{Elbow Bending Angle Simulation}We use the Biovision (BVH) file to calculate the elbow angle of SMPL. BVH file is a general human feature animation file format containing the rotation data of the character's bone and limb files. Various popular animation software, such as Maya and 3DMax, are widely supported by it. The file mainly consists of two parts: skeleton information and data block:
\begin{itemize}
    \item Skeleton information: According to the hierarchical relationship, the required three-dimensional coordinate positions, such as the forearm, arm, hand, and Euler angles of rotation components, are defined to form a complete skeleton.
    \item Data block: the data information of each frame part is marked.
\end{itemize}

Since the 3D position of each part of the skeleton in the original BVH file is not easy to extract, hampering the subsequent input of neural network data, we use a script called bvh-converter to preprocess the original BVH file and obtain the file worldpose.csv that records the 3D coordinate position information of each frame of motion joint. The forearm, arm vector, and arm, hand vector are obtained through the 3D position calculation and then utilize formula $\cos \theta=\frac{\mathbf{V}_{1} \times \mathbf{V}_{2}}{\left|\mathbf{V}_{1} \times \mathbf{V}_{2}\right|}$ to obtain the elbow bending angle.

\paragraph{Sensor Readings Simulation}The Geodesic Distance of Fabric was used to calculate simulated sensor readings. First, we split the processed FBX file into a bunch of OBJ files, which contain the vertex data free-form curves, rendering attributes, and other information of each frame. Then we used \textbf{pygeodesic} to calculate the geodesic distance for simulative fabric. As Figure \ref{fig:Strain} shows, the soft sensors exhibit high linearity and repeatability ($R^2 = 0.99$) with low hysteresis.

%%%%%%%%%%%%%%%%%%%%%%%%%%%%%%%%%%%%%%%%%%%%%%%%%%%%%%%%%%%%%%%%%%%%%%%%%%%%%%%
%%%%%%%%%%%%%%%%%%%%%%%%%%%%%%%%%%%%%%%%%%%%%%%%%%%%%%%%%%%%%%%%%%%%%%%%%%%%%%%

\end{document}